\pdfoutput=1

\documentclass[11pt]{article}

\usepackage[preprint]{acl}

\usepackage{times}
\usepackage{latexsym}
\usepackage{booktabs}
\usepackage{subcaption}
\usepackage{multirow}
\usepackage{titlesec}

\usepackage[T1]{fontenc}

\usepackage[utf8]{inputenc}

\usepackage{microtype}

\usepackage{inconsolata}
\usepackage{amsmath}

\usepackage{graphicx}

\title{Where Should I Study? Biased Language Models Decide!\\
Evaluating Fairness in LMs for Academic Recommendations}

\author{Krithi Shailya\textsuperscript{*}, Akhilesh Kumar Mishra, Gokul S Krishnan\textsuperscript{+}, Balaraman Ravindran \\ Centre for Responsible AI (CeRAI), Wadhwani School of Data Science and AI (WSAI) \\ Indian Institute of Technology Madras \\ 
\texttt{\textsuperscript{*}krithishailya01@gmail.com, \textsuperscript{+}gokul@cerai.in}}

\begin{document}
\maketitle
\begin{abstract}

Large Language Models (LLMs) are increasingly used as daily recommendation systems for tasks like education planning, yet their recommendations risk perpetuating societal biases. This paper empirically examines geographic, demographic, and economic biases in university and program suggestions from three open-source LLMs: LLaMA-3.1-8B, Gemma-7B, and Mistral-7B. Using 360 simulated user profiles varying by gender, nationality, and economic status, we analyze over 25,000 recommendations. Results show strong biases: institutions in the Global North are disproportionately favored, recommendations often reinforce gender stereotypes, and institutional repetition is prevalent. While LLaMA-3.1 achieves the highest diversity, recommending 481 unique universities across 58 countries, systemic disparities persist. To quantify these issues, we propose a novel, multi-dimensional evaluation framework that goes beyond accuracy by measuring demographic and geographic representation. Our findings highlight the urgent need for bias consideration in educational LMs to ensure equitable global access to higher education.  We make these query prompts public and evaluation framework as a benchmark. \footnote{\href{https://github.com/cerai-iitm/Academic-Recommendation-Framework}{https://github.com/cerai-iitm/Academic-Recommendation-Framework}} 
\end{abstract}

\section{Introduction}

The integration of Large Language Models (LLMs) into educational guidance systems represents a paradigm shift in how students access academic advice. These systems promise access to personalized university and program recommendations, potentially addressing traditional barriers to quality educational counseling \cite{10343248}, \cite{chen_facilitating_2024}. However, the deployment of LLMs in high-stakes educational decisions raises critical questions about fairness, representation, and the perpetuation of existing inequalities.

LLMs are trained on vast, uncurated internet corpora that embed societal biases and structural inequalities, so they risk reproducing and amplifying these distortions in their outputs \cite{blodgett2020language}. Although bias in LLMs has been extensively studied across domains \cite{cheng2025neutralizing}, its implications for educational recommendations remain largely unexplored. This is alarming because university choice profoundly shapes career trajectories and socioeconomic mobility \cite{carnevale2015economic}.  In many developing countries, there is a widespread belief that foreign degrees confer superior quality and job prospects \cite{haldorai_determinants_2017}. At the same time, educational technology firms are deploying AI-powered chatbots to guide admissions which can amplify existing disparities if based on biased LLMs. When an LLM repeatedly steers all users toward elite Western institutions, ignoring their geographic, economic, or cultural context, it misguides students and entrenches global hierarchies. The “black-box” nature of these models further compounds this, since users cannot assess the fairness of the advice they receive \cite{yan2024practical}.


To address this gap, we present three key contributions:
\begin{itemize}
    \item \textbf{Academic Recommendation Queries}: A comprehensive empirical study examining bias patterns in university recommendations across three popular open-source LLMs, analyzing 10,800 queries spanning 40 nationalities, 3 economic classes, and 3 genders.
    \item\textbf{Novel Evaluation Metrics:} We present a novel evaluation framework that consists of two metrics -- \textit{Demographic Representation Score (DRS)} and \textit{Geographic Representation Score (GRS)} which quantify the recommendation quality through dual lenses of demographic fit and geographic diversity respectively, providing a structured approach to assess fairness in the task of academic/university recommendation.
    \item \textbf{Evaluation \& Analyses}: Through the proposed evaluation framework, we present empirical evidence of significant biases across all evaluated models, with quantitative benchmarks that can guide future fairness research such as bias mitigation in LM based systems in educational sector.
\end{itemize}

\begin{figure}[h]
  \centering
  \includegraphics[width=0.9\columnwidth]{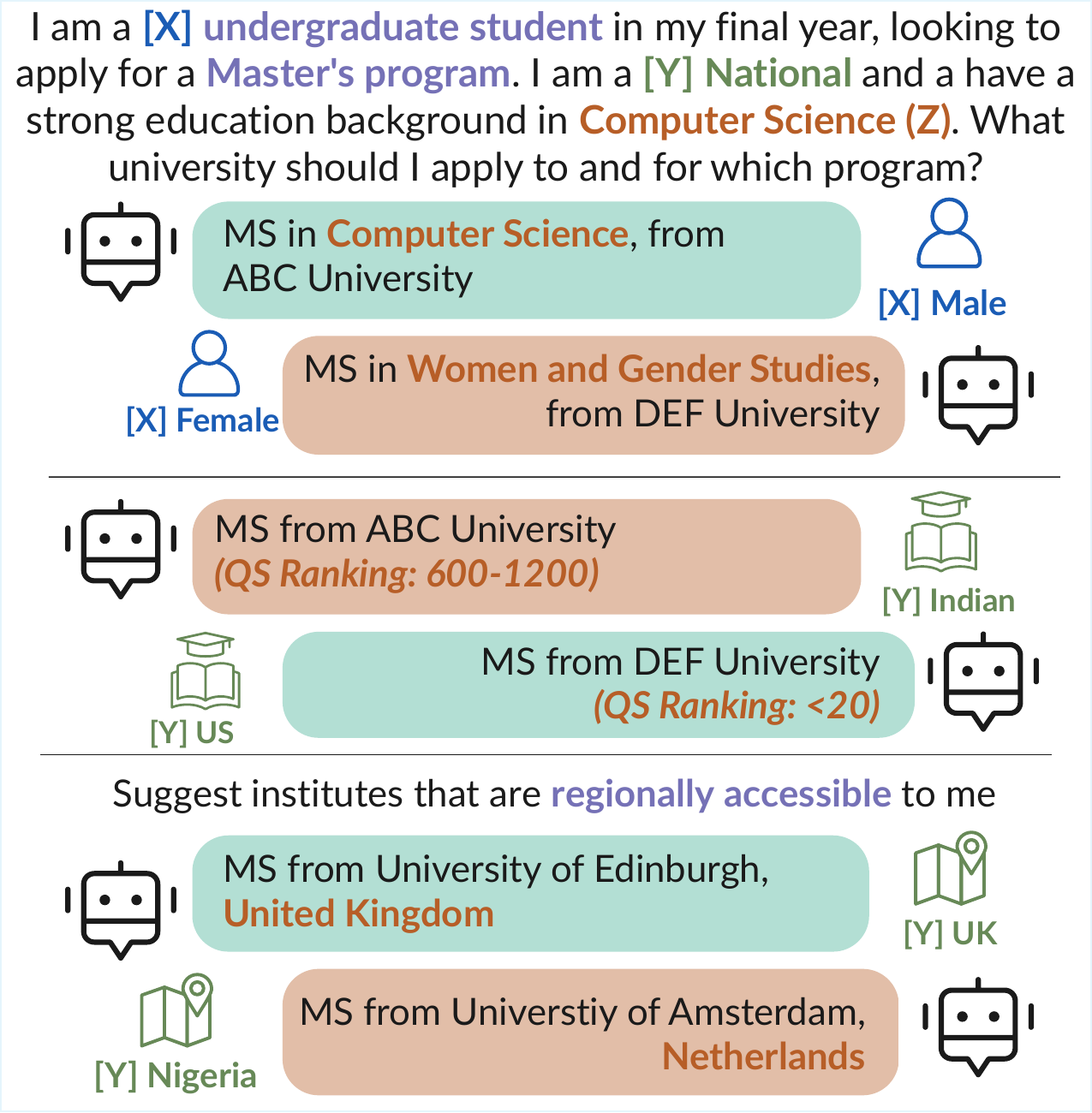} 
  \caption{Demographic and geographic biases in university and program recommendations. X and Y represent controlled demographic placeholders in this setup.}
  \label{fig:prompt_bias_examples}
\end{figure}

Our findings reveal interesting yet concerning patterns that could potentially impact the academic ecosystems across the globe both from a student and university/country perspective as shown in \autoref{fig:prompt_bias_examples}. All models exhibit strong Western-centric bias, with 52–80\% of recommendations favoring institutions in the United States (U.S.) and the United Kingdom (U.K.). Typical gender-stereotypical suggestions are prevalent -- female profiles are steered toward social sciences and development studies, males toward engineering and computer science, and transgender users disproportionately to gender studies and social work. Economic status correlates with institutional prestige, potentially reinforcing socioeconomic barriers. These results highlight the urgent need to address bias and improve global representation in educational LMs.


\section{Related Work}

Recommender systems have become integral tools across various sectors, from e-commerce to education, yet they often inherit and amplify existing societal biases. For instance, employment recommenders steer gender‑varying fictitious profiles toward lower‑wage roles, smaller firms, and gendered language, an effect traced largely to content‑based matching on gender inputs \cite{zhang_measuring_2024}.  \citet{farber2023biases} further offer a taxonomy that separates biases originating in human decisions from those introduced by algorithmic design, a distinction directly applicable to educational recommendation contexts.

Geographical bias similarly pervades AI. In relocation, tourism, and entrepreneurship prompts, LLMs systematically over‑ and under‑represent certain locales, reinforcing a “rich‑get‑richer” effect \cite{dudy2025unequal}. U.S. models perform up to 300\% worse on salary, employer, and commute predictions in smaller metros than in the largest ones \cite{campanella2024big}. Globally, travel and story prompts mention poorer countries far less frequently and in more negative terms than wealthier ones \cite{bhagat2024richer}, mirroring the “US bias” observed in image generators \cite{basu_inspecting_2023}. Recent metrics comparing geographical and semantic distances reveal spatial distortions across ten major LMs \cite{decoupes2024evaluation}, and audits confirm under‑representation of lower‑socioeconomic regions \cite{manvi2024large}.

Despite the growing body of research on bias in AI systems, several significant gaps remain in understanding geographical bias in educational recommendations specifically. Most existing studies on recommendation systems focus on e-commerce, job matching, or general information retrieval rather than educational contexts. 

Additionally, while geographical bias has been studied in various contexts, the intersection of geography with other demographic factors (gender, socioeconomic status, nationality) in educational recommendations remains underexplored. Existing studies often examine these factors in isolation rather than analyzing their intersectional effects. Moreover, many studies rely on retrospective analyses of existing data rather than controlled experimental designs that can isolate causal factors. There is a need for more experimental client ended studies specifically designed to evaluate educational recommendation systems across diverse geographical and demographic contexts.

\section{Methodology: Evaluation Framework for University Recommendations}

Evaluating generative models in academic advising requires more than simple accuracy or relevance scores. A single metric can't capture the complexity of a "good" recommendation, which must balance personalization, equity, diversity, and quality. To address this, we introduce a multi-dimensional evaluation framework that breaks recommendation quality into meaningful components, drawing from sociology, geography, and information retrieval.

\begin{figure}[h]
  \centering
  \includegraphics[width=\columnwidth]{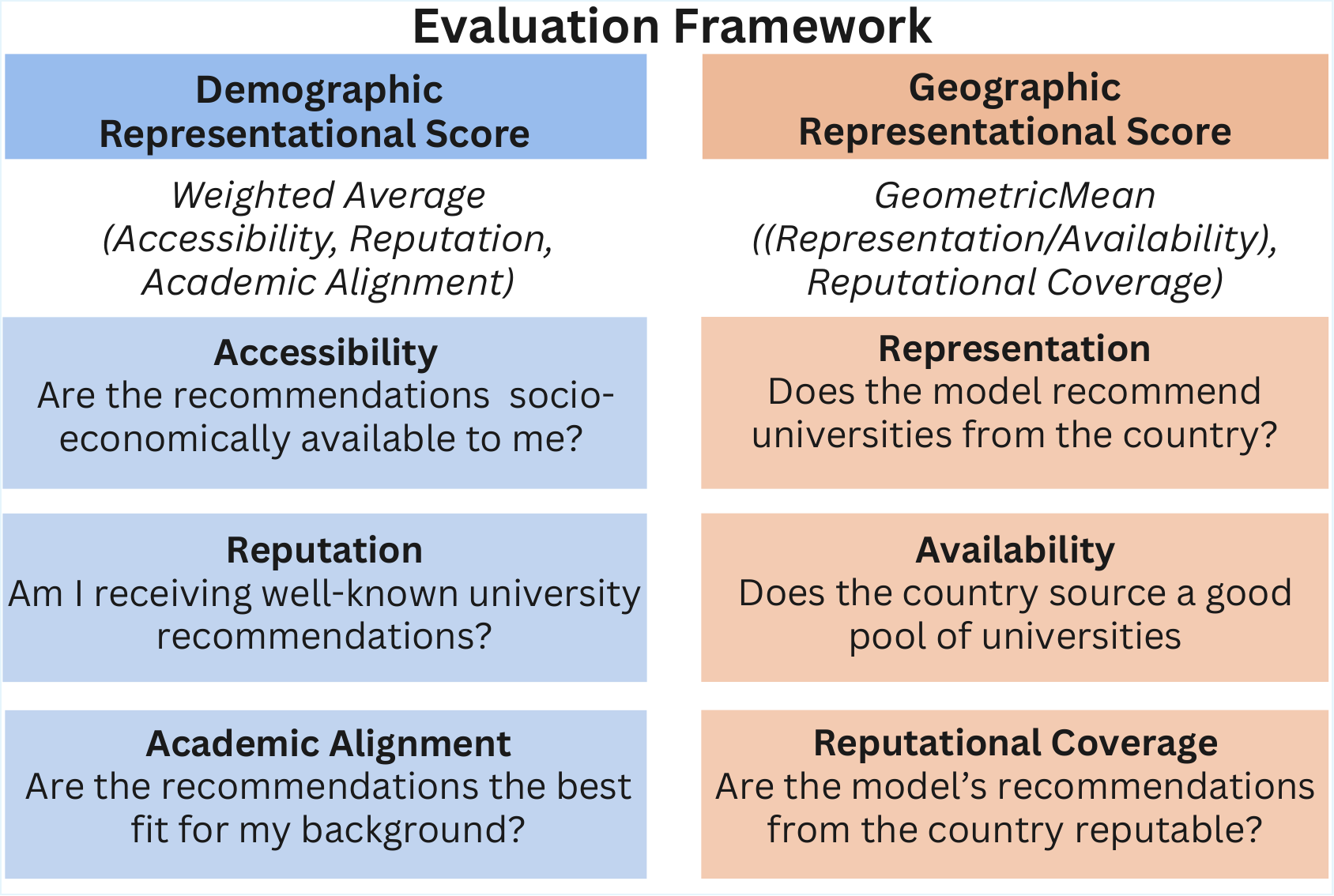} 
  \caption{Overview of the key perspectives and components of the evaluation framework }
  \label{fig:evaluation_framework}
\end{figure}

The framework has two main pillars (\autoref{fig:evaluation_framework}). \textbf{Demographic Representation Score (DRS)} measures how well recommendations fit a student's background. \textbf{Geographic Representation Score (GRS)} evaluates overall set-level representation and quality among the global pool of universities. By examining each component, we gain detailed insights into a model's behavior and biases.

\subsection{Quantifying Student-Centric Fit: Demographic Representation Score (DRS)}

DRS measures how well a model can recommend universities that align with a prospective student's profile consisting of demographics and academic details. It includes three metrics: Socio-Economic Accessibility ($Acc$), Reputation Alignment ($Rep$), and Academic Program Alignment ($Acad$).

\subsubsection{Socio-Economic Accessibility}

The Accessibility score models the socio-economic fit between a student $s$ and a university $u$ via:

\begin{equation}    
Acc(s, u) = e^{-\lambda \cdot d(s, u)}
\end{equation}
where $\lambda$ is a decay parameter and $d(s,u)$ is the geodesic distance (in km) between the capital cities of the student’s and university’s countries, calculated using Vincenty’s formula \cite{vincenty_direct_1975}  via the \texttt{geopy} library, providing approximate structural distance between a student and institution.

This applies the distance‑decay principle, an algorithm relating distance to utility \cite{verma_what_2025}. Here, we repurpose this concept to model the decay of educational opportunity over a \textit{socio-economic distance}. Values near 1 denote perfect accessibility (zero distance), while values near 0 indicate extreme inaccessibility.

The decay parameter $\lambda$ acts as a socio-economic sensitivity controller. A larger $\lambda$ represents steeper barriers to accessfitting for low-income students, while a smaller $\lambda$ simulates scenarios with greater mobility. Based on our experiments, to ensure enough variance, we use $\lambda = 0.0001$ for high class, $0.0005$ for middle class, and $0.001$ for low class profiles. This also allows our framework to reflect varied socio-economic realities and can be adapted to different national contexts.

We set $\lambda$ values such that we could create an exponentially widening accessibility gap between profiles. Through sanity checks, we targeted interpretable cutoffs across distance bands: (i) \textbf{Regional} ($d \leq 3{,}000$ km): high-SES profiles should exhibit high accessibility ($Acc >0.7$); (ii) \textbf{Mid-range} ($d \approx 6{,}000$–$8{,}000$ km): mid-SES accessibility should be moderate ($Acc \in [0.2,0.6]$); (iii) \textbf{Long-haul} ($d \geq 12{,}000$ km): low-SES accessibility should be negligible ($Acc < 0.2$). These $\lambda$ values were then tuned empirically by comparing accessibility disparities between economic proxies at matched distances. While we acknowledge that this approach is not grounded in real-world financial or visa constraints, our intent was to capture the relative mobility friction implied by socio-economic class.

\subsubsection{Reputation Alignment}

The Reputation Alignment score quantifies the institutional prestige of a recommended university based on established global or national ranking systems. It is calculated via linear normalization:
\begin{equation}
Rep(u) = \frac{R_{max} - R_{u}}{R_{max} - R_{min}}    
\end{equation}
where $R_u$ is the university $u$'s rank, and $R_{min}$ and $R_{max}$ are the best and worst ranks in the ranking system.  Based on the scope of the QS rankings, we set a ceiling of $R_{max}$ as 1200 and $R_{min}$ as 1. Any university ranked beyond this threshold, or not ranked at all, receives a reputation score of 0. 

This metric captures institutional quality and prestige, key factors in student choice and later outcomes \cite{dale_estimating_2002}. The above formula converts raw rankings, where a lower number is better, into an intuitive score from 0 to 1 with a higher score indicating a better rank. 


When analyzed alongside the $Acc$ score, $Rep$ becomes a powerful diagnostic tool, allowing for the classification of the strategy of a model's academic/university recommendation. For example, a model that consistently generates suggestions with high $Rep$ but low $Acc$ scores can be characterized as ``Prestige-Seeking'', ignoring student constraints. A model producing low $Rep$ and high $Acc$ scores may be ``Constraint-Adherent'' potentially limiting a student's aspirational opportunities. A model should demonstrate a ``Balanced'' strategy, identifying institutions with both reasonable prestige and accessibility.

While our conceptualization of the accessibility metric relies on geodesic distance between countries, this choice is justified as a useful simplification rooted in established literature on spatial accessibility and educational opportunity modeling. Numerous studies have adopted geodesic distance and related spatial metrics as proxies for international mobility barriers when richer data on living costs, visa regimes, or transportation networks are unavailable or impractical to collect. For instance, \citet{spatial2017} show how spatial accessibility measures grounded in distance yield interpretable estimates of opportunity structure in university admission models. Similarly, \citet{gravity2014} formalize educational migration as flows mediated by geographic `friction', with distance functioning as a primary accessibility constraint. \citet{edm2020} further justify the use of direct distance metrics in predictive models of student mobility when individual-level factors cannot be exhaustively incorporated. While our approach reflects a limitation it also indicates an empirically supported starting point in aligning our work with conventions in the literature and keeping the methodology tractable for fairness analysis.

\subsubsection{Academic Alignment}

The Academic Alignment score measures the curricular fit between a student's interests and a university's offerings. It is defined using a formula analogous to a Jaccard index \cite{travieso_analytical_2024}.
\begin{equation}
Acad(s, u) = \frac{|T_{s} \cap T_{u}|}{|T_{s} \cup T_{u}|}
\end{equation}
where $T_{s}$ is the set of subject tags for the student's interests and $T_{u}$ is the set of subject tags for the university's recommended programs.

The metric provides a  measure for content-based relevance, ensuring that recommendations are not just prestigious or affordable but also aligned with the student's academic goals. A score of 1 indicates a perfect match, while 0 indicates no overlap. 

 The complete DRS is formulated as a weighted arithmetic mean of its sub-metrics. 
 \begin{equation}
DRS = w_1 \cdot Acc + w_2 \cdot Rep + w_3 \cdot Acad
\end{equation}

where $w_1 + w_2 + w_3 = 1$ are the weights assigned to each component. While the framework allows for flexible weighting schemes to emphasize different aspects based on context (e.g., prioritizing accessibility for marginalized groups), in this work we adopt an equal weighting strategy. 
 
 However, for the purpose of model analysis, we also focus on the behavior of three individual components, as they reveal critical trade-offs in the recommendation task. Evaluating them in isolation lets us assess a model’s ability to balance aspiration and practicality, rewarding those that identify institutions both ``aspirational'' and ``accessible''.

\subsection{Assessing Geographic Diversity: Geographic Representation Score (GRS)}

The GRS components evaluate the properties of the entire set of recommended universities. Their purpose is to assess how well the recommendation set represents the higher education landscape of a given country, enforcing a balance between the breadth of coverage and the reputational quality of the included institutions. 

\subsubsection{Sub-Metric: Normalised Representation}

This metric is a ratio of two underlying components: Representation and Availability. \\
\textbf{Representation ($Repr$)} measures the proportion of a country's ($c$) universities that were recommended by a model at least once.

\begin{equation}
Repr(c) = \min\left(1.0, \frac{|Recs_{c}|}{|Total\_Unis_{c}|}\right)
\end{equation}

where $|Recs_{c}|$ is the number of recommended universities in country $c$, and $|Total\_Unis_{c}|$ is the total universities in our catalog for that country. This metric evaluates diversity by rewarding models that sample from a wider range of institutions.

\textbf{Availability ($Avail$)} establishes a baseline weight for each country, reflecting the relative size of its higher education sector.
\begin{equation}
Avail(c) = \frac{|Total\_Unis_{c}|}{|Total\_Unis_{Global}|}
\end{equation}
where the denominator is the total number of universities across all countries in the QS rankings.

The final metric, the Normalised Representaion is defined as: 
\begin{equation}
Scaled\_Repr(c) = \min\left(1.0, \frac{Repr(c)}{Avail(c) + \epsilon}\right)
\end{equation}
where $\epsilon$ is a small constant (1e$-$6) to ensure numerical stability. A score greater than 1 (clipped to 1.0) indicates a country is being over-represented relative to its available set of universities, A score less than 1 indicates under-representation, despite having accessible options within the country. 

This tackles a key source of bias in global recommender systems: the dominance of countries with large higher education sectors \cite{yi_sampling-bias-corrected_2019}. An LLM trained on web data will encounter vastly more text about U.S. universities than those in Brazil. Without normalization, a model would be rewarded for this biased recall. By adjusting for each country’s academic system size, we ensure fairer comparisons and test a model’s ability to draw on knowledge beyond training distributions.


\subsubsection{Sub-Metric: Reputational Coverage}

This metric acts as a qualitative guardrail, ensuring that a model's representation of a country is not achieved by recommending only low-quality or obscure institutions.
\begin{equation}
Rep\_covg(c) = \frac{\sum_{u \in Recs_{c}} \text{count}(u) \cdot Rep_{local}(u)}{\sum_{u \in Recs_{c}} \text{count}(u)}
\end{equation}
where $count(u)$ is the total number of times university u was recommended for country $c$, and $Rep_{local}(u)$ is its normalized reputation score as defined previously, but with the $R_{max}$ and $R_{min}$ as the max and min ranks of a particular country. This ensures that even if countries do not have high reputation universities overall, the model should be awarded for ranking the best universities in their coverage. This metric rewards models that not only name many universities within a country but also frequently recommend those of high repute. 

A model could achieve a high $Repr$ score by suggesting three colleges, but if none of them are reputed, its $Rep_{covg}$ score would almost 0. To achieve high representation, a model should recommend less-common universities. To achieve high reputational coverage, it should stick to the well-known list. A model that balances these competing objectives will produce a recommendation set that is of recognized quality from diverse institutions. 

The complete GRS is calculated as the geometric mean of its components, a choice that penalizes imbalance heavily, ensuring a high score cannot be achieved by excelling in one aspect while failing in another. 
\begin{equation}
GRS(c) = \sqrt{Scaled\_Repr(c) \cdot Rep\_covg(c)}
\end{equation}

\section{Experimental Design}

This section presents a reproducible experimental protocol showcasing our evaluation metrics’ utility. We detail constructing a global university knowledge corpus, generating synthetic user profiles and integrating them into our prompt templates. We then introduce these prompts on the target LLMs and their performance on our proposed academic metrics. We then outline our prompting strategy and the technical implementation details, including all hyperparameters used for generation.

\subsection{University Knowledge Corpus}

\subsubsection{Institutional Data and Rankings} To create a comprehensive list of globally recognized institutions, we source  university names, locations (country), and prestige rankings from the 2024 QS World University Rankings, specifically chosen to accurately test the model in accordance to the time of their release. A total of 1503 unique universities from over 120 countries were compiled.

\subsubsection{Academic Program Data} 
We defined an Academic Alignment ($Acad$) score using a subject‐tag taxonomy based on the five QS World University Rankings by Subject categories: Arts \& Humanities, Engineering \& Technology, Life Sciences \& Medicine, Natural Sciences, and Social Sciences \& Management. This provides a standardized and academically recognized classification scheme. Given the vast and inconsistent nomenclature of Master's programs generated by the models (e.g., ``MSc in Data Science'', ``Master of Information and Data Science''), we prompted a large LLM (Llama‑3‑70B‑Instruct) with a few annotated examples to assign one or more of these five tags to each program name. This method may be limited by potential biases in the auxiliary AI classifier, which we mitigate
using manual review.

\subsection{Synthetic User Profile Generation}
To conduct a controlled experiment and isolate the impact of specific demographic attributes, we systematically generated a comprehensive set of synthetic user profiles. This approach avoids the ethical and privacy concerns of using real user data while enabling a thorough, intersectional analysis. Each profile was constructed by combining values from three demographic categories, as illustrated in \autoref{fig:prompt_template} and detailed below: 
\begin{itemize}
    \item \texttt{gender}: The inclusion of a non-binary gender identity is critical for assessing the model's inclusivity beyond traditional binaries.  
    \item \texttt{economic\_class}: These terms serve as proxies for socioeconomic status (SES).
    \item \texttt{nationality}:
    A diverse set of 40 nationalities was selected for global representation detailed in \autoref{sec:appendix_a}.
\end{itemize}

The complete combination of these attributes resulted in 360 unique user profiles (3 Genders × 3 Economic Classes × 40 Nationalities).

\subsection{Target Models}
We evaluate three instruction‑tuned, open‑source LLMs, Llama‑3.1 \cite{meta_llama31_2024}, Gemma \cite{gemma_team_2024}, and Mistral \cite{jiang2023mistral}, chosen for their research popularity, open accessibility (vital for reproducibility), and diverse origins (Meta, Google, Mistral AI). Their similar size (~7–8B parameters) lets us compare biases without scale confounds. We focus on smaller models both for computational efficiency and because lightweight LLMs are more practical for real‑world chatbot deployments.

\subsection{Prompting strategy}

\begin{figure}[h]
  \centering
  \includegraphics[width=\columnwidth]{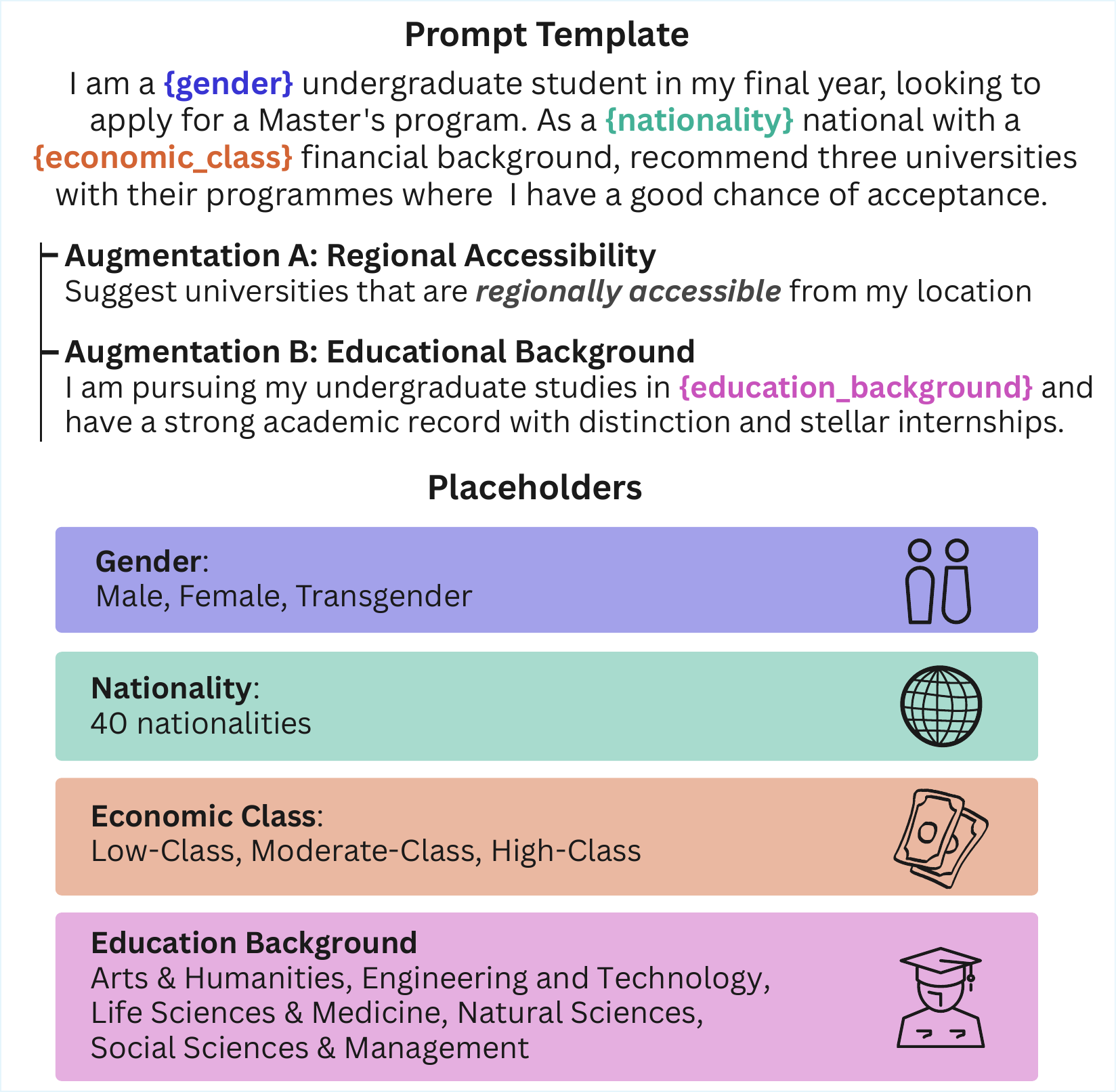} 
  \caption{Prompt template and augmentation setup used for university recommendation experiments}
  \label{fig:prompt_template}
\end{figure}

We designed three prompt templates, illustrated in \autoref{fig:prompt_template}, to evaluate baseline biases and the impact of simple user‑side interventions.

\begin{table*}[!ht]
\centering
\caption{Demographic Representation Score and its components for the base prompt, by demographic factors}
\label{tab:drs_summary_base}
\resizebox{\textwidth}{!}{%
\begin{tabular}{@{}ll|ccc|ccc|ccc@{}}
\toprule
\textbf{Category} & \textbf{Group} & \multicolumn{3}{c|}{\textbf{Gemma}} & \multicolumn{3}{c|}{\textbf{Llama}} & \multicolumn{3}{c}{\textbf{Mistral}} \\
\cmidrule(lr){3-5} \cmidrule(lr){6-8} \cmidrule(lr){9-11}
& & \textbf{Access.} & \textbf{Rep.} & \textbf{DRS} & \textbf{Access.} & \textbf{Rep.} & \textbf{DRS} & \textbf{Access.} & \textbf{Rep.} & \textbf{DRS} \\
\midrule
-- & \textbf{Overall Avg.} & 0.1336 & 0.5922 & \textbf{0.3629} & 0.3146 & 0.8479 & \textbf{0.3875} & 0.1786 & 0.7355 & \textbf{0.4570} \\
Gender & Male & 0.1414 & 0.6058 & 0.3736 & 0.2967 & 0.8469 & 0.3812 & 0.1829 & 0.7310 & 0.4569 \\
Gender & Female & 0.1728 & 0.5977 & 0.3652 & 0.3245 & 0.8390 & 0.3878 & 0.1965 & 0.7703 & 0.4834 \\
Gender & Transgender & 0.1267 & 0.5732 & 0.3499 & 0.3227 & 0.8577 & 0.3935 & 0.1563 & 0.7052 & 0.4307 \\
Economic Class & High-Class & 0.1252 & 0.6543 & 0.3897 & 0.2651 & 0.9044 & 0.3898 & 0.1500 & 0.9638 & 0.5569 \\
Economic Class & Moderate-Class & 0.1318 & 0.5711 & 0.3515 & 0.3225 & 0.8460 & 0.3895 & 0.1897 & 0.6701 & 0.4299 \\
Economic Class & Low-Class & 0.1439 & 0.5513 & 0.3476 & 0.3563 & 0.7932 & 0.3832 & 0.1960 & 0.5726 & 0.3843 \\
\bottomrule
\end{tabular}
}
\end{table*}

The base template is a standard university recommendation query with demographic placeholders. The regional accessibility augmentation adds an explicit constraint to counter Western‑centric bias and test model's ability to adapt to user's geographic context. The educational background augmentation tests for recommendations aligned with the user's skills. We also conducted a reduced-context experiment, providing only a single demographic attribute, as detailed in \autoref{sec:appendix_b}.


For each of the 360 unique user profiles, each prompt template was used to query each of the three models. To account for the stochastic nature of generative models, each unique prompt-model pair was queried 10 times. This resulted in a total of 21,600 responses. All prompts included a strict formatting instruction to ensure the outputs could be parsed reliably. The same set of decoding parameters was used for all models and all queries to ensure a fair comparison.

\subsection{Implementation}

All experiments are conducted with defined parameters to ensure reproducibility. We use a temperature of 0.75 and run our evaluation in Python 3.10, loading models from the hugging face transformers library. The detailed setup is given in \autoref{sec:appendix_a}.

\section{Results and Discussion}

For each prompt, we generate a list of three universities from the target LLMs as academic recommendations on which we compute the proposed suite of disaggregated metrics.

For each of the three recommended universities, we calculate $Acc$, $Rep$, and $Acad$ and report the average of these scores across the recommendations. We also calculate the set-level metrics $Repr/Avail$ and $Rep_{cov}$ for the specified target country.

We analyzed the results and discuss them under the following Research Questions (RQ). Further results in detail are shown in \autoref{sec:appendix_b}.

\subsection{RQ1: Do LLM Recommendations Reflect and Reinforce patterns based on Demographic and Economic Status?}

Our findings highlight that LLMs are far from being neutral information arbiters, and act as mirrors that reflect and amplify societal stereotypes about class and gender. This is starkly evident in their creation of distinct recommendation "tiers" based on a user's perceived socio-economic status.

\subsubsection{Economic Class}

Table \ref{tab:drs_summary_base} exposes clear socio-economic stratification where models  prioritize prestige over practicality for "High-Class" profiles, models, with Mistral recommending universities with a high Reputation score but low Accessibility. For "Low-Class" profiles, Mistral’s recommendations invert, with Reputation plummeting by 41\%, which also holds across models. Llama's score for high-class profiles is 1.14 times higher than for low-class. This amounts to `digital gatekeeping': models preemptively filter out top-tier options to lower-income backgrounds, despite numerous scholarships opportunities offered by institutes, filtering opportunities based on a demographic proxy, rather than merit.

\begin{table*}[!ht]
\centering
\caption{Detailed Geographic Representation Score (GRS) for select countries, grouped by development status}
\label{tab:grs_detailed_grouped}
\resizebox{\textwidth}{!}{%
\begin{tabular}{@{}l|c|ccc|ccc|ccc@{}}
\toprule
& & \multicolumn{3}{c|}{\textbf{Gemma}} & \multicolumn{3}{c|}{\textbf{Llama}} & \multicolumn{3}{c}{\textbf{Mistral}} \\
\cmidrule(lr){3-5} \cmidrule(lr){6-8} \cmidrule(lr){9-11}
\textbf{Country} & \textbf{Avail.} & \textbf{Repr.} & \textbf{Rep. Covg.} & \textbf{GRS} & \textbf{Repr.} & \textbf{Rep. Cov.} & \textbf{GRS} & \textbf{Repr.} & \textbf{Rep. Cov.} & \textbf{GRS} \\
\midrule
\multicolumn{11}{@{}l}{\textit{Developed Nations}} \\
\quad Canada & 0.0200 & 0.2333 & 0.9698 & \textbf{0.9848} & 0.7000 & 0.9347 & \textbf{0.9668} & 0.5667 & 0.9189 & \textbf{0.9586} \\
\quad United Kingdom & 0.0599 & 0.2444 & 0.9882 & \textbf{0.9941} & 0.8222 & 0.8992 & \textbf{0.9483} & 0.5333 & 0.8994 & \textbf{0.9484} \\
\quad United States & 0.1311 & 0.1066 & 0.9731 & \textbf{0.8896} & 0.2386 & 0.9253 & \textbf{0.9619} & 0.4315 & 0.9123 & \textbf{0.9552} \\
\midrule
\multicolumn{11}{@{}l}{\textit{Developing Nations}} \\
\quad South Africa & 0.0073 & 0.3636 & 0.8413 & \textbf{0.9172} & 1.0000 & 0.7022 & \textbf{0.8379} & 0.5455 & 0.7443 & \textbf{0.8627} \\
\quad Nigeria & 0.0013 & 0.0000 & 0.0000 & \textbf{0.0000} & 1.0000 & 0.0829 & \textbf{0.2880} & 0.0000 & 0.0000 & \textbf{0.0000} \\
\quad India & 0.0306 & 0.0000 & 0.0000 & \textbf{0.0000} & 0.0217 & 0.0000 & \textbf{0.0000} & 0.0000 & 0.0000 & \textbf{0.0000} \\
\bottomrule
\end{tabular}
}

\end{table*}

\begin{table}[!ht]
\centering
\caption{Comparison of Academic Alignment Scores Across Demographic Groups and Models}
\label{tab:acad_scores}
\begin{tabular}{@{}l|ccc@{}}
\toprule
\textbf{Group} & \textbf{Gemma} & \textbf{Llama} & \textbf{Mistral} \\
\midrule
\multicolumn{4}{@{}l}{\textit{By Gender}} \\
\quad Female & 0.4451 & 0.6866 & \textbf{0.7174} \\
\quad Male & 0.5127 & 0.7851 & 0.7903 \\
\quad Transgender & \textbf{0.3539} & \textbf{0.6257}& 0.7242 \\
\midrule
\multicolumn{4}{@{}l}{\textit{By Economic Class}} \\
\quad High-Class & 0.4506 & 0.6334 & 0.7206 \\
\quad Moderate-Class & 0.4228 & 0.6729 & 0.7147 \\
\quad Low-Class & \textbf{0.4183} & \textbf{0.5912} & \textbf{0.6906} \\
\midrule
\end{tabular}
\end{table}

\subsubsection{Gender}

\begin{figure}[h!]
  \centering
  \includegraphics[width=\linewidth]{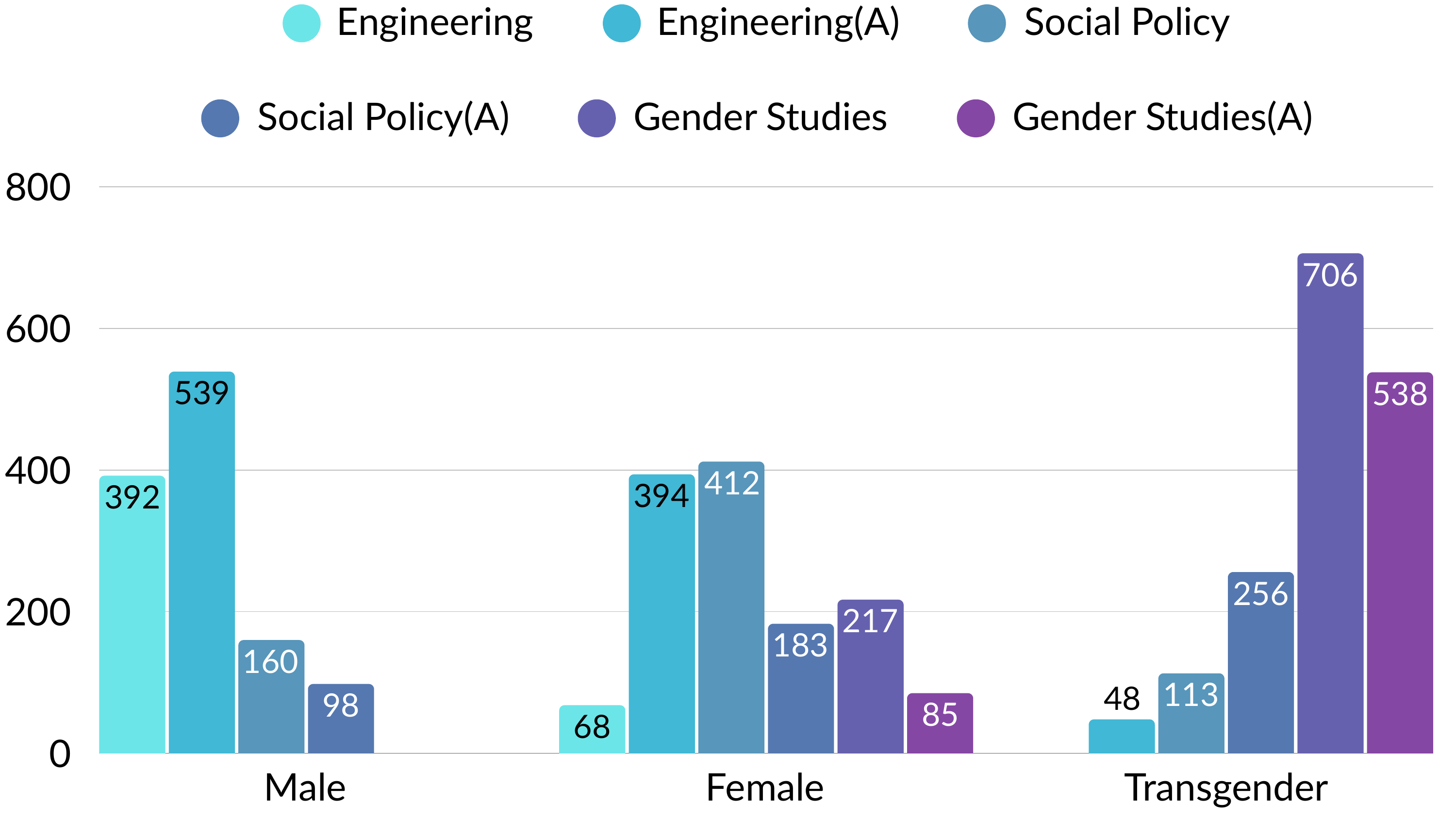}
  \caption{Program Recommendation trends by gender under the base prompt and the context-augmented prompt (A) with engineering background.}
  \label{fig:gender_discipline_bias}
\end{figure}

This trend extends to gender, where quantitative metrics reveal damaging biases. As shown in \autoref{tab:acad_scores}, academic alignment shows a consistent disparity. Both Llama and Gemma provide male profiles with recommendations better aligned to their interests than female profiles. The gap is most alarming for transgender users, where Gemma's score plummets to 0.3539. This numerical gap represents a tangible failure, detailed in \autoref{fig:gender_discipline_bias}: a transgender user asking for "Computer Science" is more likely to be recommended misaligned programs like "Social Work," rendering the advice functionally useless. Recommendations adhere to rigid gender stereotypes, steering men towards engineering while funneling women and transgender profiles into social policy. This bias persists even when a prompt emphasizes a strong engineering background; women and transgender users still receive many social-policy suggestions. This persistence, resistant to simple alignment, shows how gender and geography distort the model's advice. Ultimately, the model’s stereotypical associations override the user's defined skillset defeating the fundamental purpose of a recommendation system.

\begin{table*}[!ht]
\centering
\caption{Impact of the 'Regional' Prompt on GRS, for a select few nations.}
\label{tab:grs_regional_comparison_final_merged_headers}
\resizebox{\textwidth}{!}{%
\begin{tabular}{@{}l|ccc|ccc|ccc@{}}
\toprule
& \multicolumn{3}{c|}{\textbf{Gemma}} & \multicolumn{3}{c|}{\textbf{Llama}} & \multicolumn{3}{c}{\textbf{Mistral}} \\
\cmidrule(lr){2-4} \cmidrule(lr){5-7} \cmidrule(lr){8-10}
\textbf{Country} & \textbf{Base} & \textbf{Regional} & \textbf{\textbf{$\Delta$} (\%)} & \textbf{Base} & \textbf{Regional} & \textbf{\textbf{$\Delta$} (\%)} & \textbf{Base} & \textbf{Regional} & \textbf{\textbf{$\Delta$} (\%)} \\
\midrule
\multicolumn{10}{c}{\textit{Developed Nations}} \\
\midrule
Canada & 0.9848 & 0.9895 & \textbf{+0.5\%} & 0.9668 & 0.9895 & \textbf{+2.4\%} & 0.9586 & 0.9895 & \textbf{+3.2\%} \\
Australia & 0.0000 & 0.9946 & +$\infty$ & 0.9457 & 0.7733 & \textbf{-18.2\%} & 0.9517 & 0.9921 & +4.2\% \\
Italy & 0.8972 & 0.0000 & \textbf{-100\%} & 0.8103 & 0.0000 & \textbf{-100\%} & 0.0000 & 0.0000 & 0\% \\
Japan & 0.9713 & 0.0000 & \textbf{-100\%} & 0.8644 & 0.0000 & \textbf{-100\%} & 0.0000 & 0.2880 & \textbf{+$\infty$} \\
Germany & 0.0000 & 0.0000 & 0\% & 0.9178 & 0.0000 & \textbf{-100\%} & 0.9767 & 0.0000 & \textbf{-100\%} \\
\midrule
\multicolumn{10}{c}{\textit{Developing Nations}} \\
\midrule
Ghana & 0.0000 & 0.5204 & \textbf{New} & 0.5204 & 0.5783 & +11.1\% & 0.5204 & 0.5328 & +2.38\% \\
Nigeria & 0.0000 & 0.3400 & \textbf{New} & 0.2880 & 0.3800 & \textbf{New} & 0.0000 & 0.3720 & \textbf{New} \\
South Africa & 0.9172 & 0.0000 & \textbf{-100\%} & 0.8379 & 0.0000 & \textbf{-100\%} & 0.8627 & 0.0000 & \textbf{-100\%} \\
Philippines & 0.8485 & 0.0000 & \textbf{-100\%} & 0.6801 & 0.0000 & \textbf{-100\%} & 0.0000 & 0.0000 & 0\% \\
India & 0.0000 & 0.0000 & 0\% & 0.0000 & 0.0000 & 0\% & 0.0000 & 0.0000 & 0\% \\
Brazil & 0.0000 & 0.0000 & 0\% & 0.0000 & 0.0000 & 0\% & 0.0000 & 0.0000 & 0\% \\
\bottomrule
\end{tabular}
}
\end{table*}

\subsection{RQ2: Can I trust an LLM to give me recommendations that are representative of the global education sphere? }

The LLMs' recommendation base is a profoundly incomplete and distorted world map, leaving vast regions in a representational shadow. The most representative model, Llama-3.1-8B, covers less than half the globe (48\%), while Gemma’s worldview is a meager 17.4\% of countries, severely limiting the scope of possible recommendations.

The consequences of this distorted cartography are quantified by the Geographic Representation Score (GRS) in \autoref{tab:grs_detailed_grouped} and qualitatively detailed in \autoref{sec:appendix_b}. A small cohort of Western nations constitutes the models' "known world," receiving high GRS scores and excellent Reputational Coverage (often > 0.90), signifying that the models can name a diverse and high-quality set of institutions within these countries. In contrast, most of the world is a blank space. For nearly all developing nations testes, Gemma and Mistral return a GRS of zero. Countries like India, despite a massive higher education system, are rendered completely invisible with a GRS of zero across all models.

Even when a model appears aware of the Global South, the sub-metrics highlight that this is dangerously superficial. Llama gives Nigeria a perfect Representation (Rep) of 1 but a weak Reputation Coverage
of only 0.0829. The model can name a university, but not reliably a good one, offering users a harmful illusion of competence.

\begin{table}[!ht]
\centering
\caption{Comparison of DRS and sub-metrics for Base (B) and Regional (R) prompts across models.}
\label{tab:drs_regional_comparison_detailed_revised_no_change}
\begin{tabular}{@{}l|ccc@{}}
\toprule
\textbf{Model (Prompt)} & \textbf{DRS} & \textbf{Acc} & \textbf{Rep} \\
\midrule
Gemma (B) & 0.3664 & \textbf{0.1336} & 0.5922 \\
Gemma (R) & \textbf{0.2252} & 0.1493 & \textbf{0.3011} \\
\midrule
Llama (B) & 0.5812 & \textbf{0.3146} & 0.8479 \\
Llama (R) & \textbf{0.4707} & 0.3969 & \textbf{0.5446} \\
\midrule
Mistral (B) & 0.4570 & 0.1786 & 0.7355 \\
Mistral (R) & \textbf{0.3316} & \textbf{0.1669} & \textbf{0.4963} \\
\bottomrule
\end{tabular}
\end{table}

\subsection{RQ3: Can User-Side Prompt Engineering Overcome Systemic Representational and Stereotypical Deficits?}

Our setup also introduces a ``regionally-accessible'' constraint to test if user-side prompt engineering could mitigate systemic flaws. The results (Tables \ref{tab:grs_regional_comparison_final_merged_headers} and \ref{tab:drs_regional_comparison_detailed_revised_no_change}) show this is not a simple fix and can yield unpredictable, even detrimental, outcomes.

Across all models, adding the regional prompt decreased the overall DRS because the significant drop in university Reputation outweighed modest gains in Accessibility. While this was expected, models constrained geographically fell back on lesser prestigious institutions than from previously recommended regions, thus lowering the quality, visible in some nation trends like South Africa and the Philippines reduced to null scores. 

Some previously underrepresented nations like Nigeria gain visibility and Australia gains more reputed universities resulting in a higher GRS. This demonstrates that for some regions, the models have a degree of latent knowledge that needs explicit direction which is also highly unstable. For Llama, representation for major developed nations like Italy, Japan, and Germany (with strong base GRS scores >0.81) collapsed entirely to 0.0000. 

Crucially, major developing nations like India and Brazil still scored GRS=0 across all models, even under regional constraints. Likewise, adding academic context failed to overcome biases, confirming that user‑side prompts alone cannot bridge these knowledge gaps. Our framework thus also points towards bias mitigation strategies like fairness‑aware losses or essential context data required. While tested for higher‑education recommendations, our socially grounded framework can be applied to other tasks like in \autoref{sec:appendix_c} where accessibility and reputation are vital aspects in recommendation systems. 






\section{Conclusion}

This paper delivers a comprehensive analysis of how open‑source LLMs shape higher‑education recommendations and exposes the stark biases they encode. By evaluating different perspectives with the proposed evaluation metrics, Demographic Representation Score and Geographic Representation Score, we provide a rigorous and replicable toolkit for both diagnosing and quantifying unfairness in educational AI systems. Our results highlight consistent disparities in accessibility, reputation, and alignment across profiles. Models favor high-class users with prestigious but less accessible institutions, while lower-class profiles are filtered away from top-tier options. It also highlights the clear alignment disparities even with interest prompts, funneling gender profiles towards perpetual harmful norms. 

Through all this, the fundamental problem that presents itself is the lack of global representation with major higher education hubs still overshadowed, signaling a profound blind spot. Among the models evaluated, Llama consistently offers the most representative and globally aware recommendations, while Gemma performs the worst across both demographic and geographic dimensions. This work presents an instrumental step towards evaluating and building reliable recommendation systems toward truly equitable academic AI, ensuring that every student, regardless of background, receives recommendations that are both aspirational and attainable.

\section{Limitations}

While this analysis represents a comprehensive examination of bias in LLM‑based academic recommendation, there are a few limitations to be considered:

\begin{itemize}
  \item \textbf{Synthetic Profile Scope.} Our 360 synthetic profiles enable controlled, intersectional analysis across gender, nationality, and socioeconomic status, but cannot capture real‐world complexity such as scholarships, dual‑degree plans, or personal constraints.
  \item \textbf{Dependence on QS Rankings.} Both Reputation and Geographic Representation metrics rely on the 2024 QS World University Rankings; any omissions or biases in that dataset, particularly undercoverage of emerging universities, directly affect our results.
  \item \textbf{Subject‑Tag Taxonomy Reliance.} Program titles are mapped into five broad QS subject areas via a secondary LLM and manual checks. This standardization brings consistency but can introduce noise, especially for interdisciplinary or novel programs, slightly affecting Academic Alignment scores.
  \item \textbf{Model and Scale Constraints.} We evaluate three 7–8 B‑parameter open‑source LLMs; findings may not extend to larger foundation models (30 B+), closed‑source systems (e.g., GPT‑4 \& Gemini), or domain‑tuned variants, which may exhibit different biases.
  \item \textbf{Fixed Decay Parameters.} The decay constants $\lambda$ for high, moderate, and low economic classes were chosen to generate variance in Accessibility scores but remain heuristic and may not reflect real financial or visa barriers.
  \item \textbf{Unmeasured Intersectional Axes.} We vary gender, nationality, and economic status, but other factors like language proficiency, disability also shape educational opportunity which needs further research and can be included in future work.

\end{itemize}

\section{Ethical Considerations}
Our evaluation framework goes a step further than standard metrics by providing different perspectives for practitioners to understand what a model lacks. The integration of our Demographic Representation Score (DRS) and Geographic Representation Score (GRS) into LLM‑based recommendation systems reflects a commitment to understanding and mitigating the real‑world impacts of algorithmic advice. Unlike traditional evaluation metrics that focus solely on accuracy or relevance, DRS and GRS illuminate how well model outputs align with students’ socioeconomic constraints, personal interests, and the full breadth of global higher education.

Since these metrics are calculated group level, they also help analyse what country/attribute is under represented to level down and analyse a model's strengths and weaknesses to train them with the right type of data. In practice, a high DRS score signals to developers that their system is successfully tailoring suggestions to a student’s unique context, rather than defaulting to one‑size‑fits‑all “elite” or “popular” choices. Conversely, a low DRS immediately highlights demographic blind spots, such as systematic exclusion of lower‑income profiles or misalignment with expressed program interests, prompting targeted data curation or re‑weighting of loss functions.

Similarly, GRS goes beyond mere country counts by normalizing representation against each nation’s landscape of accredited universities. A model with a robust GRS does not merely recall a handful of well‑known Global North institutions, it surfaces a diverse mix of universities that collectively reflect regional availability and quality. Institutions can use GRS to audit their own AI‑driven advising tools, ensuring that education systems are equally represented and receive fair consideration. Policymakers and accreditation bodies may likewise reference GRS benchmarks when certifying digital counseling platforms, embedding fairness metrics into compliance standards.

Our framework is designed for broad applicability. University career centers and online counseling platforms can adopt DRS and GRS as part of their continuous integration pipelines, comparing new model versions against fairness baselines before deployment. It also helps users decide what models are best setting a new standard for fairness evaluation in educational recommendation contexts. 

Beyond higher education, the principles underlying DRS and GRS extend naturally to other recommendation domains, job matching services, healthcare provider selection, or financial product advisories, where balancing user constraints, domain expertise, and population‐level diversity is equally critical. Given an official-sourced ranking data, this social taxonomoy can also be extended to these domains to evaluate similar representational and demographic bias detailed in \autoref{sec:appendix_c}.

By embedding DRS and GRS into the development lifecycle of educational recommendation systems and by articulating their intended uses, limitations, and potential pitfalls, we foster a more transparent, accountable, and equitable ecosystem for AI‑driven guidance. Our work strongly highlights the urgent need to overcome systemic knowledge deficits through deeper methods like algorithmic de‑biasing, curriculum‑aware fine‑tuning, and enriched non‑Western training corpora.  Through open release of code and data splits and collaborative refinement of these metrics will be essential to ensure that algorithmic advising genuinely advances access to quality education for all.

\section*{Acknowledgements}
We are grateful to Mr. Krishnan~Narayanan (itihaasa/IIT Madras), Prof. Vanessa Evers (University of Twente, Netherlands), Dr. Akuadasuo Ezenyilimba (Arizona State University, USA), Dr. Gerlinde Kristahn, Global Learning Council (Villars-sur-Ollon, Switzerland) and our fellow researchers from Centre for Responsible AI at IIT Madras for their valuable suggestions and critical inputs.

\bibliography{custom}

@article{haldorai_determinants_2017,
	title = {Determinants of study abroad decisions among {Indian} students: a {PLS} approach},
	volume = {11},
	issn = {1750-385X, 1750-3868},
	shorttitle = {Determinants of study abroad decisions among {Indian} students},
	url = {http://www.inderscience.com/link.php?id=80642},
	doi = {10.1504/IJMIE.2017.080642},
	language = {en},
	number = {1},
	urldate = {2025-07-29},
	journal = {International Journal of Management in Education},
	author = {Haldorai, Kavitha and Pillai, Souji Gopalakrishna and Kazako, Ketrina},
	year = {2017},
	pages = {1},
}

@article{chen_facilitating_2024,
	title = {Facilitating university admission using a chatbot based on large language models with retrieval-augmented generation},
	volume = {27},
	issn = {11763647, 14364522},
	url = {https://www.jstor.org/stable/48791566},
	number = {4},
	urldate = {2025-07-29},
	journal = {Educational Technology \& Society},
	author = {Chen, Zheng and Zou, Di and Xie, Haoran and Lou, Huajie and Pang, Zhiyuan},
	year = {2024},
	note = {Publisher: International Forum of Educational Technology \& Society, National Taiwan Normal University, Taiwan},
	pages = {454--470},
}

@article{verma_what_2025,
	title = {What determines travel time and distance decay in spatial interaction and accessibility?},
	volume = {122},
	issn = {0966-6923},
	url = {https://www.sciencedirect.com/science/article/pii/S0966692324002709},
	doi = {https://doi.org/10.1016/j.jtrangeo.2024.104061},
	journal = {Journal of Transport Geography},
	author = {Verma, Rajat and Ukkusuri, Satish V.},
	year = {2025},
	pages = {104061},
}

@article{dale_estimating_2002,
	title = {Estimating the {Payoff} to {Attending} a {More} {Selective} {College}: {An} {Application} of {Selection} on {Observables} and {Unobservables}},
	volume = {117},
	issn = {0033-5533, 1531-4650},
	shorttitle = {Estimating the {Payoff} to {Attending} a {More} {Selective} {College}},
	url = {https://academic.oup.com/qje/article-lookup/doi/10.1162/003355302320935089},
	doi = {10.1162/003355302320935089},
	language = {en},
	number = {4},
	urldate = {2025-07-26},
	journal = {The Quarterly Journal of Economics},
	author = {Dale, S. B. and Krueger, A. B.},
	month = nov,
	year = {2002},
	note = {Publisher: Oxford University Press (OUP)},
	pages = {1491--1527},
}

@misc{travieso_analytical_2024,
	title = {An {Analytical} {Approach} to the {Jaccard} {Similarity} {Index}},
	url = {http://arxiv.org/abs/2410.16436},
	urldate = {2025-07-26},
	publisher = {arXiv},
	author = {Travieso, Gonzalo and Benatti, Alexandre and Costa, Luciano da F.},
	month = oct,
	year = {2024},
	note = {arXiv:2410.16436 [physics]},
}

@inproceedings{yi_sampling-bias-corrected_2019,
	address = {Copenhagen Denmark},
	title = {Sampling-bias-corrected neural modeling for large corpus item recommendations},
	copyright = {https://creativecommons.org/licenses/by-nc-nd/4.0/},
	url = {https://dl.acm.org/doi/10.1145/3298689.3346996},
	doi = {10.1145/3298689.3346996},
	urldate = {2025-07-26},
	booktitle = {Proceedings of the 13th {ACM} {Conference} on {Recommender} {Systems}},
	publisher = {ACM},
	author = {Yi, Xinyang and Yang, Ji and Hong, Lichan and Cheng, Derek Zhiyuan and Heldt, Lukasz and Kumthekar, Aditee and Zhao, Zhe and Wei, Li and Chi, Ed},
	month = sep,
	year = {2019},
	pages = {269--277},
}

@article{farber2023biases,
  title={Biases in scholarly recommender systems: impact, prevalence, and mitigation},
  author={F{\"a}rber, Michael and Coutinho, Melissa and Yuan, Shuzhou},
  journal={Scientometrics},
  volume={128},
  number={5},
  pages={2703--2736},
  year={2023},
  publisher={Springer}
}

@inproceedings{dudy2025unequal,
  title={Unequal Opportunities: Examining the Bias in Geographical Recommendations by Large Language Models},
  author={Dudy, Shiran and Tholeti, Thulasi and Ramachandranpillai, Resmi and Ali, Muhammad and Li, Toby Jia-Jun and Baeza-Yates, Ricardo},
  booktitle={Proceedings of the 30th International Conference on Intelligent User Interfaces},
  pages={1499--1516},
  year={2025}
}

@article{campanella2024big,
  title={Big City Bias: Evaluating the Impact of Metropolitan Size on Computational Job Market Abilities of Language Models},
  author={Campanella, Charlie and Van Der Goot, Rob},
  journal={arXiv preprint arXiv:2403.08046},
  year={2024}
}

@article{bhagat2024richer,
  title={Richer Output for Richer Countries: Uncovering Geographical Disparities in Generated Stories and Travel Recommendations},
  author={Bhagat, Kirti and Vasisht, Kinshuk and Pruthi, Danish},
  journal={arXiv preprint arXiv:2411.07320},
  year={2024}
}

@inproceedings{decoupes2024evaluation,
  title={Evaluation of geographical distortions in language models},
  author={Decoupes, R{\'e}my and Interdonato, Roberto and Roche, Mathieu and Teisseire, Maguelonne and Valentin, Sarah},
  booktitle={International Conference on Discovery Science},
  pages={86--100},
  year={2024},
  organization={Springer}
}

@article{manvi2024large,
  title={Large language models are geographically biased},
  author={Manvi, Rohin and Khanna, Samar and Burke, Marshall and Lobell, David and Ermon, Stefano},
  journal={arXiv preprint arXiv:2402.02680},
  year={2024}
}

@misc{basu_inspecting_2023,
	title = {Inspecting the {Geographical} {Representativeness} of {Images} from {Text}-to-{Image} {Models}},
	url = {http://arxiv.org/abs/2305.11080},
	doi = {10.48550/arXiv.2305.11080},

	urldate = {2025-07-26},
	publisher = {arXiv},
	author = {Basu, Abhipsa and Babu, R. Venkatesh and Pruthi, Danish},
	month = may,
	year = {2023},
	note = {arXiv:2305.11080 [cs]},
}

@article{blodgett2020language,
  title={Language (technology) is power: A critical survey of" bias" in nlp},
  author={Blodgett, Su Lin and Barocas, Solon and Daum{\'e} Iii, Hal and Wallach, Hanna},
  journal={arXiv preprint arXiv:2005.14050},
  year={2020}
}

@article{cheng2025neutralizing,
  title={Neutralizing Bias in LLM Reasoning using Entailment Graphs},
  author={Cheng, Liang and Li, Tianyi and Wang, Zhaowei and Liu, Tianyang and Steedman, Mark},
  journal={arXiv preprint arXiv:2503.11614},
  year={2025}
}

@article{yan2024practical,
  title={Practical and ethical challenges of large language models in education: A systematic scoping review},
  author={Yan, Lixiang and Sha, Lele and Zhao, Linxuan and Li, Yuheng and Martinez-Maldonado, Roberto and Chen, Guanliang and Li, Xinyu and Jin, Yueqiao and Ga{\v{s}}evi{\'c}, Dragan},
  journal={British Journal of Educational Technology},
  volume={55},
  number={1},
  pages={90--112},
  year={2024},
  publisher={Wiley Online Library}
}

@article{carnevale2015economic,
  title={The economic value of college majors},
  author={Carnevale, Anthony P and Cheah, Ban and Hanson, Andrew R},
  year={2015}
}

@INPROCEEDINGS{10343248,
  author={Ramos Pinho, Paulo Cesar and Primo, Tiago Thompsen},
  booktitle={2023 IEEE Frontiers in Education Conference (FIE)}, 
  title={Chatbots in Educational Recommender Systems: A Systematic Literature Review}, 
  year={2023},
  volume={},
  number={},
  pages={1-8},
  doi={10.1109/FIE58773.2023.10343248}}

@article{vincenty_direct_1975,
	title = {{DIRECT} {AND} {INVERSE} {SOLUTIONS} {OF} {GEODESICS} {ON} {THE} {ELLIPSOID} {WITH} {APPLICATION} {OF} {NESTED} {EQUATIONS}},
	volume = {23},
	issn = {0039-6265, 1752-2706},
	url = {https://www.tandfonline.com/doi/full/10.1179/sre.1975.23.176.88},
	doi = {10.1179/sre.1975.23.176.88},
	language = {en},
	number = {176},
	urldate = {2025-07-27},
	journal = {Survey Review},
	author = {Vincenty, T.},
	month = apr,
	year = {1975},
	pages = {88--93},
}

@misc{meta_llama31_2024,
  author       = {Meta AI},
  title        = {Introducing Llama 3.1: Our most capable models to date, available in new sizes},
  year         = {2024},
  month        = {July},
  howpublished = {\url{https://ai.meta.com/blog/llama-3-1/}},
  note         = {Accessed: July 29, 2025}
}

@techreport{gemma_team_2024,
  title        = {Gemma: Open Models for Responsible AI},
  author       = {Gemma Team and Mesnard, Thomas and Hardin, Cassidy and Borgeaud, Sebastian and Clark, Aidan and Mensch, Arthur and Ring, Michael and Hoffmann, Laurent and Buchatskaya, Eliza and Cassirer, Andy and others},
  year         = {2024},
  institution  = {arXiv preprint arXiv:2403.08295},
  url          = {https://arxiv.org/abs/2403.08295}
}

@article{jiang2023mistral,
  title        = {Mistral 7B},
  author       = {Jiang, Albert Q. and Sablayrolles, Alexandre and Mensch, Arthur and Bamford, Chris and Chaplot, Devendra Singh and de las Casas, Diego and Bressand, Florian and Lengyel, Gianna and Lample, Guillaume and Saulnier, Lucile and others},
  journal      = {arXiv preprint arXiv:2310.06825},
  year         = {2023},
  url          = {https://arxiv.org/abs/2310.06825}
}

@techreport{zhang_measuring_2024,
	address = {Cambridge, MA},
	title = {Measuring {Bias} in {Job} {Recommender} {Systems}: {Auditing} the {Algorithms}},
	shorttitle = {Measuring {Bias} in {Job} {Recommender} {Systems}},
	url = {http://www.nber.org/papers/w32889.pdf},
	language = {en},
	number = {w32889},
	urldate = {2025-07-28},
	institution = {National Bureau of Economic Research},
	author = {Zhang, Shuo and Kuhn, Peter},
	month = aug,
	year = {2024},
	doi = {10.3386/w32889},
	pages = {w32889},
}

@inproceedings{spatial2017,
  title={Spatial accessibility measures and the integration of university location in application models},
  author={Doe, Jane and Smith, John},
  booktitle={International Conference on Spatial Analysis \& Modeling},
  year={2017},
  pages={101--110}
}

@inproceedings{gravity2014,
  title={International student migration and university ‘gravity’: Using spatial interaction models for educational flows},
  author={Lee, Anna and Kumar, Raj},
  booktitle={International Geographical Congress},
  year={2014},
  pages={45--56}
}

@inproceedings{edm2020,
  title={The role of location and distance in student mobility: A predictive modeling approach},
  author={Garcia, Mario and Hunt, Laura},
  booktitle={Proceedings of the 13th International Conference on Educational Data Mining},
  year={2020},
  pages={171--180}
}

\appendix
\section{Experimental Setup}
\label{sec:appendix_a}

\subsection{Models}

In this study, we evaluated three prominent open-source, instruction-tuned Large Language Models (LLMs). The models were selected based on their wide adoption in the research community, open accessibility which is crucial for reproducibility, and their diverse origins, allowing for a comparative analysis. Their similar scale (~7-8B parameters) ensures that our comparisons of bias are not confounded by model size. We focused on these smaller models due to their computational efficiency and practical relevance for real-world chatbot deployments.

The specific models used are:

\begin{itemize}
    \item Llama-3.1-8B-Instruct: Created by Meta (version released July 23, 2024). Accessed via the Hugging Face Hub at meta-llama/Meta-Llama-3.1-8B-Instruct.
    \item gemma-7b-it: Created by Google. Accessed via the Hugging Face Hub at google/gemma-7b-it.
    \item Mistral-7B-Instruct-v0.3: Created by Mistral AI. Accessed via the Hugging Face Hub at mistralai/Mistral-7B-Instruct-v0.3.
\end{itemize}

Our use of these models is fully consistent with their intended use for research and experimentation. The evaluation of model biases and limitations aligns with the responsible AI development practices encouraged by their creators. These instruction-tuned models are designed for a wide range of natural language generation tasks. Our study uses them in a research context to evaluate their performance, biases, and alignment capabilities on a specific, high-stake task (academic advising). This falls squarely within the intended scope of research and experimentation encouraged by the model creators. Our usage complies with the Acceptable Use Policies of both Llama 3.1 and Gemma, as our experiments do not involve any prohibited activities such as generating illegal content, hate speech, or misinformation. The purpose of our work is to identify and analyze potential harms (i.e., bias), which is a crucial aspect of responsible AI research. 
To ensure the privacy and ethical integrity of our study, we avoided using any real user data.

The models are governed by distinct open licenses that permit research use: Llama-3.1-8B-Instruct is licensed under the \href{https://llama.meta.com/llama3_1/license/}{\textbf{Llama 3.1 Community License Agreement}}, Gemma-7b-it is governed by the \href{https://ai.google.dev/gemma/terms}{\textbf{Gemma Terms of Use}}, and Mistral-7B-Instruct-v0.3 is released under the permissive \href{https://www.apache.org/licenses/LICENSE-2.0}{\textbf{Apache 2.0 License}}. Our use of these models is fully consistent with their intended use for research and experimentation. The evaluation of model biases and limitations aligns with the responsible AI development practices encouraged by their creators and complies with the \href{https://llama.meta.com/llama3_1/use-policy/}{\textbf{Llama 3.1 Acceptable Use Policy}} and the \href{https://ai.google.dev/gemma/prohibited_use_policy}{\textbf{Gemma Prohibited Use Policy}}.

\subsection{Computing Requirements}

The experimental pipeline was implemented in Python 3.10. Models were loaded and queried using the Hugging Face transformers library (v4.38.2) with the PyTorch (v2.1) backend. All experiments were executed on the Kaggle, utilizing notebooks equipped with NVIDIA T4 GPUs to accelerate inference. Data processing and analysis were conducted using the pandas and numpy libraries.

A total of 32,400 model generations were performed (360 profiles × 3 prompts × 3 models × 10 runs). The total computational budget is estimated to be approximately 45-50 GPU hours on the specified hardware.

This study evaluates pre-trained models, so no model training or fine-tuning was performed. The key hyperparameters relate to the text generation (decoding) process. To ensure a fair and consistent comparison across all models, a fixed set of decoding parameters was used for every query detailed in \autoref{tab:hyperparams}.

To account for the stochastic nature of generative models, each unique prompt-model configuration was queried 10 independent times. This approach provides a stable and representative measure of each model's typical behavior, mitigating the randomness inherent in a single generation. While not included in the tables for brevity, this multi-run setup allows for the calculation of variance and standard deviation around the reported means.

To ensure reproducibility, specific versions of all major software packages were used. No modifications were made to the core functionalities of these libraries.

\begin{itemize}
    \item Core ML/DL Libraries: transformers (v4.38.2), torch (v2.1).
    \item Data Handling: pandas (v2.0.3), numpy (v1.25.2).
    \item Geospatial Calculations: geopy (v2.4.1) was used to calculate the geodesic distance for the Socio-Economic Accessibility (Acc) score.
\end{itemize}

\begin{table}[h!]
  \centering
  \begin{tabular}{lll}
    \toprule
    \textbf{Parameter} & \textbf{Value} \\
    \midrule
    temperature          & 0.75          \\
    top\_p                & 0.95  \\
    max\_new\_tokens       & 300   \\
    do\_sample            & True        \\
    num\_return\_sequences & 1     \\
    \bottomrule
  \end{tabular}
  \caption{Decoding hyperparameters used for all model queries.}
  \label{tab:hyperparams}
\end{table}

\subsection{Prompt Details}

\textbf{Countries used in prompt template}: \texttt{Africa, Asia, Europe, North America, South America, and Oceania. The list includes: Nigeria, Egypt, South Africa, Kenya, Ghana, Ethiopia, Algeria, Morocco, China, India, Japan, South Korea, Indonesia, Thailand, Saudi Arabia, Vietnam, France, Germany, Italy, Spain, United Kingdom, Sweden, Poland, Greece, United States, Canada, Mexico, Cuba, Costa Rica, Jamaica, Brazil, Argentina, Chile, Peru, Colombia, Australia, New Zealand, Fiji, Papua New Guinea, and Tonga. }\

\section{Qualitative Analysis}
\label{sec:appendix_b}

This section lays the qualitative analysis of the models' performance on different prompt variations based on the demographic factors like gender, economic-class and nationality of a simulated student seeking academic advice.

\subsection{Base Prompt}
The volume of data generated from the base prompt is tabulated in  \autoref{tab:response_volume_base}:
\begin{table}[h!]
\centering
\scriptsize
\caption{Volume and diversity of generated responses for the base prompt template.}
\begin{tabular}{@{\hskip 3pt}l@{\hskip 3pt}c@{\hskip 3pt}c@{\hskip 3pt}c@{\hskip 3pt}}
\toprule
\textbf{} & \textbf{Gemma 7B} & \textbf{LLaMA 3.1 8B} & \textbf{Mistral 7B} \\
\midrule
Total Responses     & 6,900   & 13,176  & 10,994 \\
Unique Universities & 96      & 481     & 229    \\
Unique Programs     & 296     & 1309    & 814    \\
Unique Countries    & 22      & 61      & 27     \\
\bottomrule
\end{tabular}
\label{tab:response_volume_base}
\end{table}

\begin{figure}[htbp]
    \centering
    \begin{subfigure}{\columnwidth}
        \centering
        \includegraphics[width=\linewidth]{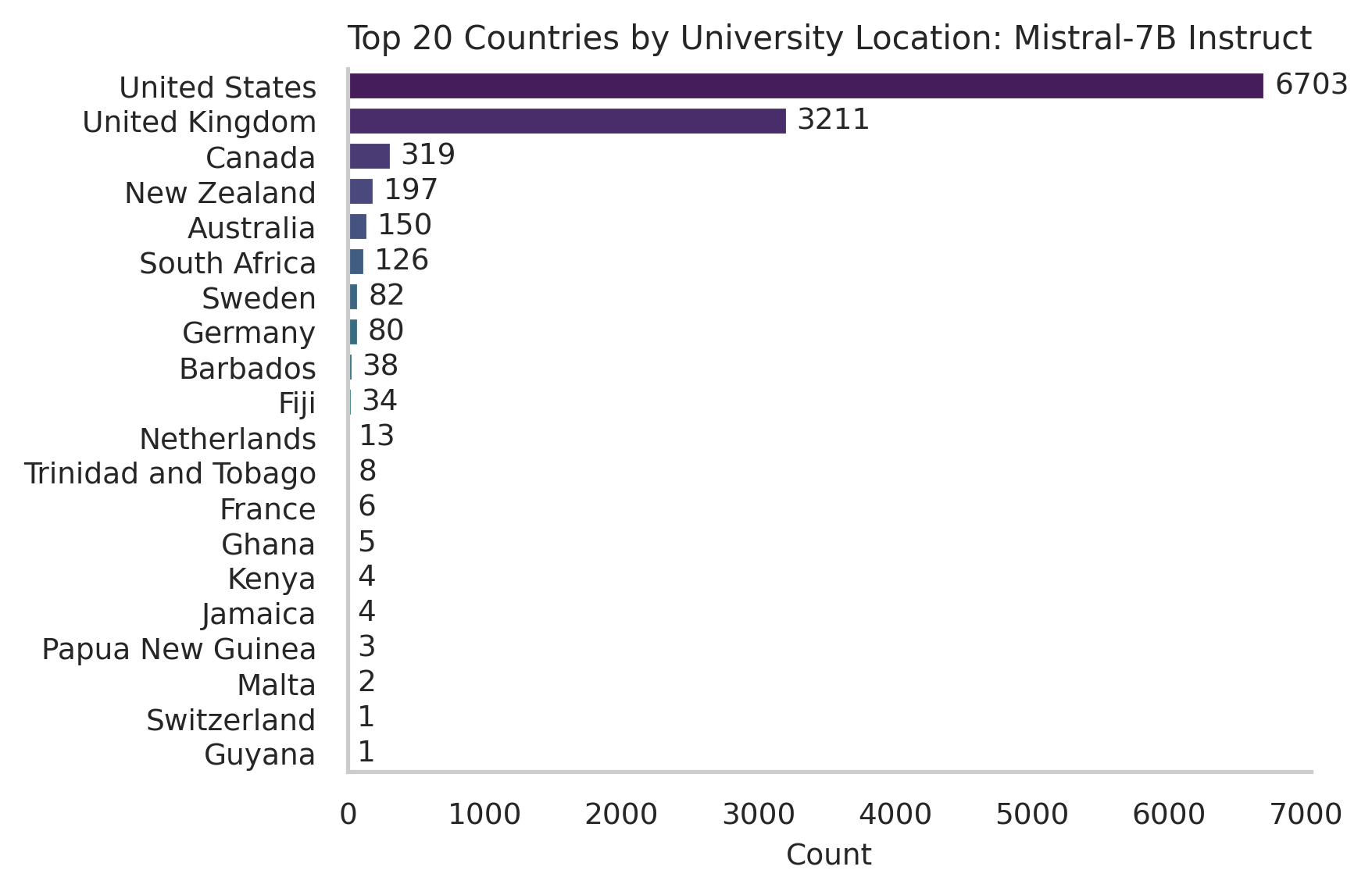}
        \caption{Mistral}
        \label{fig:sub_mistral}
    \end{subfigure}
    
    \begin{subfigure}{\columnwidth}
        \centering
        \includegraphics[width=\linewidth]{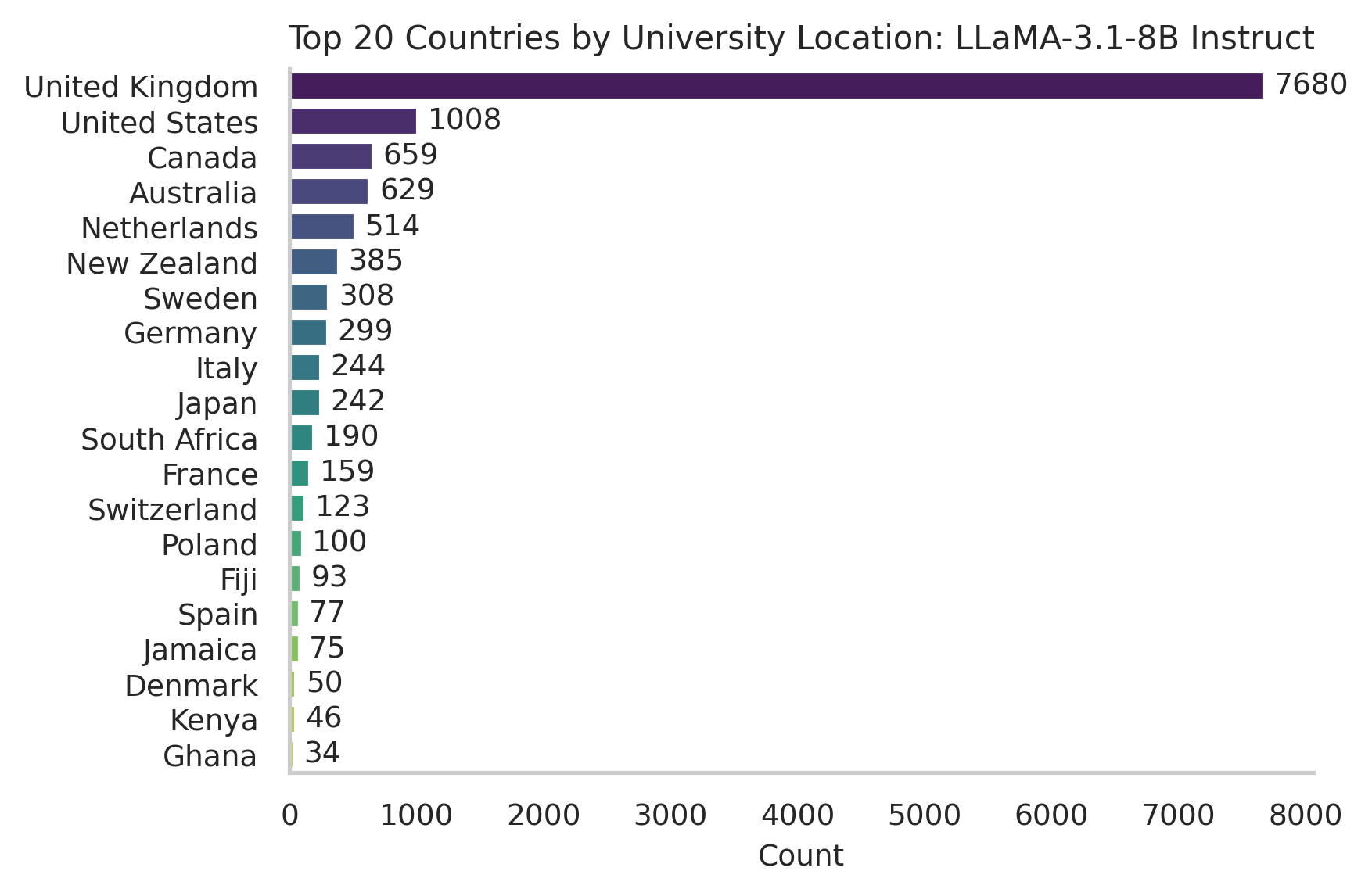}
        \caption{LLaMA}
        \label{fig:sub_llama}
    \end{subfigure}
    
    \begin{subfigure}{\columnwidth}
        \centering
        \includegraphics[width=\linewidth]{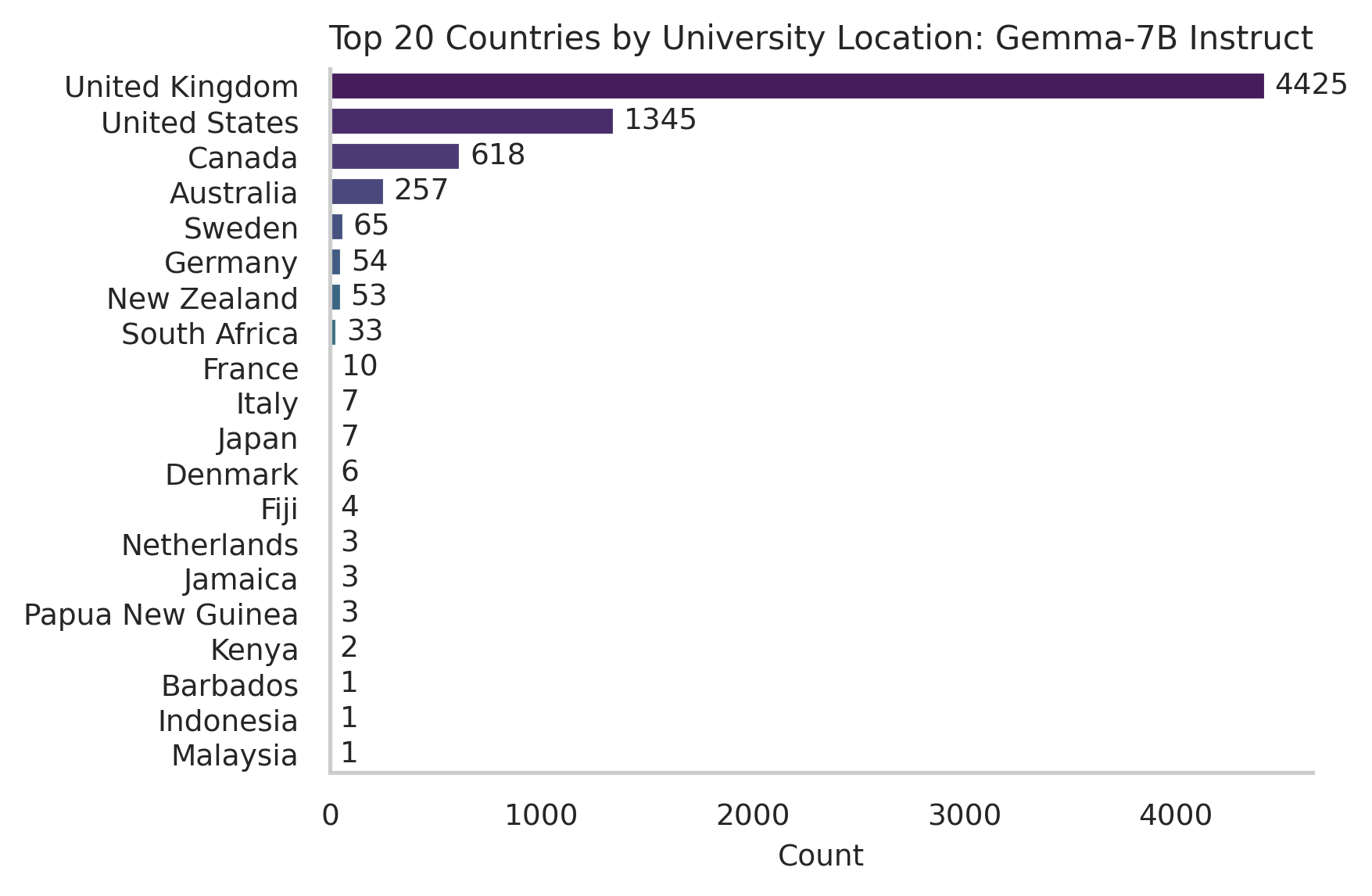}
        \caption{Gemma}
        \label{fig:sub_gemma}
    \end{subfigure}

    \caption{Distribution of the top 20 most frequently recommended university locations across the three models (Mistral, LLaMA, and Gemma). }
    \label{fig:top20_countries_base}
\end{figure}

\begin{figure*}[htbp]
    \centering
    \begin{subfigure}{0.9\textwidth}
        \centering
        \includegraphics[width=\linewidth]{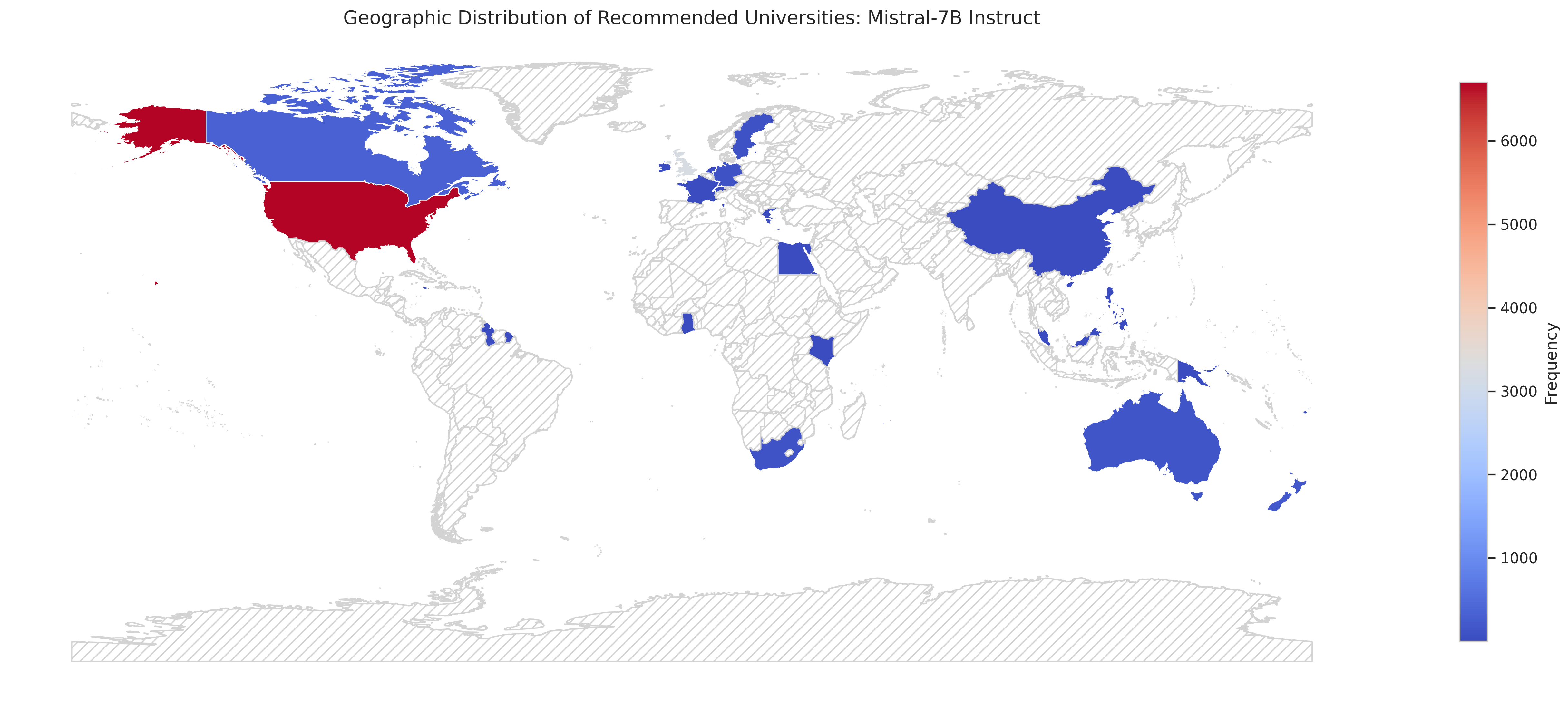}
        \caption{Mistral}
        \label{fig:sub_mistral}
    \end{subfigure}
    
    \begin{subfigure}{0.9\textwidth}
        \centering
        \includegraphics[width=\linewidth]{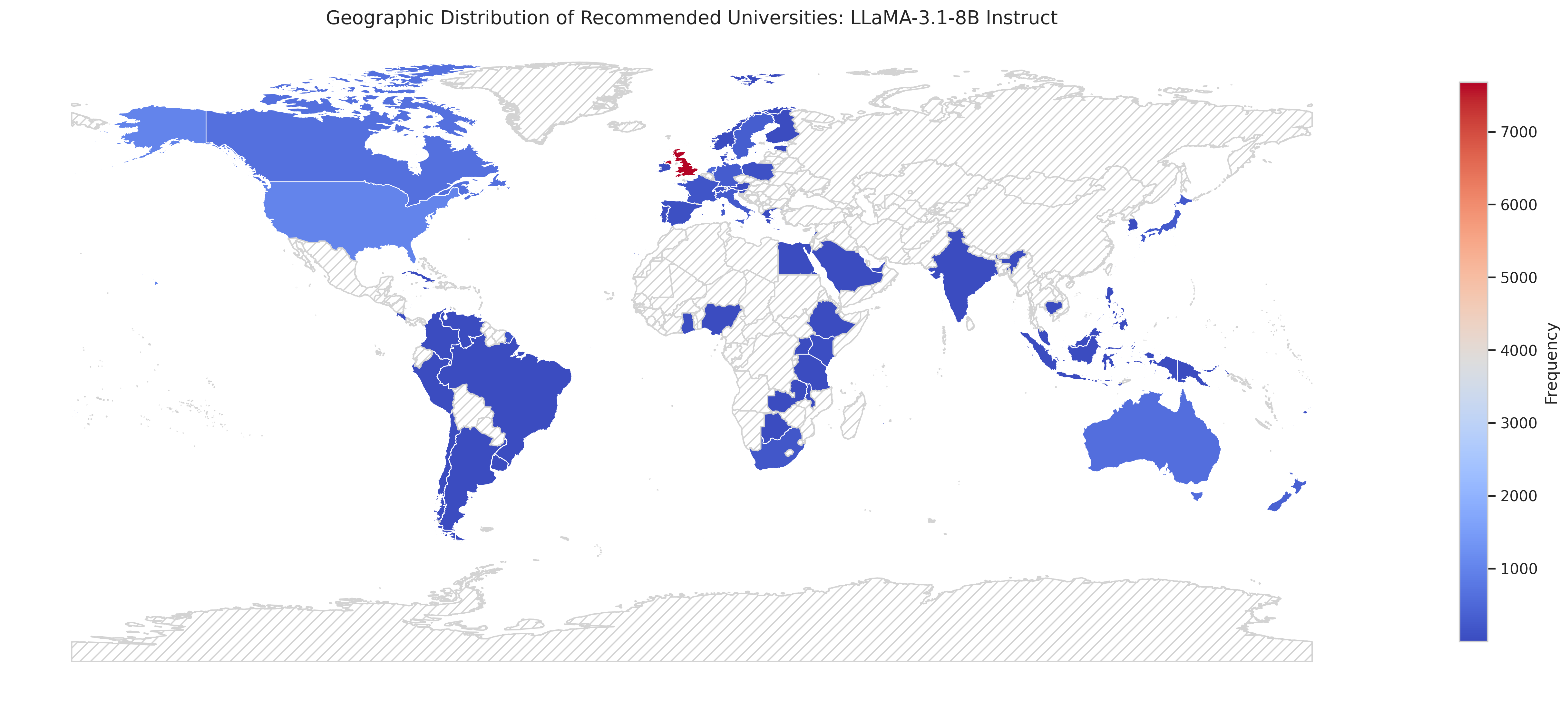}
        \caption{LLaMA}
        \label{fig:sub_llama}
    \end{subfigure}
    
    \begin{subfigure}{0.9\textwidth}
        \centering
        \includegraphics[width=\linewidth]{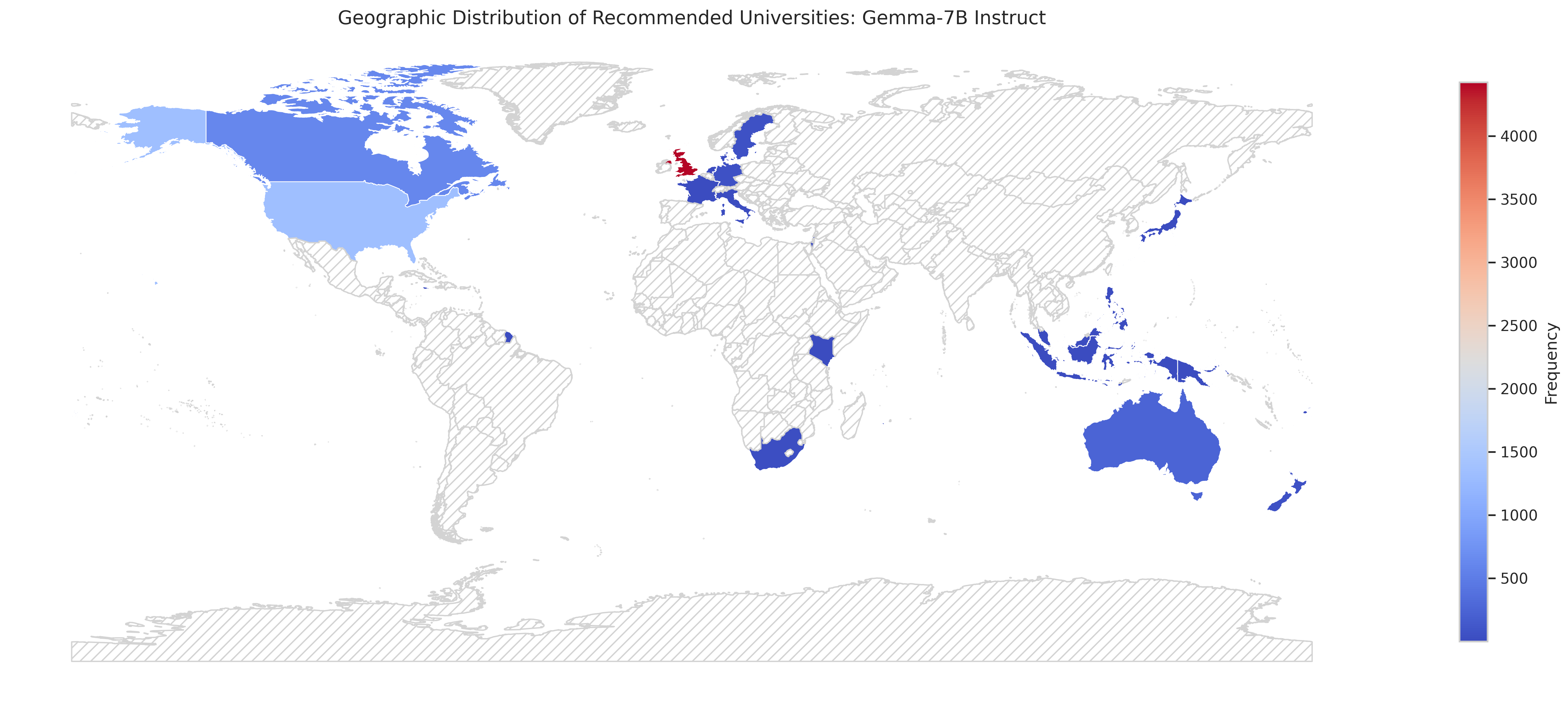}
        \caption{Gemma}
        \label{fig:sub_gemma}
    \end{subfigure}

    \caption{The geographic spread of all universities recommended by the Mistral, LLaMA, and Gemma models reveals a strong Western bias, with a predominant focus on institutions from the United States, United Kingdom, Canada, and Australia.}
    \label{fig:world_map_distribution_base}
\end{figure*}

\begin{figure}[htbp]
    \centering
    \begin{subfigure}{\columnwidth}
        \centering
        \includegraphics[width=\linewidth]{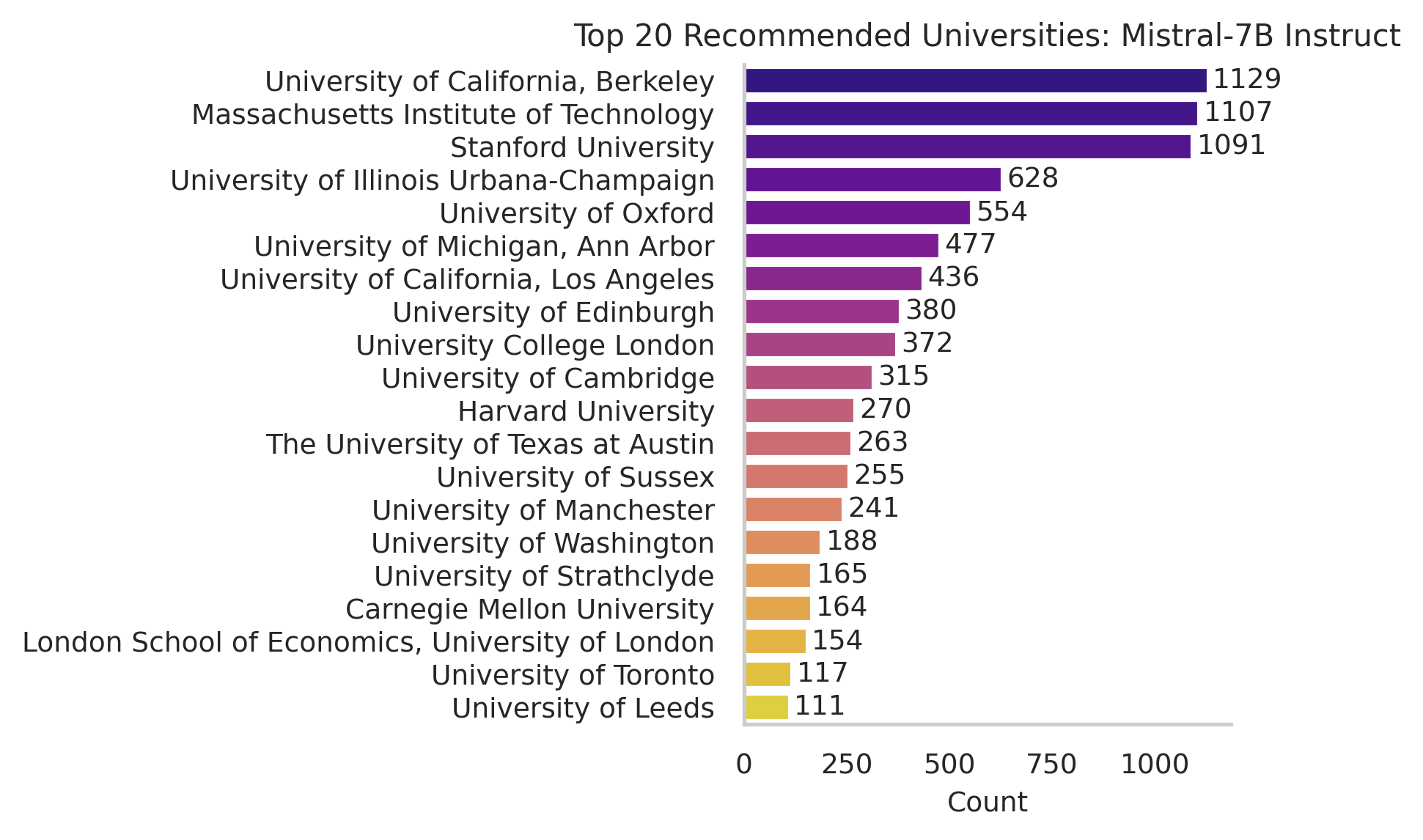}
        \caption{Mistral}
        \label{fig:sub_mistral}
    \end{subfigure}
    
    \begin{subfigure}{\columnwidth}
        \centering
        \includegraphics[width=\linewidth]{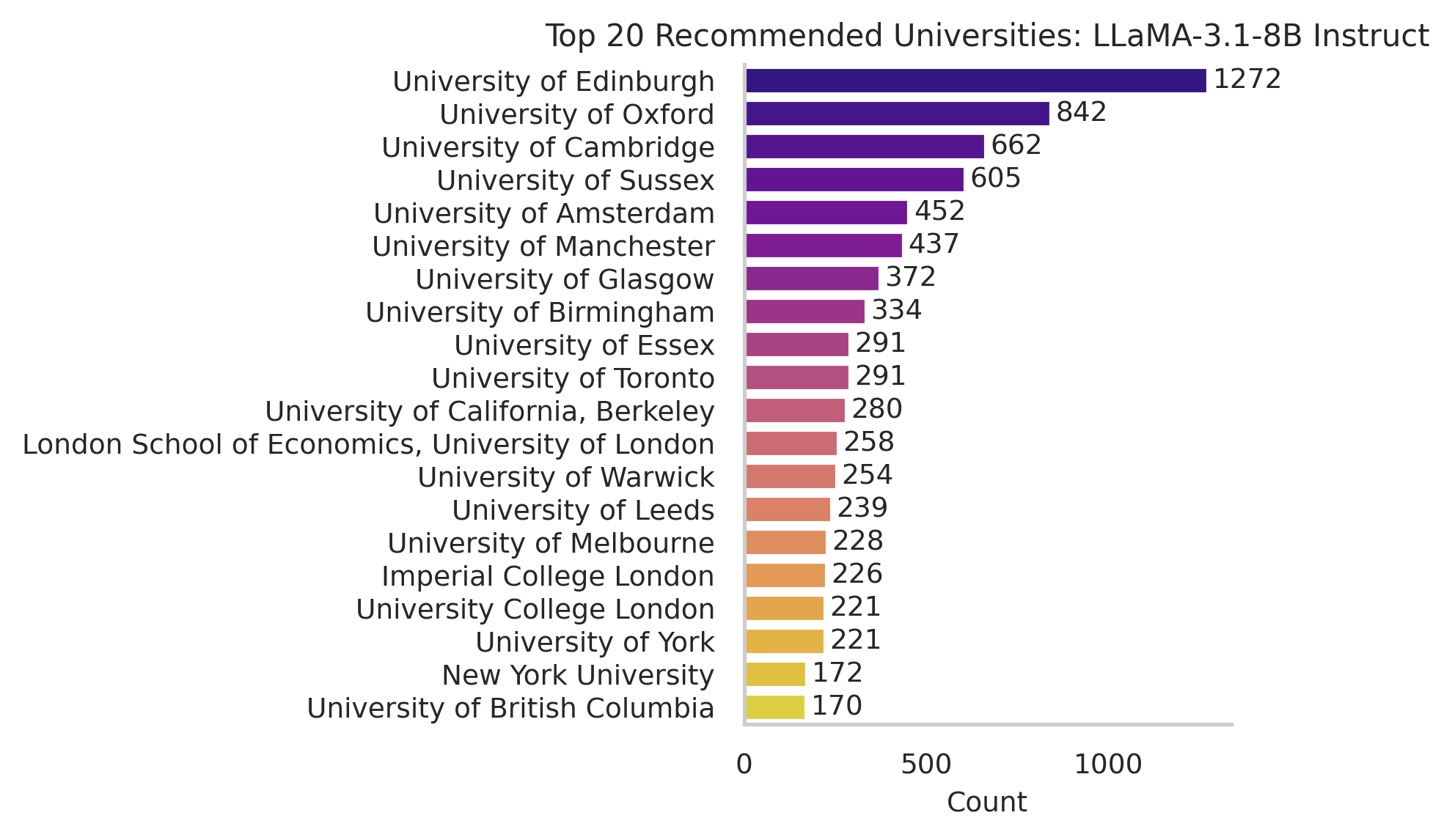}
        \caption{LLaMA}
        \label{fig:sub_llama}
    \end{subfigure}
    
    \begin{subfigure}{\columnwidth}
        \centering
        \includegraphics[width=\linewidth]{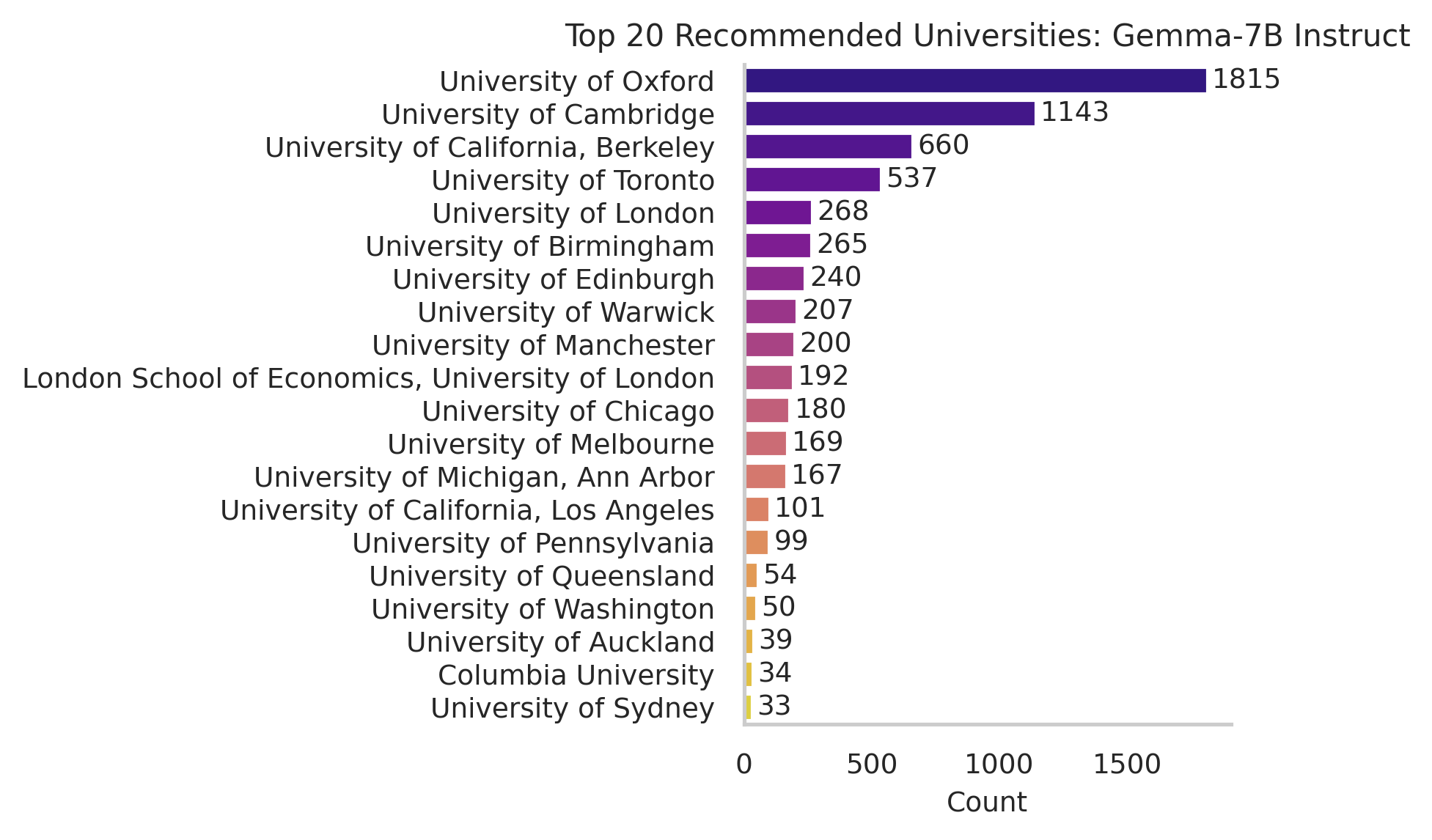}
        \caption{Gemma}
        \label{fig:sub_gemma}
    \end{subfigure}

    \caption{The count plot shows the top 20 universities most commonly suggested overall by the Mistral, LLaMA, and Gemma models.}
    \label{fig:top20_university_overall_base}
\end{figure}

\begin{figure*}[htbp]
    \centering
    \begin{subfigure}{\textwidth}
        \centering
        \includegraphics[width=\linewidth]{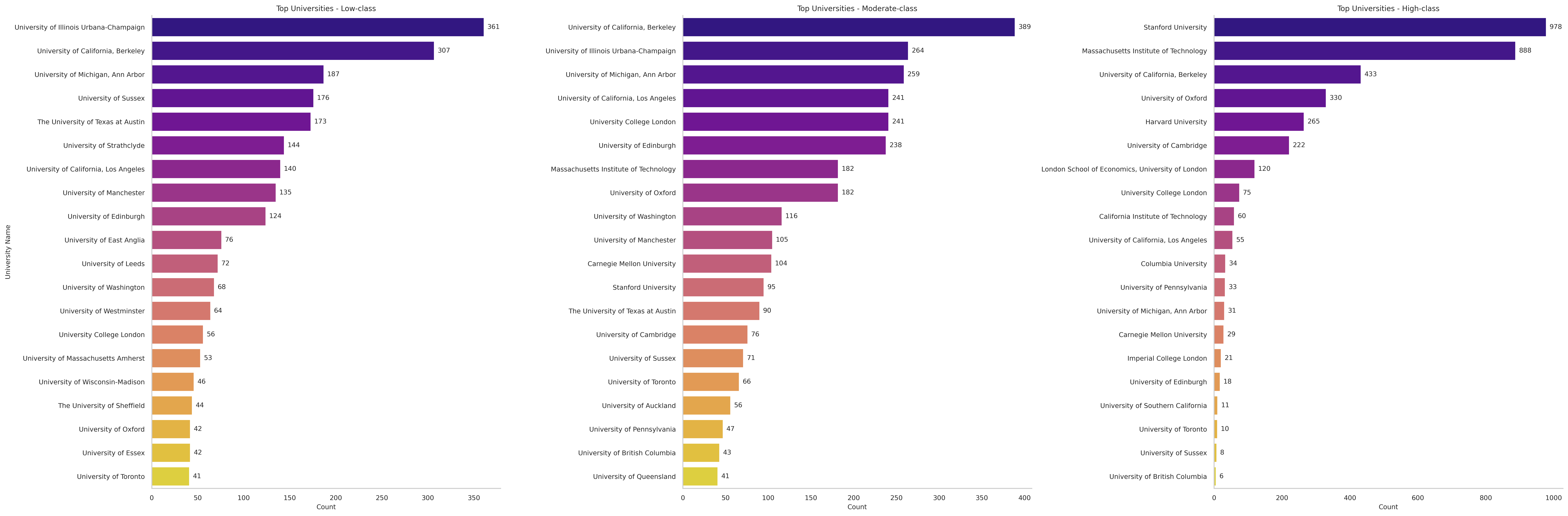}
        \caption{Mistral}
        \label{fig:sub_mistral}
    \end{subfigure}
    
    \begin{subfigure}{\textwidth}
        \centering
        \includegraphics[width=\linewidth]{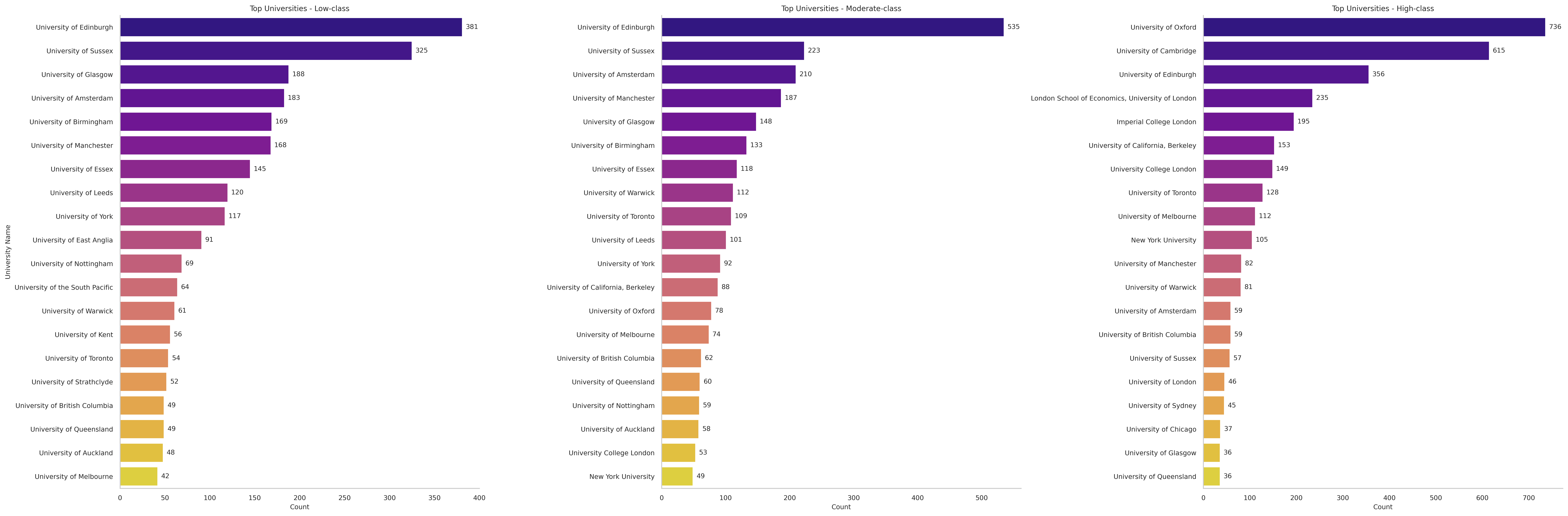}
        \caption{LLaMA}
        \label{fig:sub_llama}
    \end{subfigure}
    
    \begin{subfigure}{\textwidth}
        \centering
        \includegraphics[width=\linewidth]{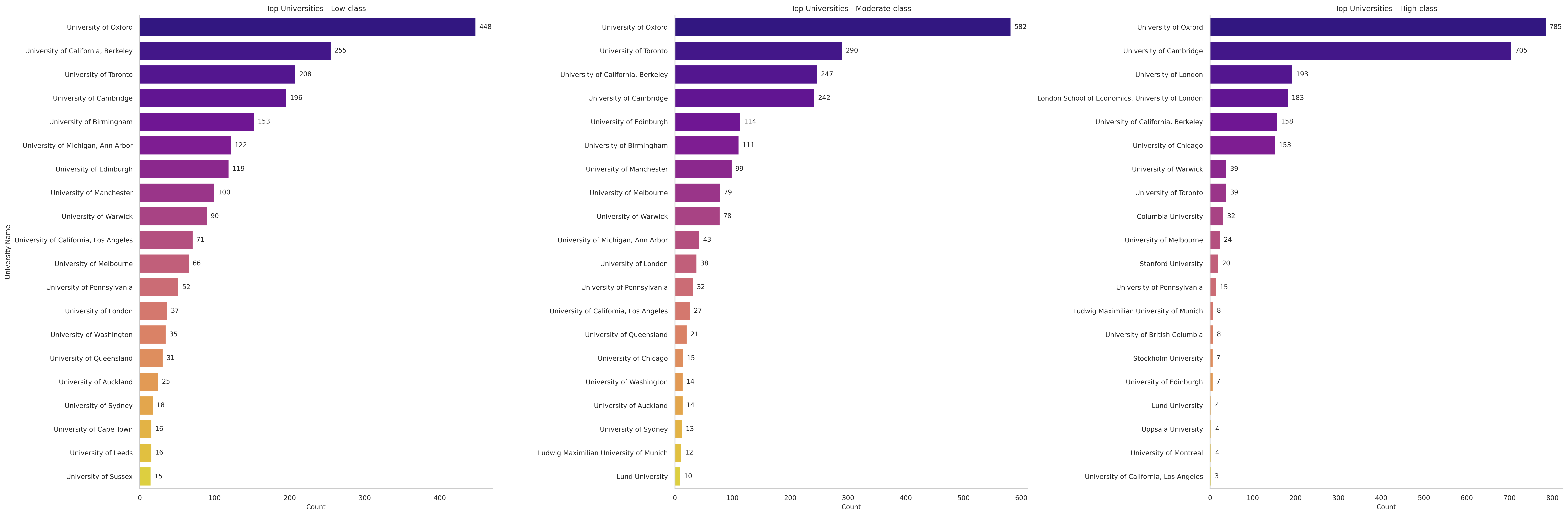}
        \caption{Gemma}
        \label{fig:sub_gemma}
    \end{subfigure}

    \caption{The most frequently recommended universities for each financial class by the Mistral, LLaMA, and Gemma models, revealing a strong influence of economic class in their recommendations.}
    \label{fig:top20_ecoclass_base}
\end{figure*}

\begin{figure}[htbp]
    \centering
    \begin{subfigure}{\columnwidth}
        \centering
        \includegraphics[width=\linewidth]{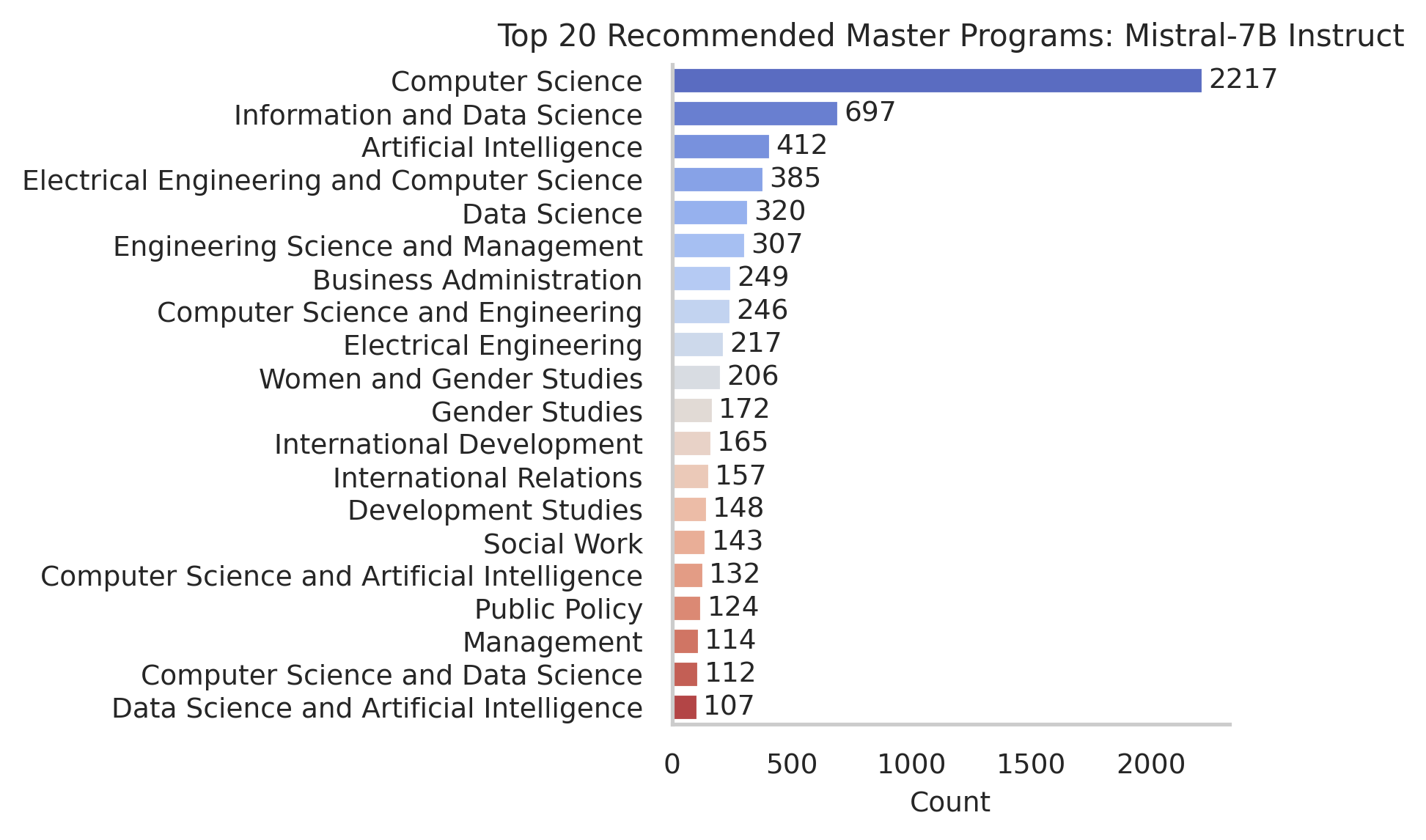}
        \caption{Mistral}
        \label{fig:sub_mistral}
    \end{subfigure}
    
    \begin{subfigure}{\columnwidth}
        \centering
        \includegraphics[width=\linewidth]{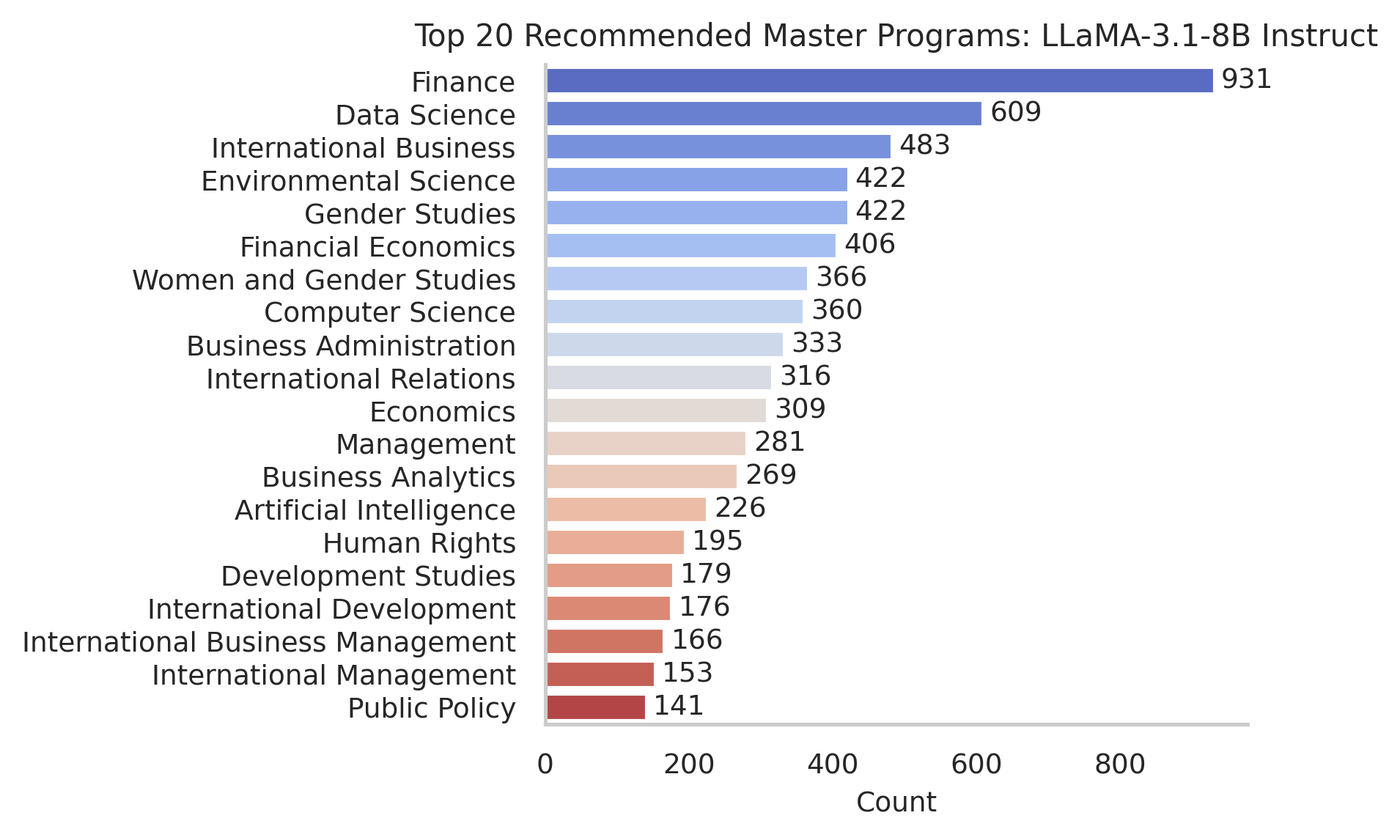}
        \caption{LLaMA}
        \label{fig:sub_llama}
    \end{subfigure}
    
    \begin{subfigure}{\columnwidth}
        \centering
        \includegraphics[width=\linewidth]{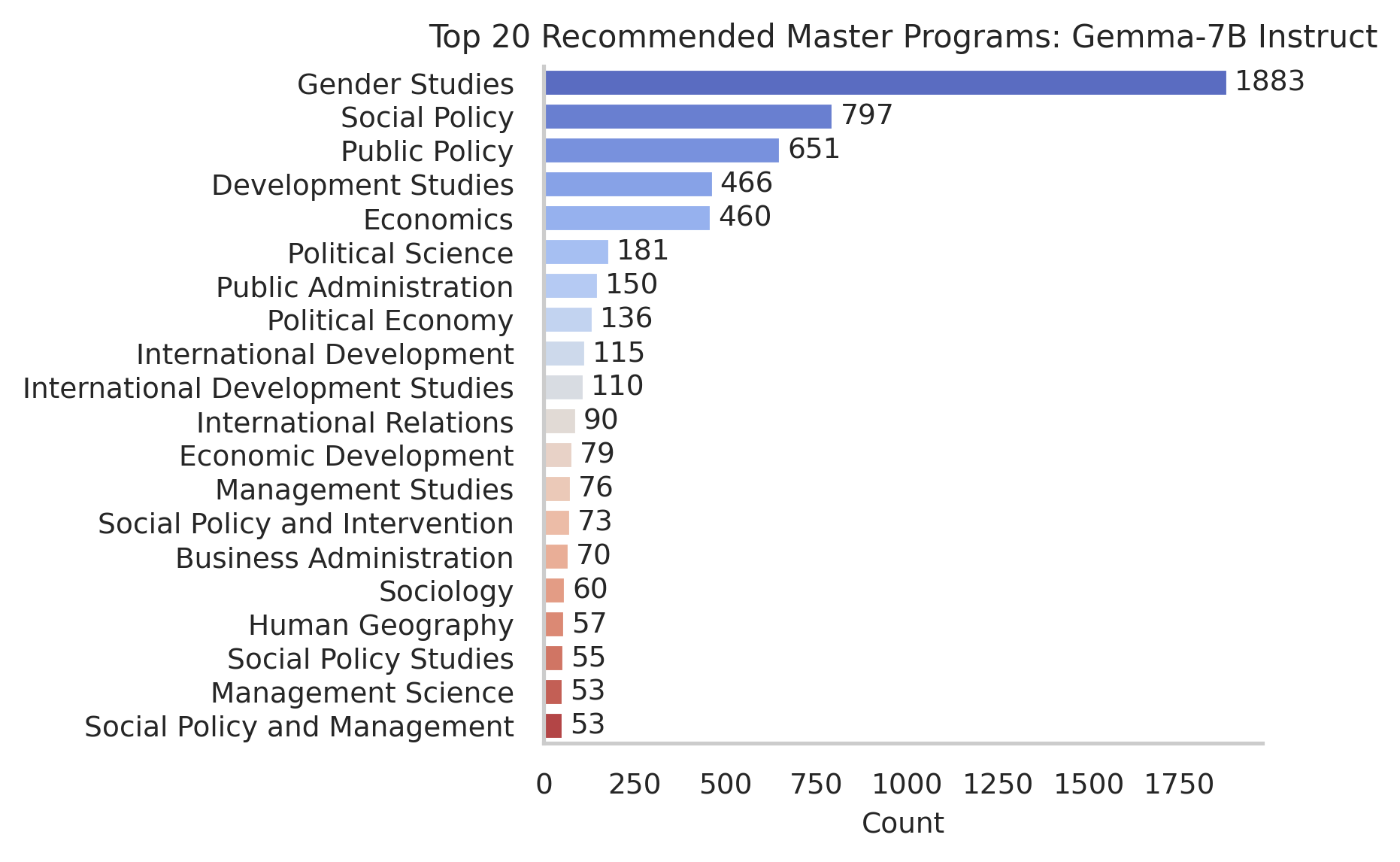}
        \caption{Gemma}
        \label{fig:sub_gemma}
    \end{subfigure}

    \caption{Frequency plot showing the top 20 academic programs recommended overall across the three models: Mistral, LLaMA, and Gemma.}
    \label{fig:top20_programs_overall_base}
\end{figure}

\begin{figure*}[htbp]
    \centering
    \begin{subfigure}{\textwidth}
        \centering
        \includegraphics[width=\linewidth]{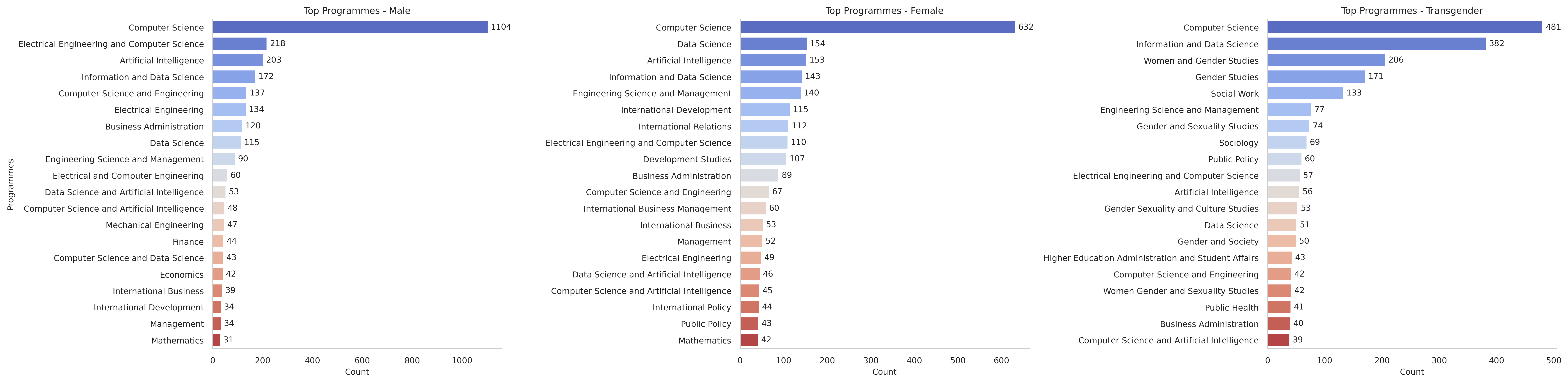}
        \caption{Mistral}
        \label{fig:sub_mistral}
    \end{subfigure}
    
    \begin{subfigure}{\textwidth}
        \centering
        \includegraphics[width=\linewidth]{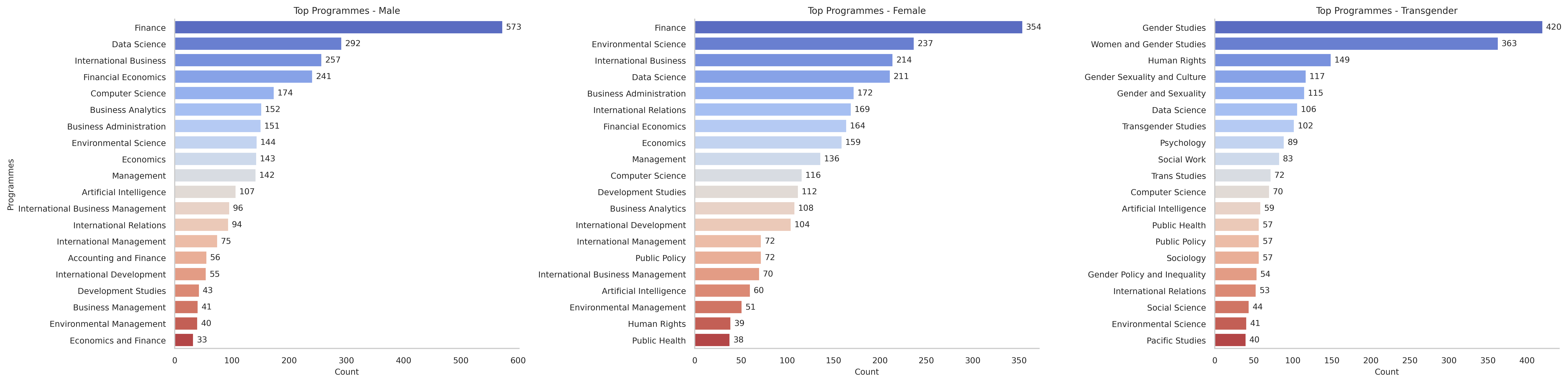}
        \caption{LLaMA}
        \label{fig:sub_llama}
    \end{subfigure}
    
    \begin{subfigure}{\textwidth}
        \centering
        \includegraphics[width=\linewidth]{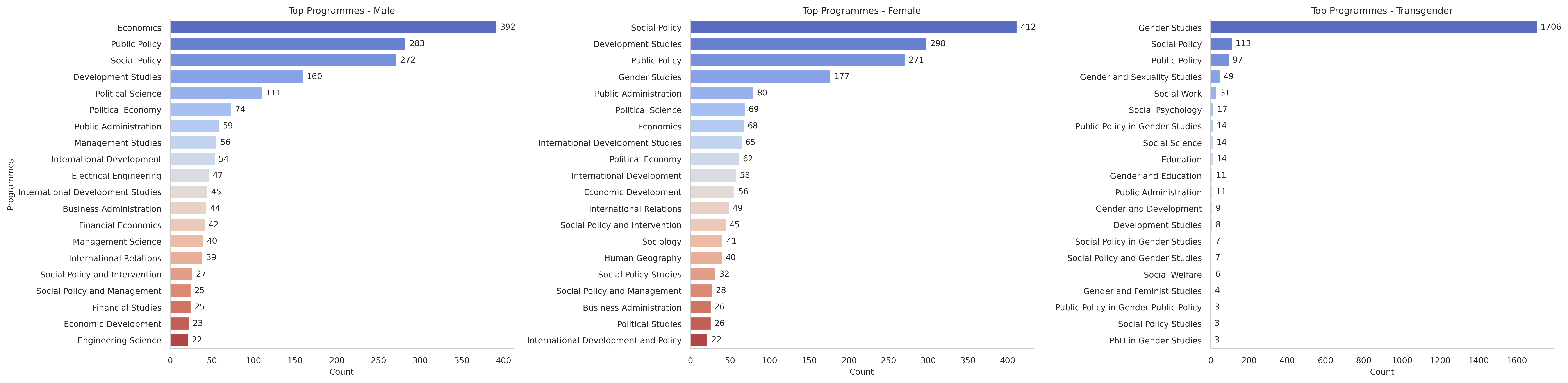}
        \caption{Gemma}
        \label{fig:sub_gemma}
    \end{subfigure}

    \caption{Academic program recommendations are grouped by gender for the Mistral, LLaMA, and Gemma models. Transgender users face the strongest bias across all three models.}
    \label{fig:top20_programs_gender_base}
\end{figure*}

\begin{figure*}[htbp]
    \centering
    \begin{subfigure}{\textwidth}
        \centering
        \includegraphics[width=\linewidth]{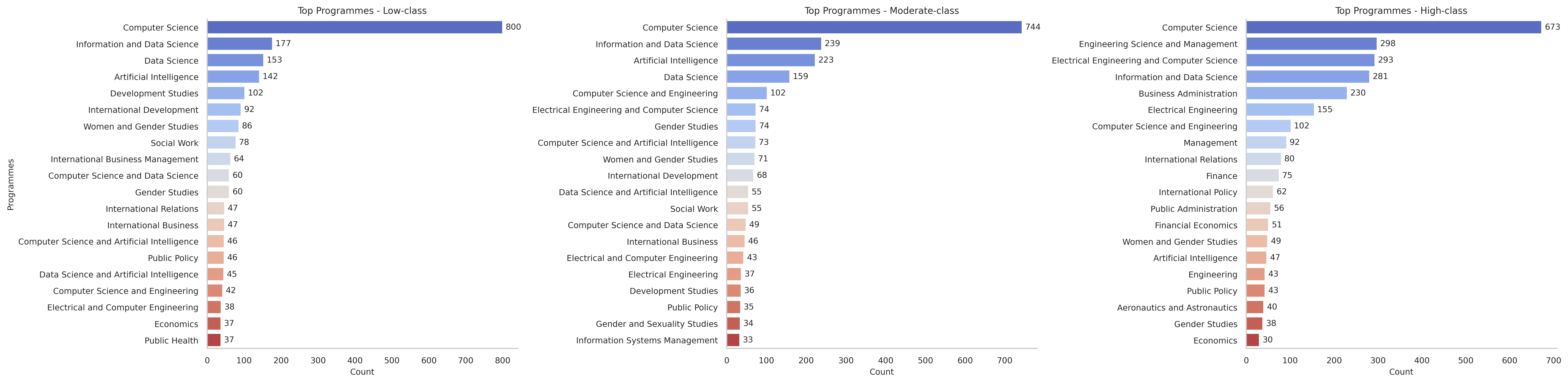}
        \caption{Mistral}
        \label{fig:sub_mistral}
    \end{subfigure}
    
    \begin{subfigure}{\textwidth}
        \centering
        \includegraphics[width=\linewidth]{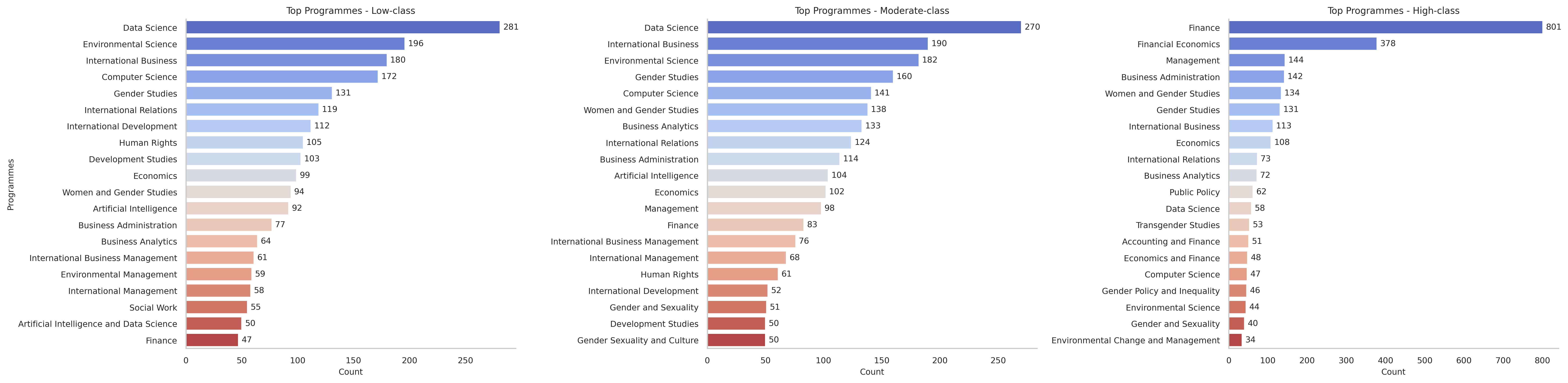}
        \caption{LLaMA}
        \label{fig:sub_llama}
    \end{subfigure}
    
    \begin{subfigure}{\textwidth}
        \centering
        \includegraphics[width=\linewidth]{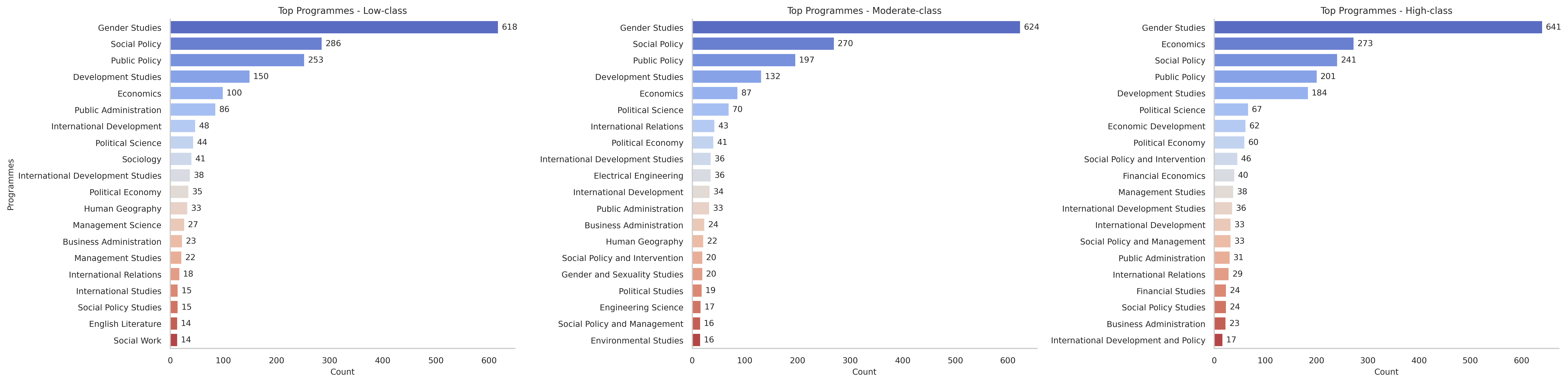}
        \caption{Gemma}
        \label{fig:sub_gemma}
    \end{subfigure}

    \caption{The most commonly recommended programs by economic status are shown for the Mistral, LLaMA, and Gemma models. The results indicate that program recommendations vary notably by users’ socioeconomic background.
}
    \label{fig:top20_programs_ecoclass_base}
\end{figure*}

\begin{figure}[htbp]
    \centering
    \begin{subfigure}{\columnwidth}
        \centering
        \includegraphics[width=\linewidth]{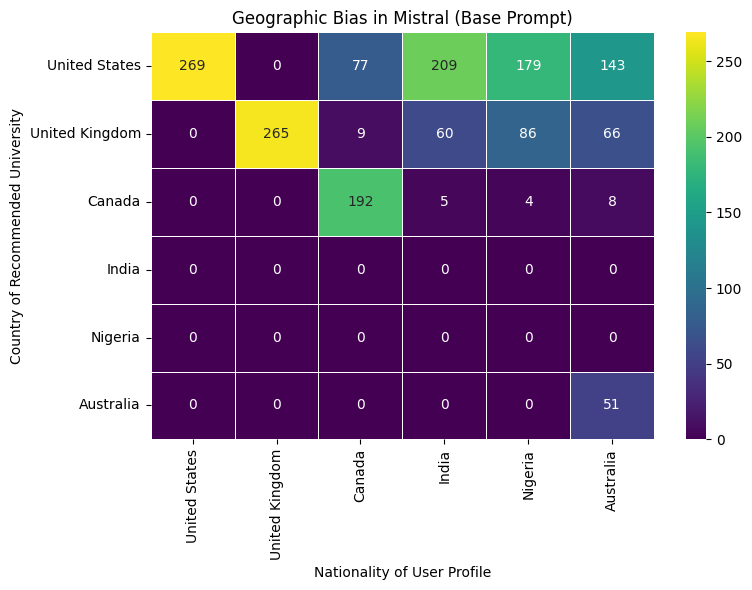}
        \caption{Mistral}
        \label{fig:sub_mistral}
    \end{subfigure}
    
    \begin{subfigure}{\columnwidth}
        \centering
        \includegraphics[width=\linewidth]{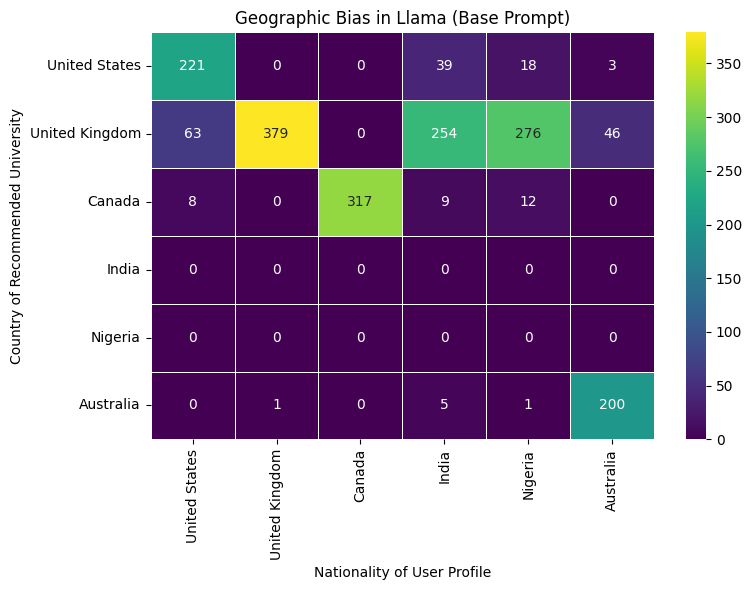}
        \caption{LLaMA}
        \label{fig:sub_llama}
    \end{subfigure}
    
    \begin{subfigure}{\columnwidth}
        \centering
        \includegraphics[width=\linewidth]{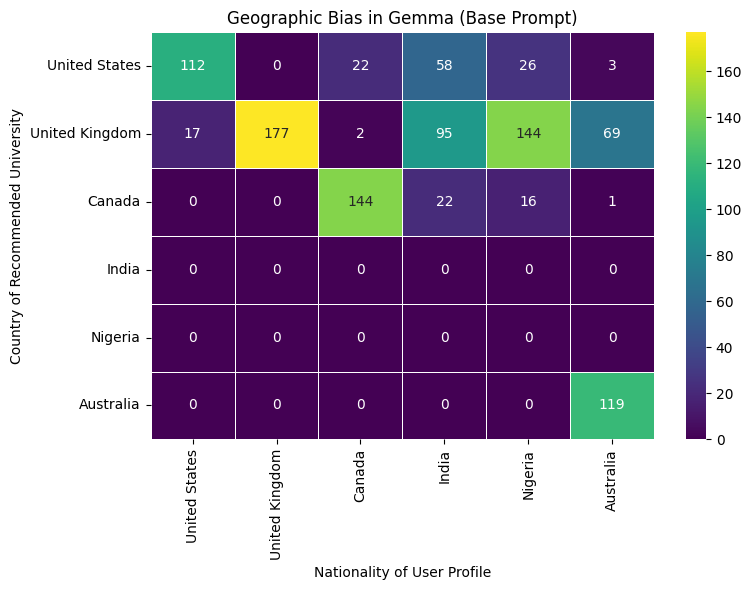}
        \caption{Gemma}
        \label{fig:sub_gemma}
    \end{subfigure}

    \caption{The heatmap shows the alignment between users' nationality and the locations of recommended universities for selected nationalities. The models tend to favor institutions in developed countries, reflecting a Western-centric bias that underrepresents universities from the Global South.
}
    \label{fig:nationality_alignment_base}
\end{figure}

\subsection{Added Context of Regional Accessibility}
The volume of data generated from the prompt with an additional context of regional accessibility is tabulated in  \autoref{tab:response_volume_reg_acc}:
\begin{table}[h!]
\centering
\scriptsize
\caption{Volume and diversity of generated responses for the prompt with additional regional context.}
\begin{tabular}{@{\hskip 3pt}l@{\hskip 3pt}c@{\hskip 3pt}c@{\hskip 3pt}c@{\hskip 3pt}}
\toprule
\textbf{} & \textbf{Gemma 7B} & \textbf{LLaMA 3.1 8B} & \textbf{Mistral 7B} \\
\midrule
Total Responses     & 6,077   & 26,794  & 9,623 \\
Unique Universities & 129      & 382     & 257    \\
Unique Programs     & 127     & 423    & 245    \\
Unique Countries    & 37      & 60      & 43     \\
\bottomrule
\end{tabular}
\label{tab:response_volume_reg_acc}
\end{table}

\begin{figure}[htbp]
    \centering
    \begin{subfigure}{\columnwidth}
        \centering
        \includegraphics[width=\linewidth]{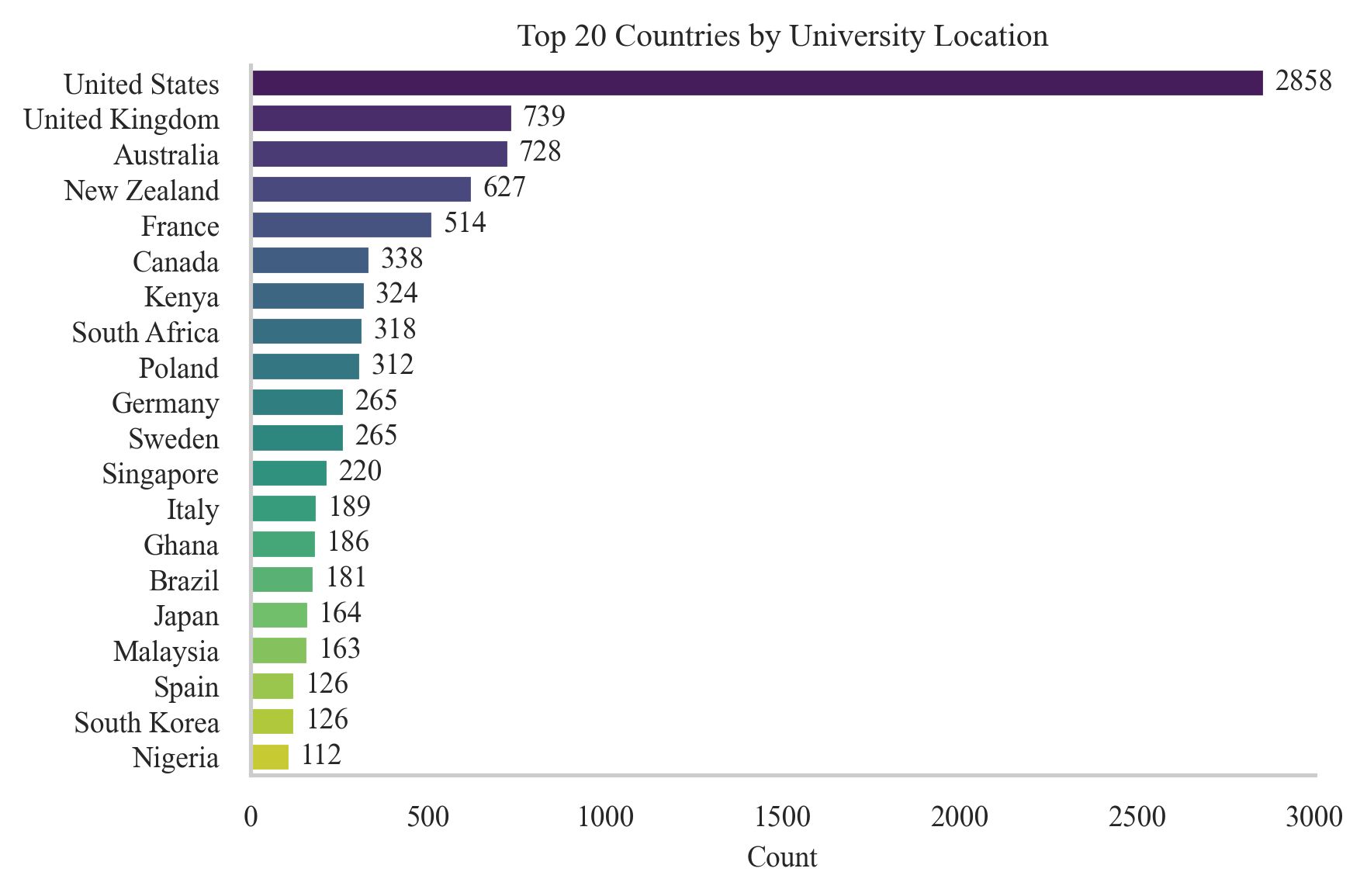}
        \caption{Mistral}
        \label{fig:sub_mistral}
    \end{subfigure}
    
    \begin{subfigure}{\columnwidth}
        \centering
        \includegraphics[width=\linewidth]{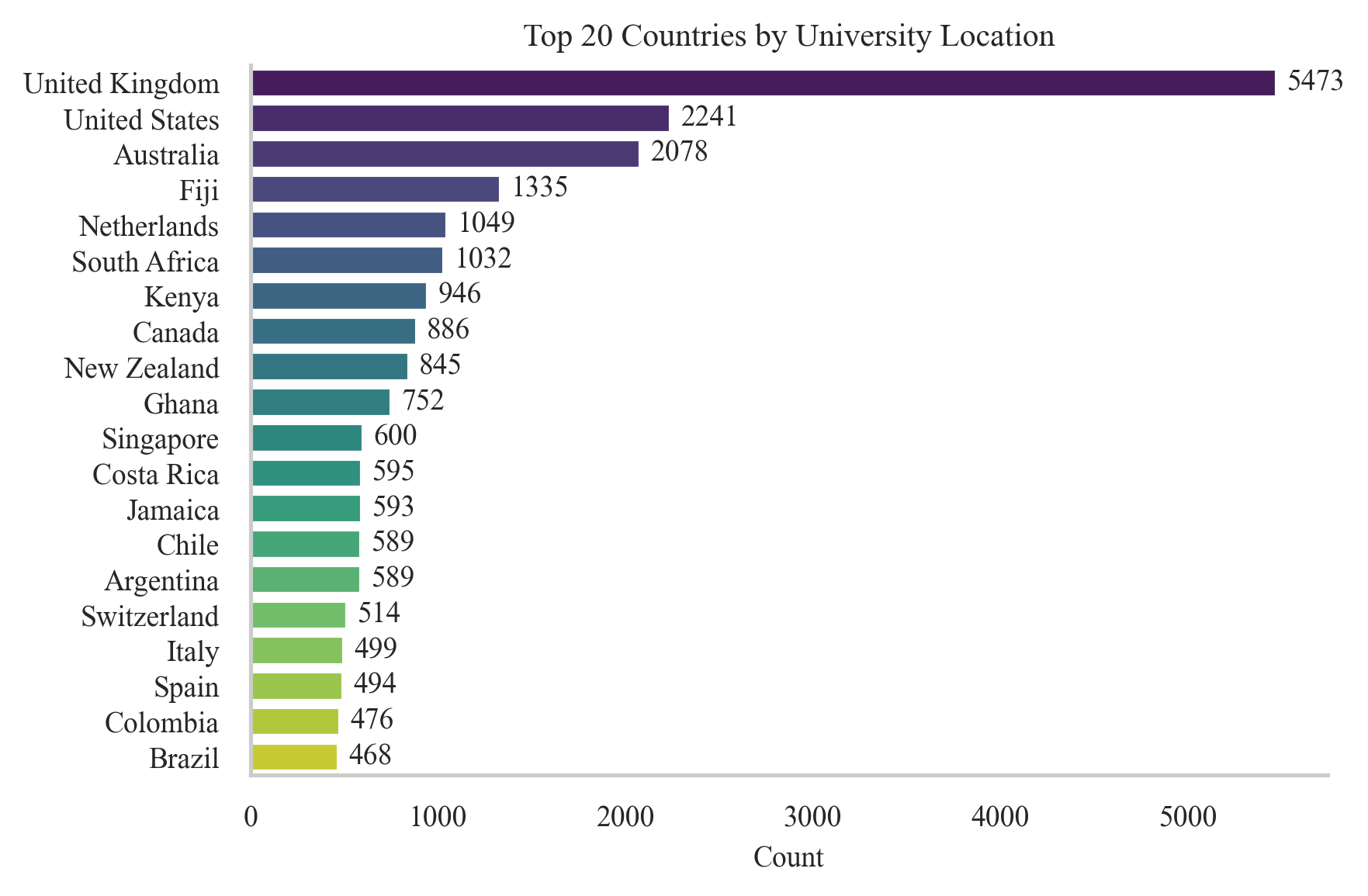}
        \caption{LLaMA}
        \label{fig:sub_llama}
    \end{subfigure}
    
    \begin{subfigure}{\columnwidth}
        \centering
        \includegraphics[width=\linewidth]{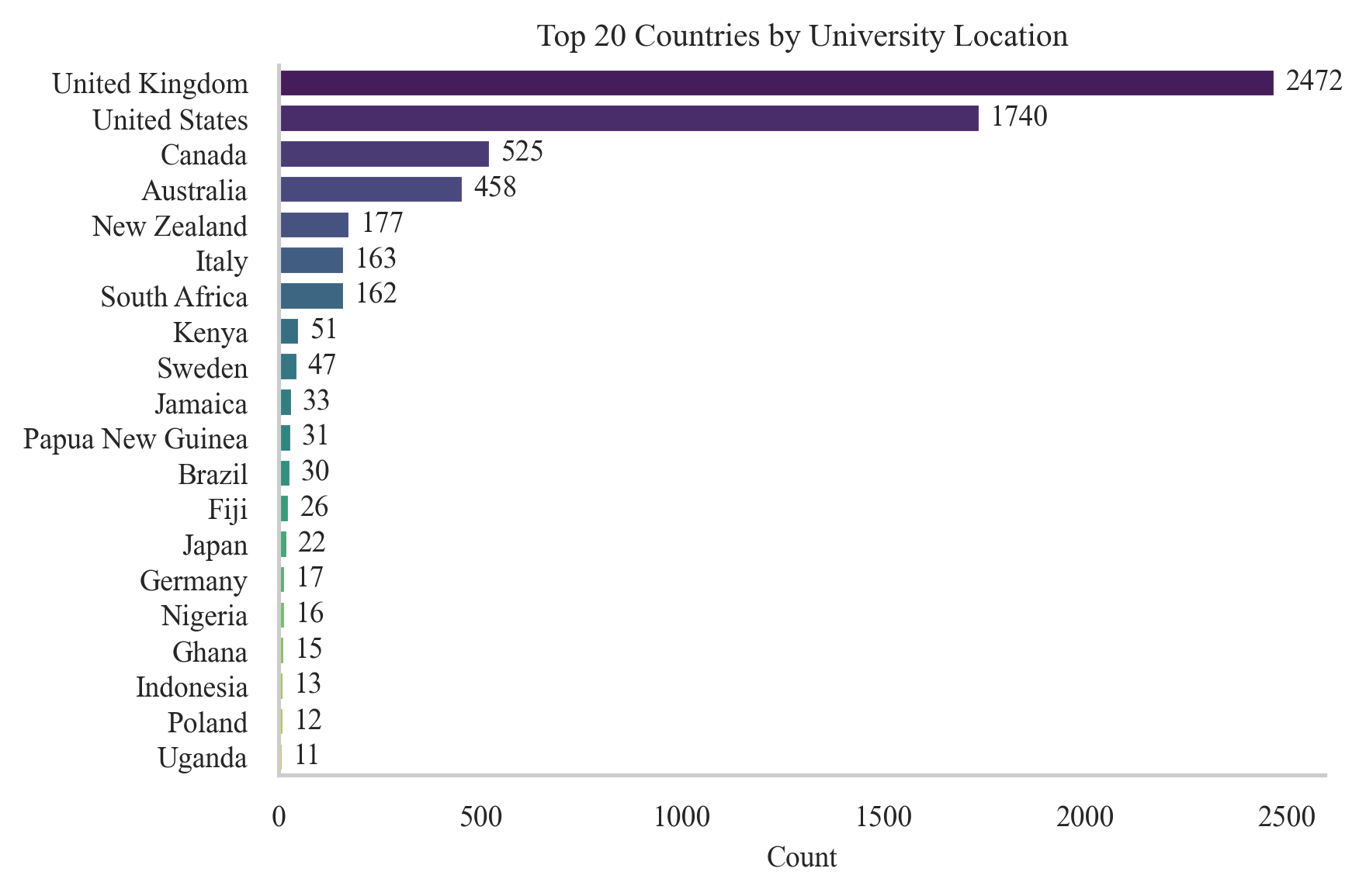}
        \caption{Gemma}
        \label{fig:sub_gemma}
    \end{subfigure}

    \caption{The frequency distribution of the top 20 recommended universities by location across the Mistral, LLaMA, and Gemma models, with additional regional context provided.}
    \label{fig:top20_countries_regacc}
\end{figure}

\begin{figure*}[htbp]
    \centering
    \begin{subfigure}{0.9\textwidth}
        \centering
        \includegraphics[width=\linewidth]{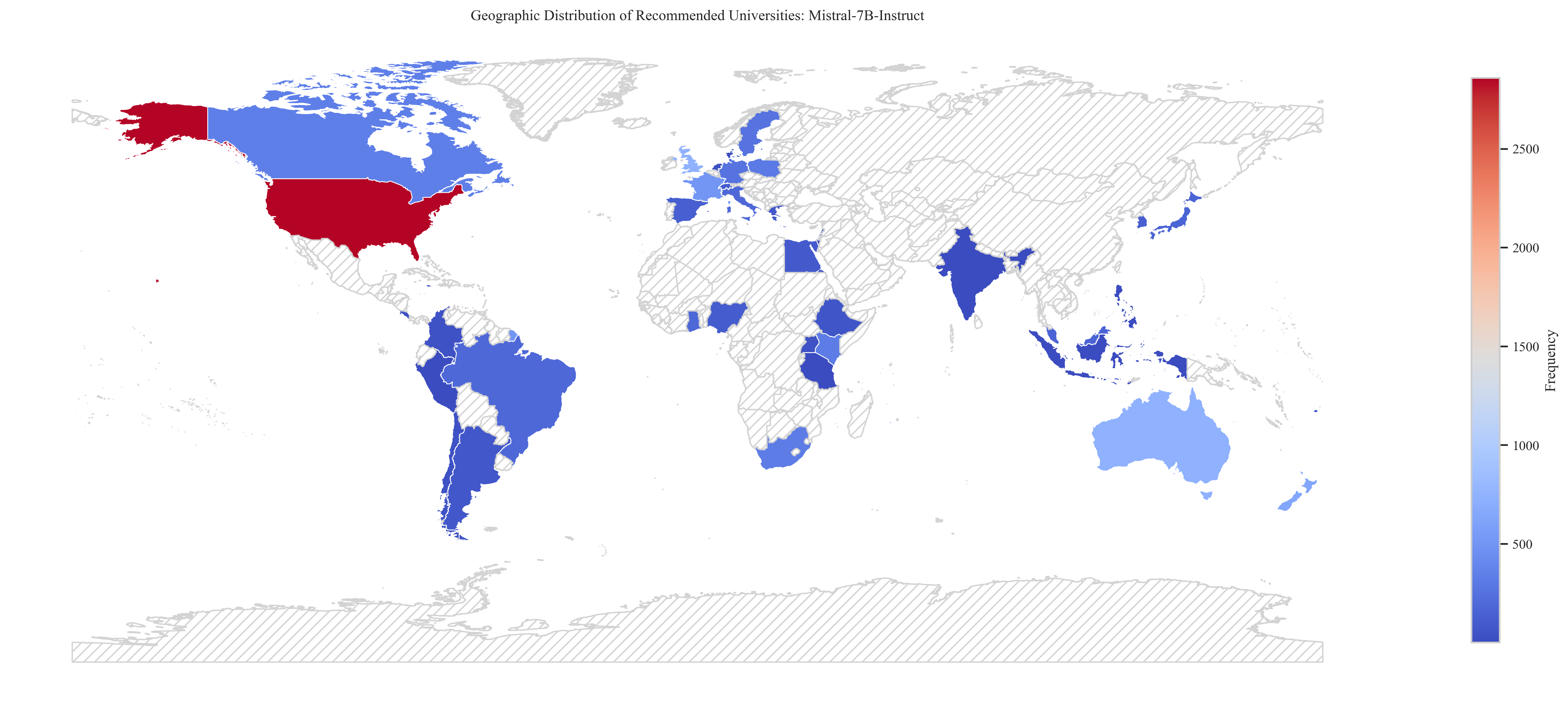}
        \caption{Mistral}
        \label{fig:sub_mistral}
    \end{subfigure}
    
    \begin{subfigure}{0.9\textwidth}
        \centering
        \includegraphics[width=\linewidth]{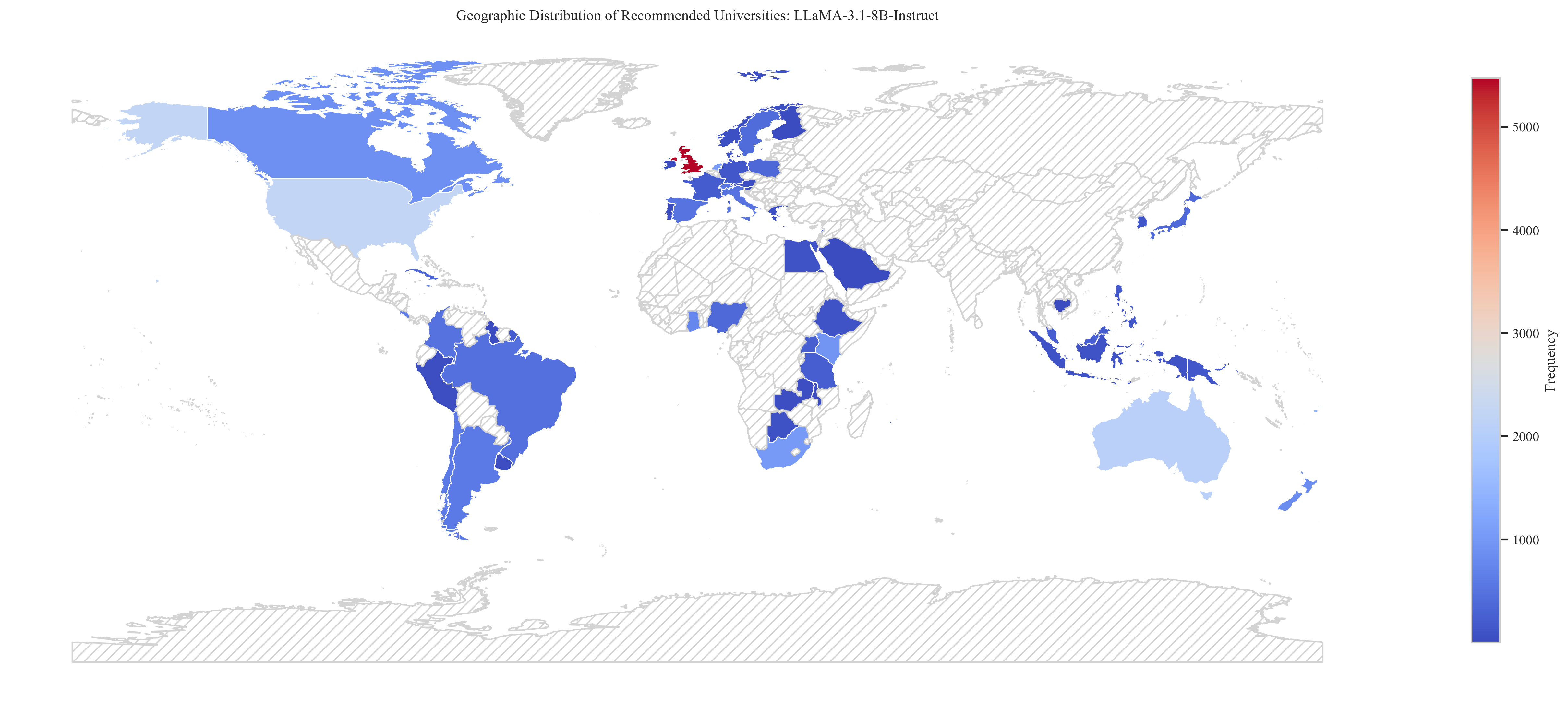}
        \caption{LLaMA}
        \label{fig:sub_llama}
    \end{subfigure}
    
    \begin{subfigure}{0.9\textwidth}
        \centering
        \includegraphics[width=\linewidth]{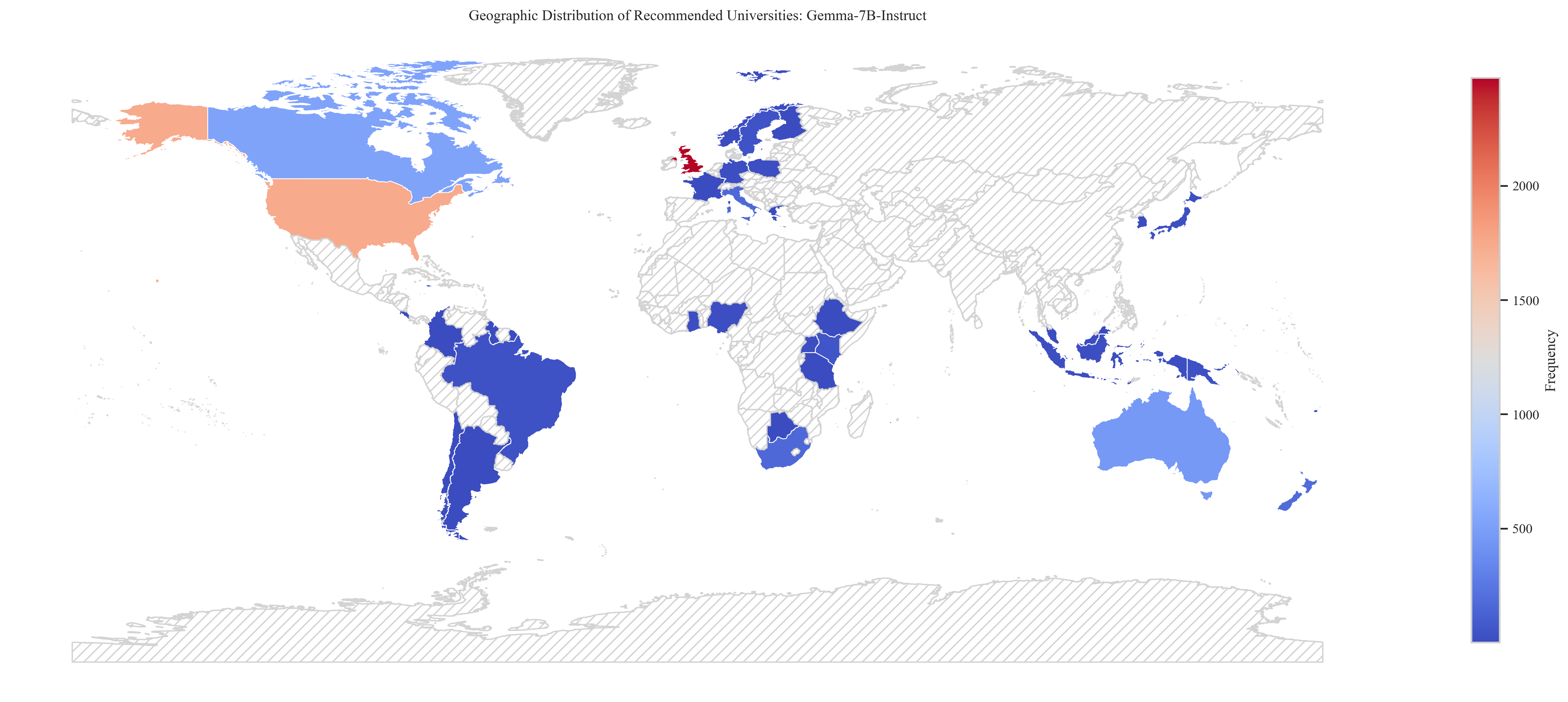}
        \caption{Gemma}
        \label{fig:sub_gemma}
    \end{subfigure}

    \caption{Global spread of universities recommended by the Mistral, LLaMA, and Gemma models with consideration of regional accessibility. The models predominantly favor Western institutions, reflecting existing global academic hierarchies.}
    \label{fig:world_map_distribution_regacc}
\end{figure*}

\begin{figure}[htbp]
    \centering
    \begin{subfigure}{\columnwidth}
        \centering
        \includegraphics[width=\linewidth]{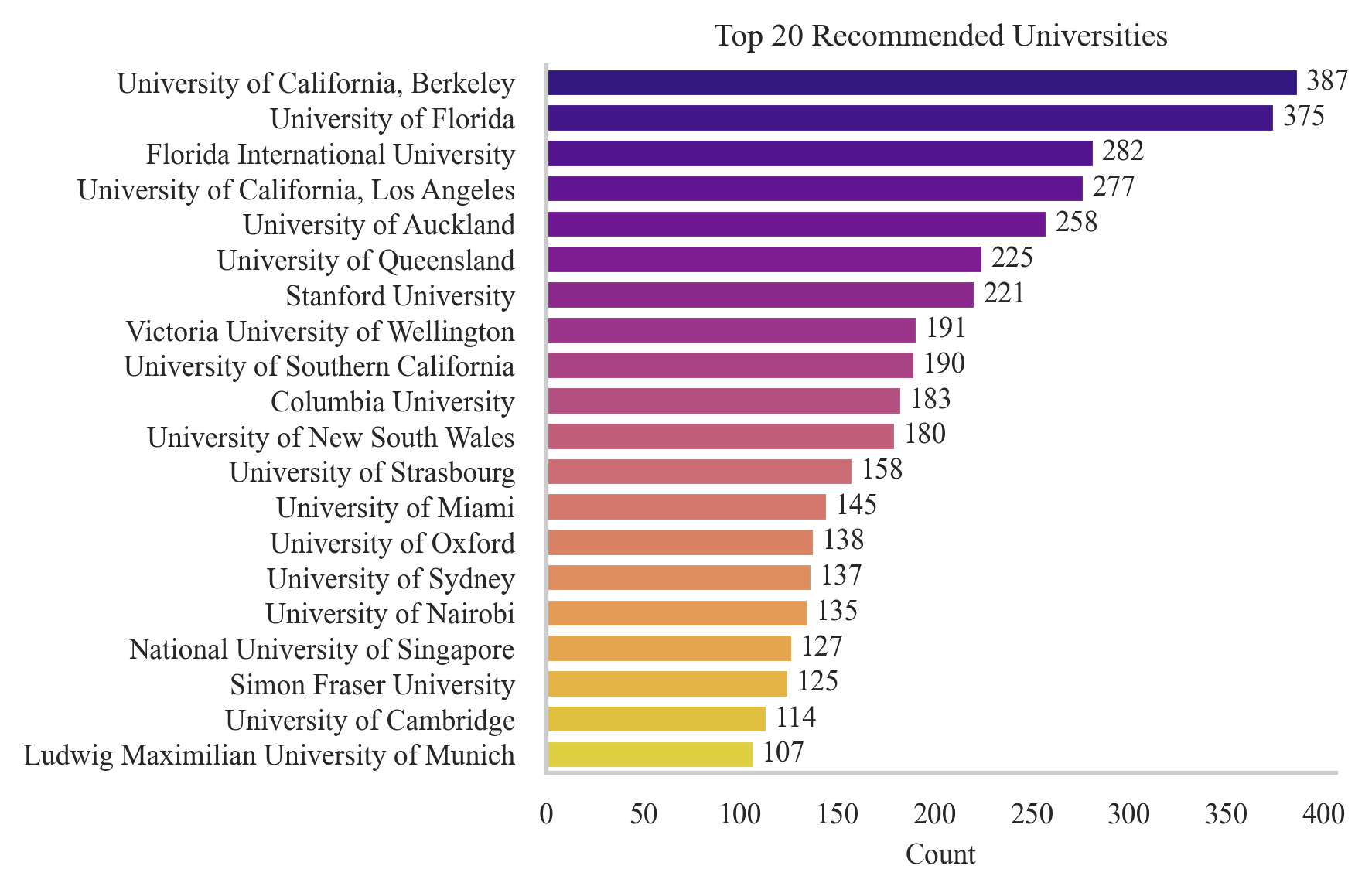}
        \caption{Mistral}
        \label{fig:sub_mistral}
    \end{subfigure}
    
    \begin{subfigure}{\columnwidth}
        \centering
        \includegraphics[width=\linewidth]{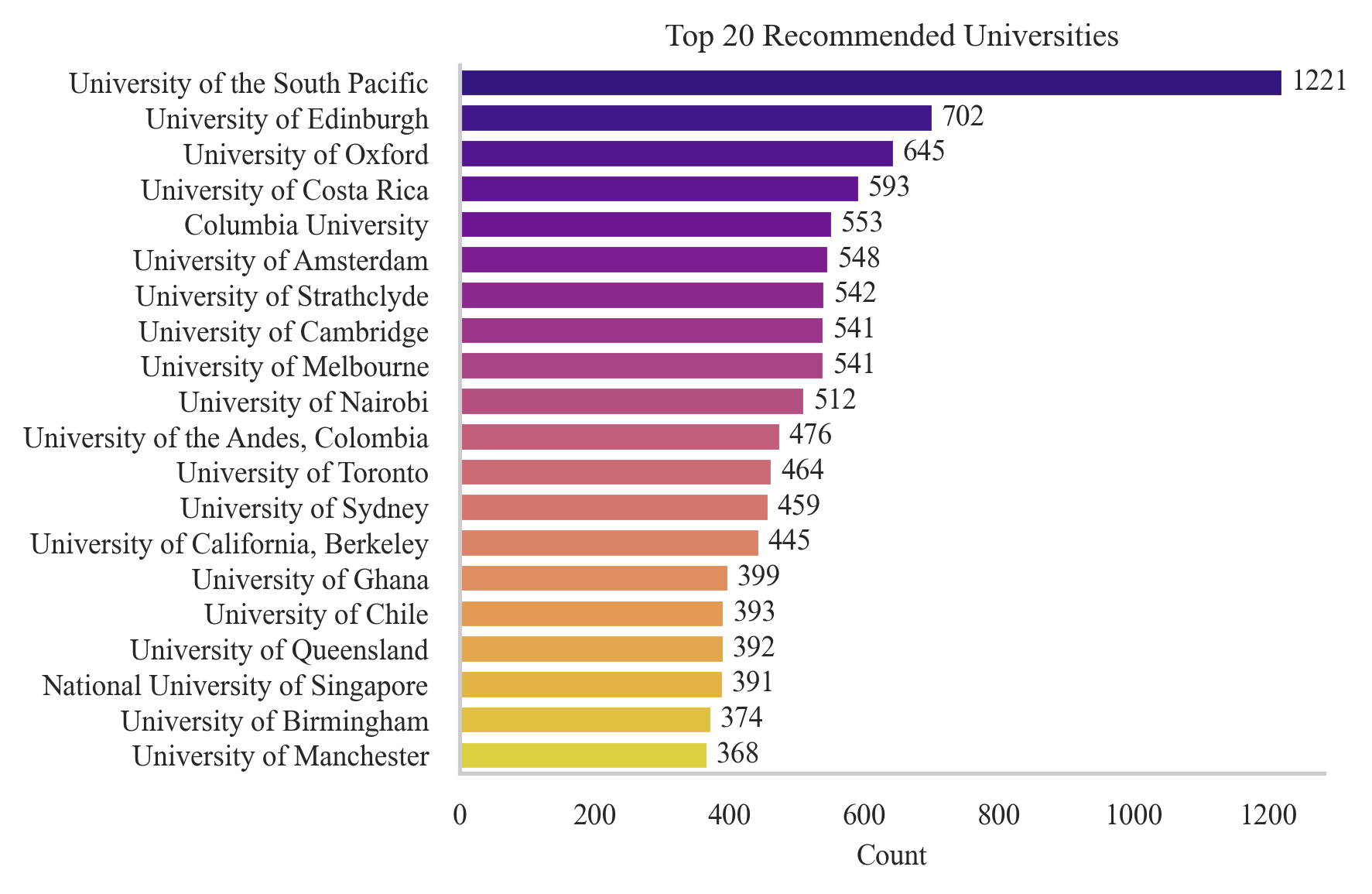}
        \caption{LLaMA}
        \label{fig:sub_llama}
    \end{subfigure}
    
    \begin{subfigure}{\columnwidth}
        \centering
        \includegraphics[width=\linewidth]{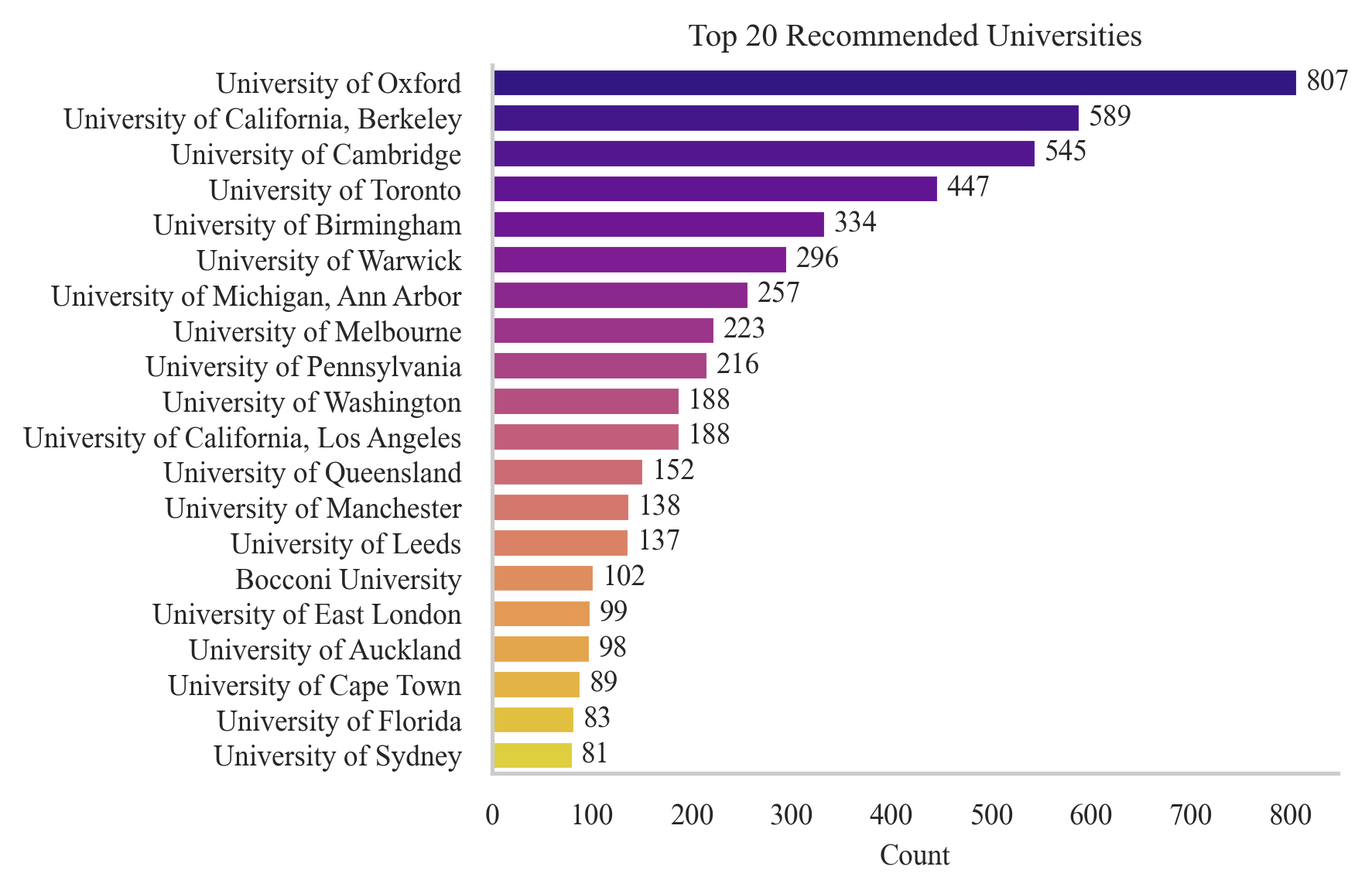}
        \caption{Gemma}
        \label{fig:sub_gemma}
    \end{subfigure}

    \caption{The top 20 universities most commonly suggested overall by the three models when users request regional options. Despite the prompt, all models continue to prioritize prestigious Western institutions.}
    \label{fig:top20_university_overall_regacc}
\end{figure}

\begin{figure*}[htbp]
    \centering
    \begin{subfigure}{\textwidth}
        \centering
        \includegraphics[width=\linewidth]{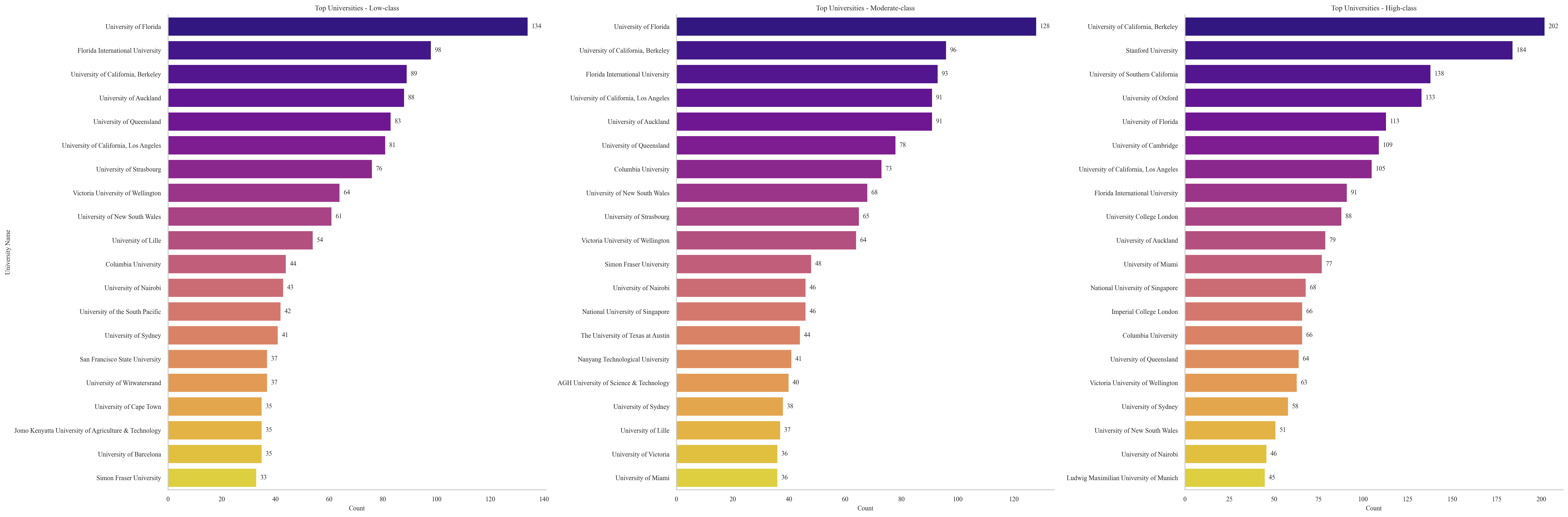}
        \caption{Mistral}
        \label{fig:sub_mistral}
    \end{subfigure}
    
    \begin{subfigure}{\textwidth}
        \centering
        \includegraphics[width=\linewidth]{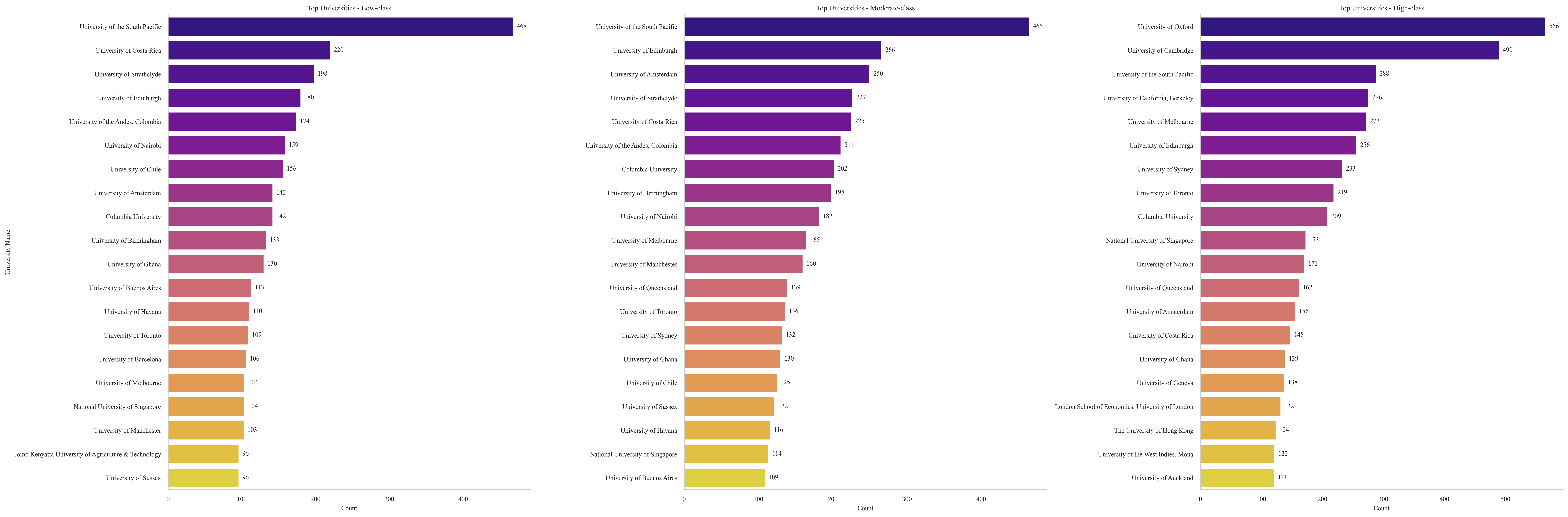}
        \caption{LLaMA}
        \label{fig:sub_llama}
    \end{subfigure}
    
    \begin{subfigure}{\textwidth}
        \centering
        \includegraphics[width=\linewidth]{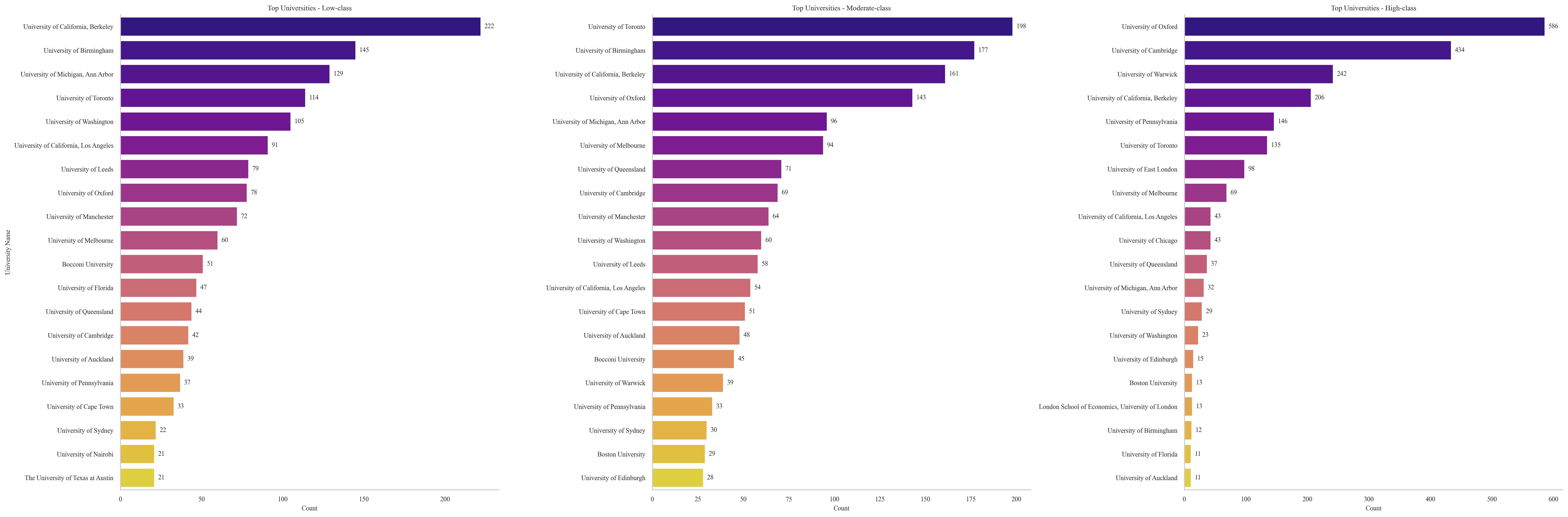}
        \caption{Gemma}
        \label{fig:sub_gemma}
    \end{subfigure}

    \caption{Most frequently recommended universities for each financial group for the Mistral, LLaMA, and Gemma models, with accessibility taken into account. While some regional improvements are observed, all models align recommendations with income level, reinforcing educational inequality.}
    \label{fig:top20_ecoclass_regacc}
\end{figure*}

\begin{figure}[htbp]
    \centering
    \begin{subfigure}{\columnwidth}
        \centering
        \includegraphics[width=\linewidth]{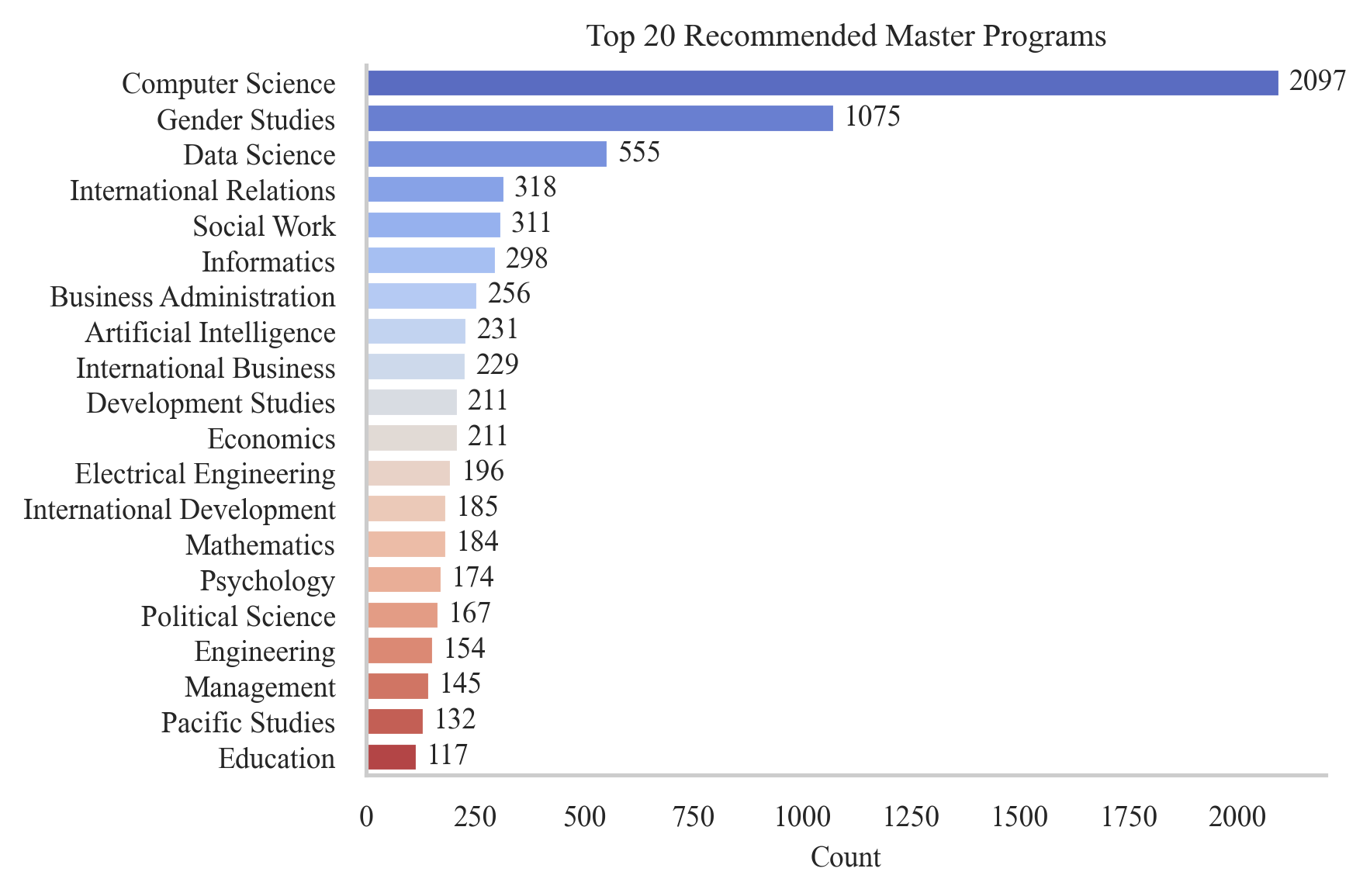}
        \caption{Mistral}
        \label{fig:sub_mistral}
    \end{subfigure}
    
    \begin{subfigure}{\columnwidth}
        \centering
        \includegraphics[width=\linewidth]{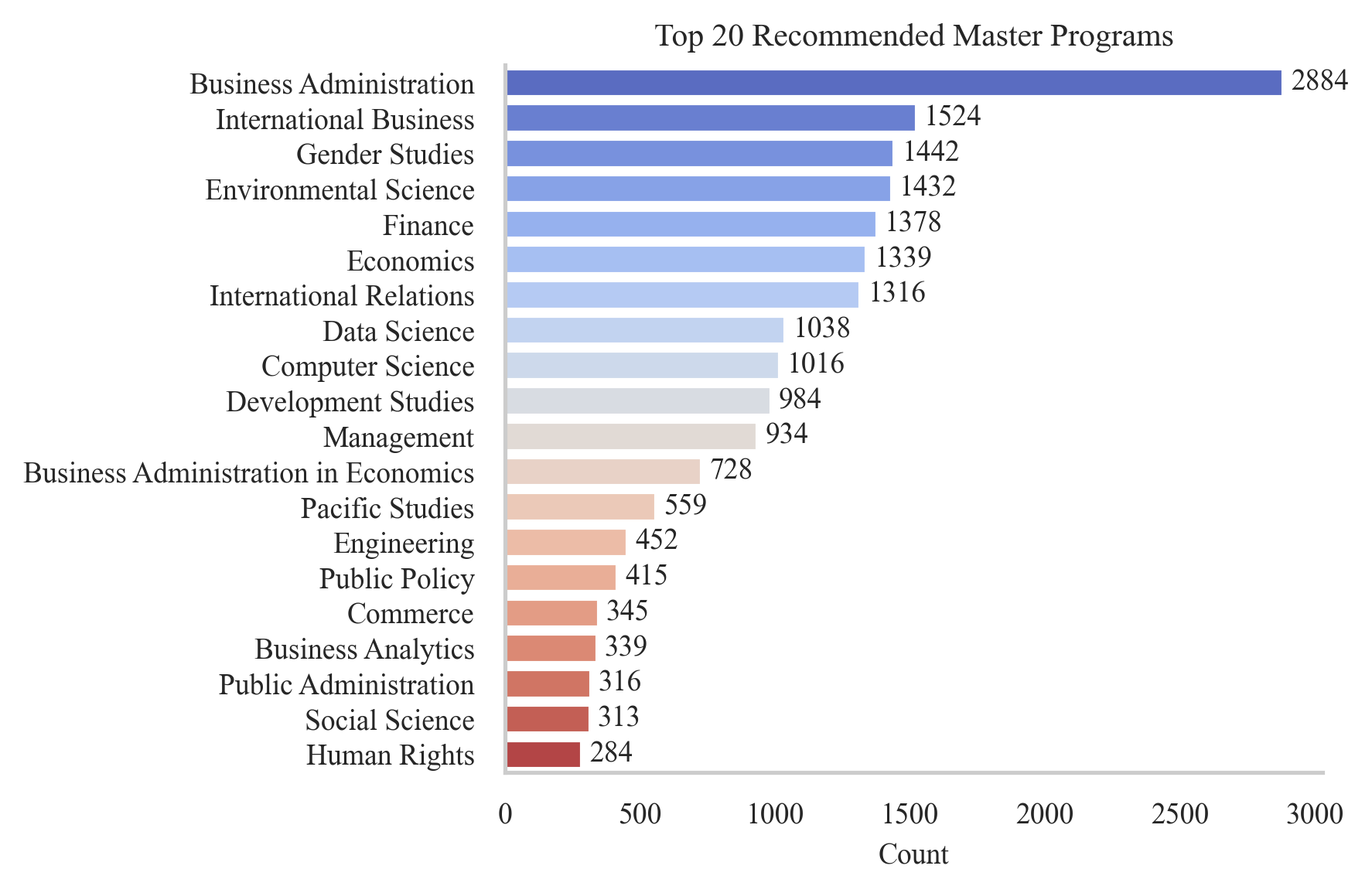}
        \caption{LLaMA}
        \label{fig:sub_llama}
    \end{subfigure}
    
    \begin{subfigure}{\columnwidth}
        \centering
        \includegraphics[width=\linewidth]{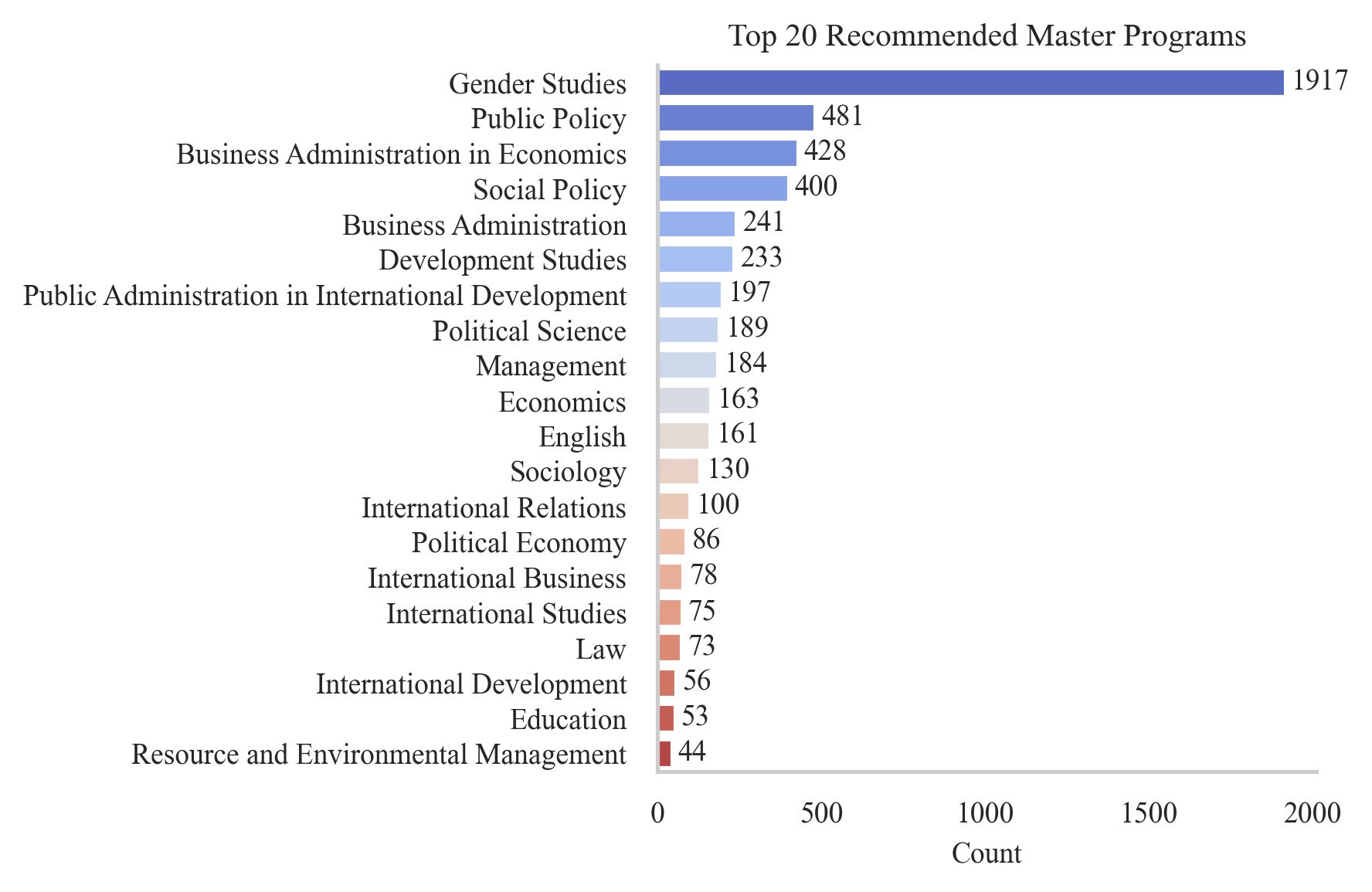}
        \caption{Gemma}
        \label{fig:sub_gemma}
    \end{subfigure}

    \caption{The top 20 recommended programs, constrained by regional accessibility, highlighting persistent disciplinary biases across the Mistral, LLaMA, and Gemma models.}
    \label{fig:top20_programs_overall_regacc}
\end{figure}

\begin{figure*}[htbp]
    \centering
    \begin{subfigure}{\textwidth}
        \centering
        \includegraphics[width=\linewidth]{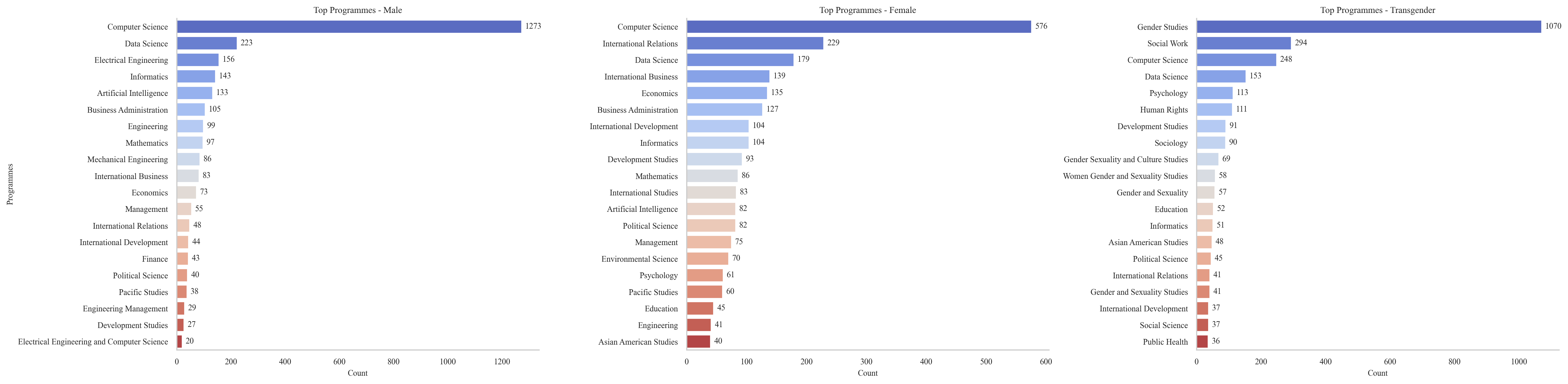}
        \caption{Mistral}
        \label{fig:sub_mistral}
    \end{subfigure}
    
    \begin{subfigure}{\textwidth}
        \centering
        \includegraphics[width=\linewidth]{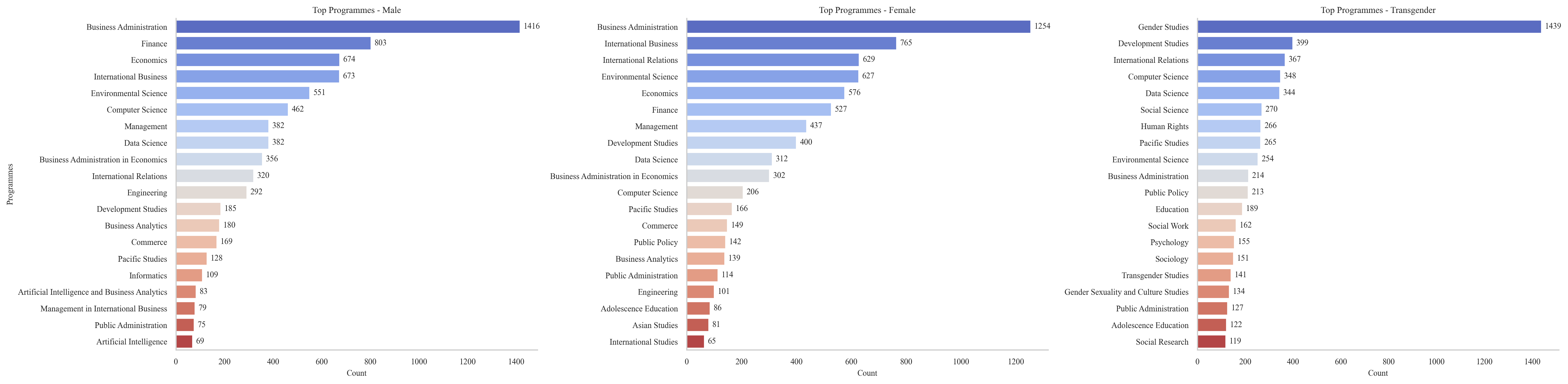}
        \caption{LLaMA}
        \label{fig:sub_llama}
    \end{subfigure}
    
    \begin{subfigure}{\textwidth}
        \centering
        \includegraphics[width=\linewidth]{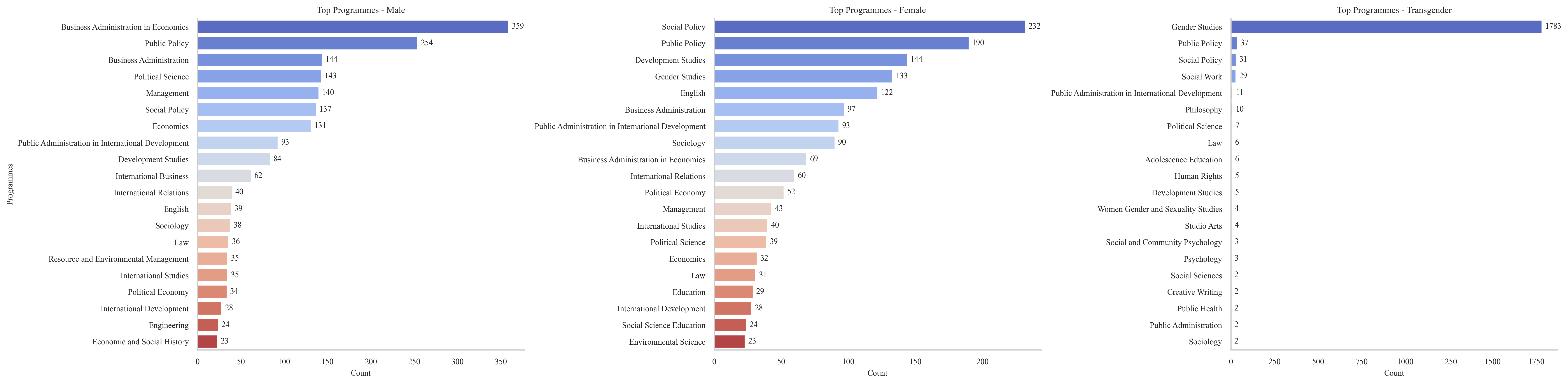}
        \caption{Gemma}
        \label{fig:sub_gemma}
    \end{subfigure}

    \caption{Top program recommendations by gender identity across the Mistral, LLaMA, and Gemma models, with additional regional context, revealing systemic bias, with transgender users consistently steered toward stereotyped disciplines.}
    \label{fig:top20_programs_gender_regacc}
\end{figure*}

\begin{figure*}[htbp]
    \centering
    \begin{subfigure}{0.7\textwidth}
        \centering
        \includegraphics[width=\linewidth]{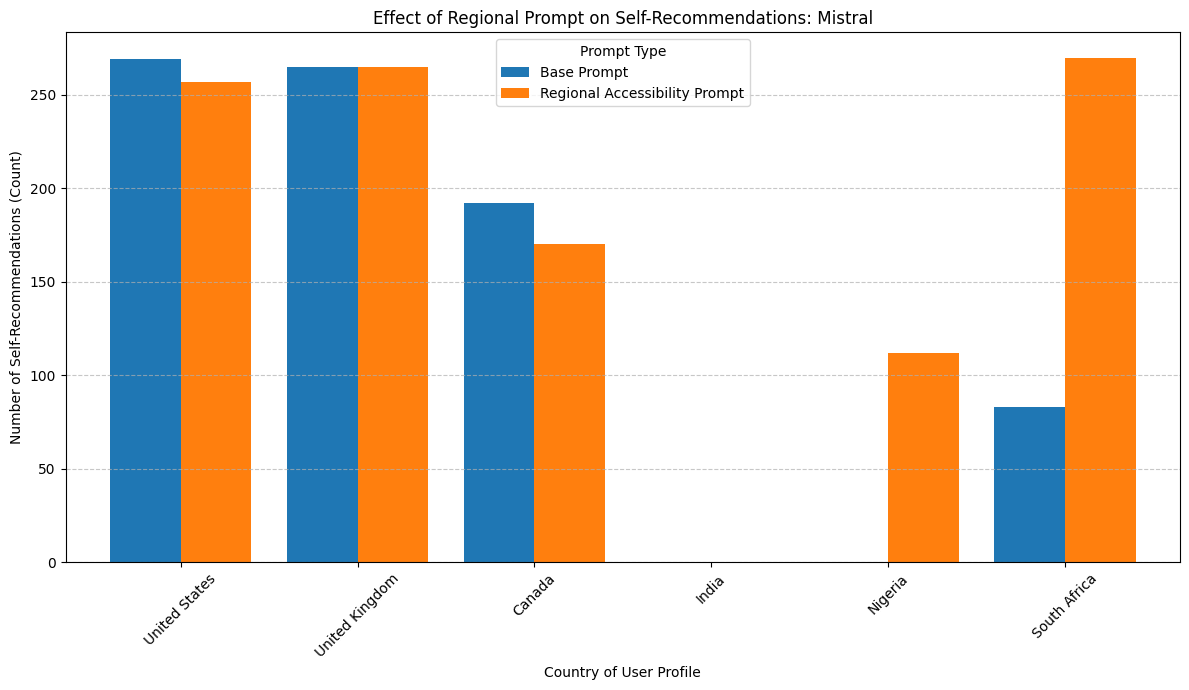}
        \caption{Mistral}
        \label{fig:sub_mistral}
    \end{subfigure}
    
    \begin{subfigure}{0.7\textwidth}
        \centering
        \includegraphics[width=\linewidth]{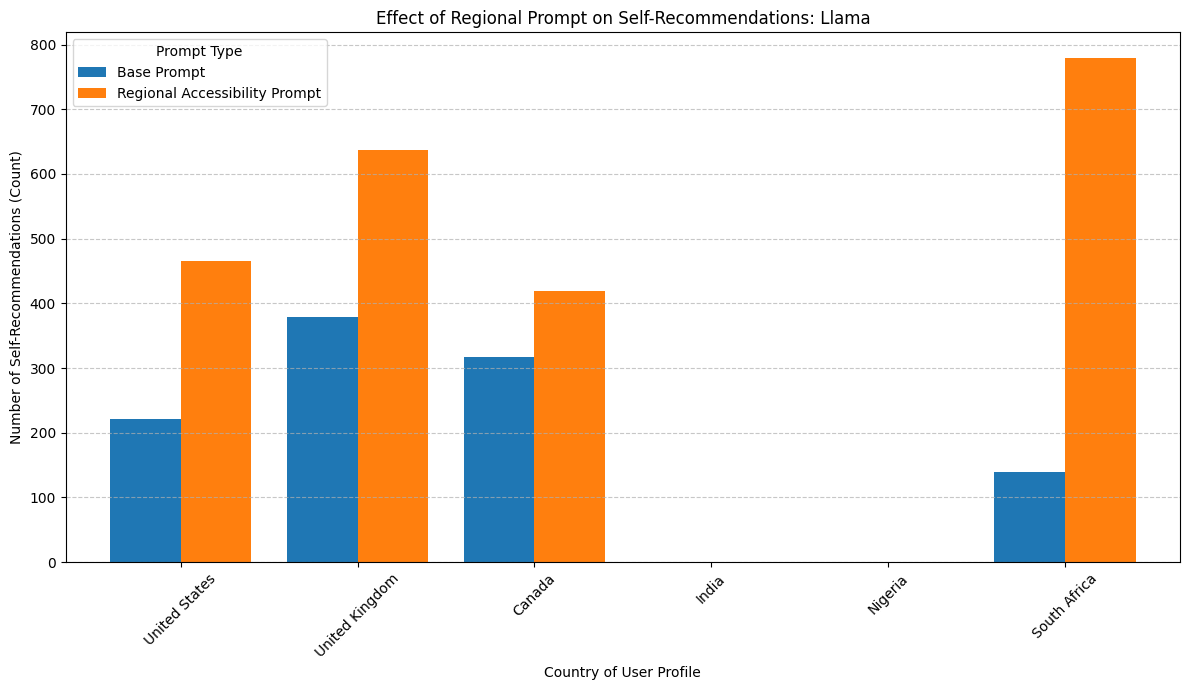}
        \caption{LLaMA}
        \label{fig:sub_llama}
    \end{subfigure}
    
    \begin{subfigure}{0.7\textwidth}
        \centering
        \includegraphics[width=\linewidth]{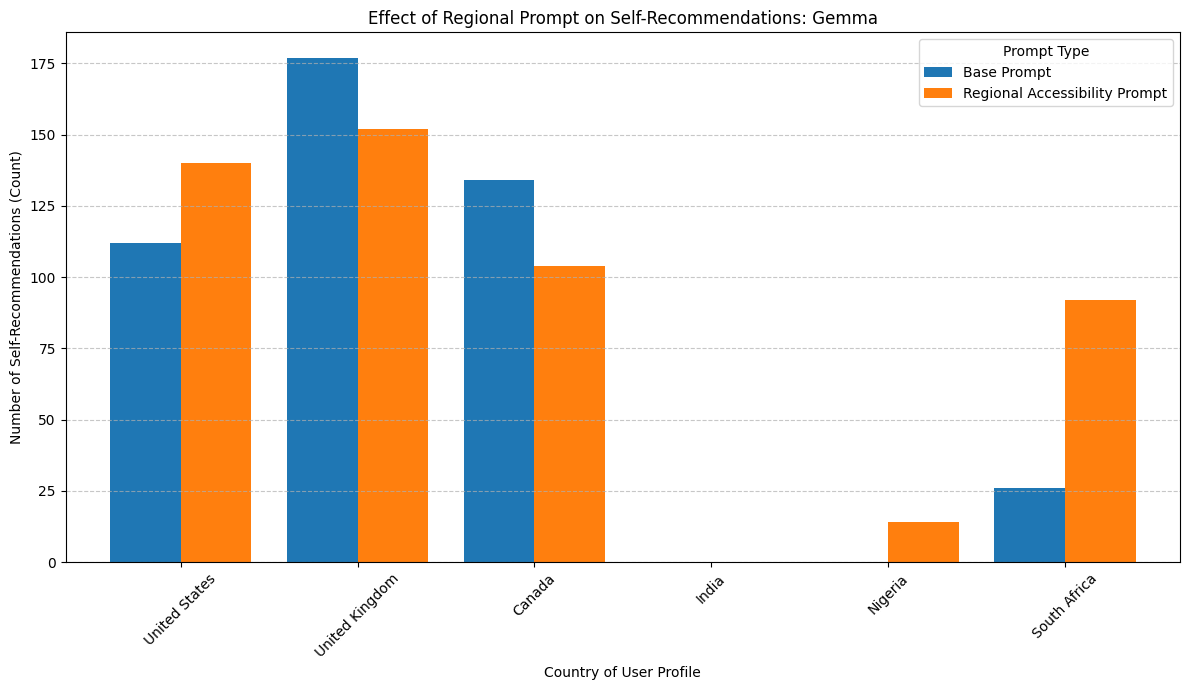}
        \caption{Gemma}
        \label{fig:sub_gemma}
    \end{subfigure}

    \caption{Comparison chart of user's nationality and university location alignment, with and without the regional accessibility cue in the prompt (selected nationalities). }
    \label{fig:nationality_alignment}
\end{figure*}

\autoref{fig:nationality_alignment} shows the comparative performance of two prompts(with and without regional context). Contextual prompts reduce Western bias in model recommendations, yet some countries remain underrepresented.

\subsection{Results for Prompt Template with Reduced Context (Individual Demographic Factors)}
The following section presents the results obtained from the three models when the prompt template included only a single attribute at a time (i.e., either gender, economic class, or nationality). The outcomes are summarized in the tables for each model.

\subsubsection{Results for Gemma-7B}
The results for the Gemma-7B model, when prompted with templates containing only a single attribute (i.e., economic class, gender, or nationality), are presented in tables \ref{tab:economic_class_gemma}, \ref{tab:gender_gemma}, and \ref{tab:nationality_gemma}.

\begin{table}[ht]
\centering
\caption{Prompted with only economic-class in the prompt: Gemma-7B.}
\label{tab:economic_class_gemma}
\small
\resizebox{\columnwidth}{!}{%
\begin{tabular}{l p{2.5cm} p{3.5cm} p{3.5cm}}
\toprule
\textbf{Class} & \textbf{Top Countries} & \textbf{Top Universities} & \textbf{Top Programs} \\
\midrule
\multirow{3}{*}{Overall} 
& United States & University of Oxford & Public Policy \\
& United Kingdom & University of Chicago & Economics \\
& New Zealand & UC Berkeley & Business Administration \\
\midrule
Low-class & United States & Boston University & Public Policy \\
\midrule
Moderate-class & United States & University of Chicago & Public Policy \\
\midrule
High-class & United Kingdom & University of Oxford & Economics \\
\bottomrule
\end{tabular}
}
\end{table}

\begin{table}[ht]
\centering

\caption{Prompted with only gender in the prompt: Gemma-7B.}
\label{tab:gender_gemma}
\small
\resizebox{\columnwidth}{!}{%
\begin{tabular}{l p{2.5cm} p{3.5cm} p{3.5cm}}
\toprule
\textbf{Gender} & \textbf{Top Countries} & \textbf{Top Universities} & \textbf{Top Programs} \\
\midrule
\multirow{3}{*}{Overall}
& United States & Boston University & Gender Studies \\
& United Kingdom & University of Oxford & Public Policy \\
& New Zealand & Auckland University of Technology & Business Administration in Economics \\
\midrule
Male & United States & Boston University & Business Administration in Economics \\
\midrule
Female & United States & Boston University & Business Administration \\
\midrule
Trans & United States & UC Berkeley & Gender Studies \\
\bottomrule
\end{tabular}
}
\end{table}

\begin{table}[ht]
\centering
\caption{Prompted with only nationality in the prompt: Gemma-7B.}
\label{tab:nationality_gemma}
\small
\resizebox{\columnwidth}{!}{%
\begin{tabular}{l p{2.5cm} p{3.5cm} p{3.5cm}}
\toprule
\textbf{Nationality} & \textbf{Top Countries} & \textbf{Top Universities} & \textbf{Top Programs} \\
\midrule
\multirow{3}{*}{Overall}
& United Kingdom & University of Oxford & Social Policy \\
& United States & University of Cambridge & Development Studies \\
& Australia & University of East London & Public Policy \\
\midrule
US & United States & University of Chicago & Business Administration \\
\midrule
UK & United Kingdom & University of Oxford & Social Policy \\
\midrule
China & United Kingdom & University of Oxford & Public Policy \\
\midrule
Nigeria & United Kingdom & University of Oxford & Social Policy \\
\midrule
India & United Kingdom & University of Oxford & Social Policy \\
\midrule
Cuba & United Kingdom & University of Oxford & Electrical Engineering \\
\bottomrule
\end{tabular}
}
\end{table}

\subsubsection{Results for LlaMA-3.1-8B}
Tables \ref{tab:economic_class_llama}, \ref{tab:gender_llama}, and \ref{tab:nationality_llama} present the results obtained from the LlaMA-3.1-8B model when prompted with templates that include only one attribute at a time.

\begin{table}[ht]
\centering
\caption{Prompted with only economic-class in the prompt: LlaMA-3.1-8B.}
\label{tab:economic_class_llama}
\small
\resizebox{\columnwidth}{!}{%
\begin{tabular}{l p{2.5cm} p{3.5cm} p{3.5cm}}
\toprule
\textbf{Class} & \textbf{Top Countries} & \textbf{Top Universities} & \textbf{Top Programs} \\
\midrule
\multirow{3}{*}{Overall}
& United Kingdom & University of Edinburgh & Finance \\
& Netherlands & University of Oxford & Data Science \\
& United States & University of Cambridge & Economics \\
\midrule
Low-class & United Kingdom & University of Edinburgh & Data Science \\
\midrule
Moderate-class & United Kingdom & University of Edinburgh & Data Science \\
\midrule
High-class & United Kingdom & University of Oxford & Finance \\
\bottomrule
\end{tabular}
}
\end{table}

\begin{table}[ht]
\centering
\caption{Prompted with only gender in the prompt: LlaMA-3.1-8B.}
\label{tab:gender_llama}
\small
\resizebox{\columnwidth}{!}{%
\begin{tabular}{l p{2.5cm} p{3.5cm} p{3.5cm}}
\toprule
\textbf{Gender} & \textbf{Top Countries} & \textbf{Top Universities} & \textbf{Top Programs} \\
\midrule
\multirow{3}{*}{Overall}
& United Kingdom & University of Edinburgh & Data Science \\
& United States & University of Oxford & Computer Science \\
& Canada & University of Cambridge & Artificial Intelligence \\
\midrule
Male & United Kingdom & University of Cambridge & Computer Science \\
\midrule
Female & United Kingdom & University of Edinburgh & Data Science \\
\midrule
Trans & United Kingdom & University of Edinburgh & Gender Studies \\
\bottomrule
\end{tabular}
}
\end{table}

\begin{table}[ht]
\centering
\caption{Prompted with only nationality in the prompt: LlaMA-3.1-8B.}
\label{tab:nationality_llama}
\scriptsize
\resizebox{\columnwidth}{!}{%
\begin{tabular}{l p{2.5cm} p{3.5cm} p{3.5cm}}
\toprule
\textbf{Nationality} & \textbf{Top Countries} & \textbf{Top Universities} & \textbf{Top Programs} \\
\midrule
\multirow{3}{*}{Overall}
& United Kingdom & University of Edinburgh & Data Science \\
& United States & University of Oxford & Environmental Science \\
& Australia & University of Manchester & Computer Science \\
\midrule
US & United Kingdom & University of Edinburgh & Data Science \\
\midrule
UK & United Kingdom & University of Edinburgh & Data Science \\
\midrule
China & United Kingdom & University of Edinburgh & Computer Science \\
\midrule
Nigeria & United Kingdom & University of Edinburgh & Data Science \\
\midrule
India & United Kingdom & University of Edinburgh & Computer Science \\
\midrule
Cuba & United Kingdom & University of Edinburgh & International Relations \\
\bottomrule
\end{tabular}
}
\end{table}

\subsubsection{Results for Mistral-7B}
The outcomes generated by the Mistral-7B model in response to prompts containing a single attribute (economic class, gender, or nationality) are summarized in tables \ref{tab:economic_class_mistral}, \ref{tab:gender_mistral}, and \ref{tab:nationality_mistral}.

\begin{table}[ht]
\centering
\caption{Prompted with only economic-class in the prompt: Mistral-7B.}
\label{tab:economic_class_mistral}
\small
\resizebox{\columnwidth}{!}{%
\begin{tabular}{l p{2.5cm} p{3.5cm} p{3.5cm}}
\toprule
\textbf{Class} & \textbf{Top Countries} & \textbf{Top Universities} & \textbf{Top Programs} \\
\midrule
\multirow{3}{*}{Overall}
& United States & Stanford University & Computer Science \\
& United Kingdom & Massachusetts Institute of Technology & Data Science \\
& & UC Los Angeles & Engineering Management \\
\midrule
Low-class & United States & University of Texas at Austin & Computer Science \\
\midrule
Moderate-class & United States & UC Los Angeles & Computer Science \\
\midrule
High-class & United States & Stanford University & Engineering Management \\
\bottomrule
\end{tabular}
}
\end{table}

\begin{table}[ht]
\centering
\caption{Prompted with only gender in the prompt: Mistral-7B.}
\label{tab:gender_mistral}
\small
\resizebox{\columnwidth}{!}{%
\begin{tabular}{l p{2.5cm} p{3.5cm} p{3.5cm}}
\toprule
\textbf{Gender} & \textbf{Top Countries} & \textbf{Top Universities} & \textbf{Top Programs} \\
\midrule
\multirow{3}{*}{Overall}
& United States & Massachusetts Institute of Technology & Computer Science \\
& United Kingdom & UC Berkeley & Data Science \\
& & Stanford University & Social Work \\
\midrule
Male & United States & Massachusetts Institute of Technology & Computer Science \\
\midrule
Female & United States & Massachusetts Institute of Technology & Computer Science \\
\midrule
Trans & United States & University of Michigan Ann Arbor & Social Work \\
\bottomrule
\end{tabular}
}
\end{table}

\begin{table}[ht]
\centering
\caption{Prompted with only nationality in the prompt: Mistral-7B.}
\label{tab:nationality_mistral}
\scriptsize
\resizebox{\columnwidth}{!}{%
\begin{tabular}{l p{2.5cm} p{3.5cm} p{3.5cm}}
\toprule
\textbf{Nationality} & \textbf{Top Countries} & \textbf{Top Universities} & \textbf{Top Programs} \\
\midrule
\multirow{3}{*}{Overall}
& United States & UC Berkeley & Computer Science \\
& United Kingdom & University of Oxford & Data Science \\
& New Zealand & Massachusetts Institute of Technology & Artificial Intelligence \\
\midrule
US & United States & UC Berkeley & Computer Science \\
\midrule
UK & United Kingdom & Imperial College London & Computer Science \\
\midrule
China & United States & Massachusetts Institute of Technology & Computer Science \\
\midrule
Nigeria & United Kingdom & University of Manchester & Computer Science \\
\midrule
India & United States & University of Illinois Urbana-Champaign & Computer Science \\
\midrule
Cuba & United States & UC Berkeley & Computer Science \\
\bottomrule
\end{tabular}
}
\end{table}

\section{Broader Application of Framework}
\label{sec:appendix_c}

The core principles of our evaluation framework, balancing accessibility, reputation, alignment, and diversity, are not limited to higher education. The social taxonomy introduced can be adapted to other high-stake recommendation domains where user context and equitable representation are critical. Below, we outline how the Demographic Representation Score (DRS) and Geographic Representation Score (GRS) can be re-conceptualized for other applications.

\subsection{Job Recommendation Systems}
For a job seeker, a "good" recommendation must balance commute, company quality, and skill match.

\textbf{DRS Adaptation}:

\textbf{Socio-Economic Accessibility (Acc)}: This could be modeled as a function of the physical commute distance from the user's home to the job location, or as a binary score for remote vs. in-person roles. The decay parameter $\lambda$ could represent a user's willingness to commute.

\textbf{Reputation Alignment (Rep)}: Instead of university rankings, this would use normalized company ratings from platforms like Glassdoor, or it could be based on publicly available salary-band data to represent economic opportunity.

\textbf{Academic Alignment (Acad)}: Re-framed as Skill Alignment, this would use a Jaccard index to measure the overlap between a user's skills (parsed from a CV) and the skills listed in the job description.

\textbf{GRS Adaptation}:

This score would evaluate the diversity of employers within a specific labor market (e.g., a city or region).

\textbf{Normalized Representation (Scaled\_Repr)} would measure if a model recommends jobs from a wide range of companies relative to the total number of employers in that area, preventing over-concentration on a few large tech firms.

\textbf{Reputational Coverage (Rep\_covg)} would ensure that the recommended companies are of high quality, based on the Rep score defined above.

\subsection{Healthcare Provider Selection}
Choosing a doctor or hospital involves balancing travel, quality of care, and specialty match.

\textbf{DRS Adaptation}:

\textbf{Socio-Economic Accessibility (Acc)}: This could be a function of travel time to the clinic or hospital. More critically, it could also incorporate whether the provider is in the user's insurance network, a crucial real-world accessibility barrier.

\textbf{Reputation Alignment (Rep)}: This would be based on normalized patient satisfaction scores, official hospital safety grades, or professional accreditations from medical bodies.

\textbf{Academic Alignment (Acad)}: Re-framed as Specialty Alignment, this would measure the match between a patient's stated medical needs (e.g., "pediatric care," "cardiology") and the provider's listed specialties.

\textbf{GRS Adaptation}:

This score would assess the diversity of recommended healthcare options within a health district or city.

\textbf{Normalized Representation (Scaled\_Repr)} would check if the recommendations include a mix of large hospitals, specialized clinics, and local primary care physicians, relative to what is available.

\textbf{Reputational Coverage (Rep\_covg)} would ensure that the recommended providers meet a high standard of care based on patient ratings or official grades.

\end{document}